\documentclass[10pt]{article}

\usepackage{hyperref}
\usepackage[utf8]{inputenc} 
\usepackage[T1]{fontenc}    
\usepackage{url}            
\usepackage{booktabs}       
\usepackage{amsfonts}       
\usepackage{nicefrac}       
\usepackage{microtype}      
\usepackage{amsmath,amssymb,subfigure,epsfig,bbm,mathrsfs,mathrsfs,dsfont, upgreek}
\usepackage{pstricks,framed}
\usepackage{subfigure}
\usepackage{amstext}
\usepackage{array,xcolor}
\usepackage{color,soul}
\usepackage{hyperref}
\usepackage{caption}
\usepackage[all,cmtip]{xy}
\usepackage{chemfig}
\usepackage{makecell}
\usepackage{diagbox}
\usepackage{rotating}
\usepackage{easybmat}
\usepackage{multirow,bigdelim}
\usepackage{makeidx,overpic}
\usepackage{mdframed,algorithmic}
\usepackage[lined,boxruled]{algorithm2e}
\usepackage{boldline}
\usepackage{float}

\DeclareGraphicsExtensions{.pdf}

\DeclareMathAlphabet{\mathpzc}{OT1}{pzc}{m}{it}
\DeclareFontFamily{OT1}{pzc}{}
\DeclareFontShape{OT1}{pzc}{m}{it}{<-> s * [1.2] pzcmi7t}{}
\DeclareMathAlphabet{\mathpzc}{OT1}{pzc}{m}{it}
\DeclareMathAlphabet{\mathbbit}{U}{bbm}{m}{sl}
\DeclareMathOperator*{\argmin}{arg\,min}
\newcommand{\AX}{\mathcal{A}_{\mX{}^{in}}}

\makeindex

\newtheorem{lemma}{Lemma}
\newtheorem{definition}{Definition}

\newtheorem{theorem}{Theorem}

\newcommand{\supp}{\mbox{supp}}
  
\newcommand{\balpha}{{\boldsymbol{\alpha}}}
\newcommand{\bbeta}{{\boldsymbol{\beta}}}
\newcommand{\bmu}{{\boldsymbol{\mu}}}

\newcommand{\bphi}{{\boldsymbol{\varphi}}}
\newcommand{\bsigma}{{\boldsymbol{\Sigma}}}
\newcommand{\ReLU}{{\mathtt{ReLU}}}
\newcommand{\net}{{\mathcal{N}\!et}}

\newcommand{\vct}[1]{\boldsymbol{#1}}

\DeclareMathOperator*{\minimize}{\text{minimize}}

\newcommand{\vzero}{\vct{0}}

\newcommand{\vb}{\vct{b}}

\newcommand{\ve}{\vct{e}}

\newcommand{\vg}{\vct{g}}
\newcommand{\vh}{\vct{h}}

\newcommand{\vv}{\vct{v}}
\newcommand{\vw}{\vct{w}}
\newcommand{\vx}{\vct{x}}
\newcommand{\vy}{\vct{y}}
\newcommand{\vz}{\vct{z}}

\newcommand{\mtx}[1]{\boldsymbol{#1}}
\newcommand{\mA}{\mtx{A}}
\newcommand{\mB}{\mtx{B}}
\newcommand{\mC}{\mtx{C}}

\newcommand{\mE}{\mtx{E}}

\newcommand{\mI}{\mtx{I}}

\newcommand{\mP}{\mtx{P}}

\newcommand{\mR}{\mtx{R}}

\newcommand{\mT}{\mtx{T}}
\newcommand{\mU}{\mtx{U}}
\newcommand{\mV}{\mtx{V}}
\newcommand{\mW}{\mtx{W}}
\newcommand{\mX}{\mtx{X}}
\newcommand{\mY}{\mtx{Y}}
\newcommand{\mZ}{\mtx{Z}}

\newcommand{\vphi}{\vct{\phi}}

\begin{document}

\title{\vspace{-2cm}\bf{Fast Convex Pruning of Deep Neural Networks}}
\author{Alireza Aghasi\thanks{(Corresponding Author) Robinson College of Business, Georgia State University, Atlanta, GA. Email: {\tt aaghasi@gsu.edu}} \and  Afshin Abdi${}^\dagger$ \and Justin Romberg\thanks{School of Electrical and Computer Engineering, Georgia Tech, Atlanta, GA. Emails:  {\tt \{abdi,jrom\}@ece.gatech.edu}.}}

\date{}    


\maketitle

\begin{abstract}
We develop a fast, tractable technique called Net-Trim for simplifying a trained neural network.    The method is a convex post-processing module, which prunes (sparsifies) a trained network layer by layer, while preserving the internal responses. We present a comprehensive analysis of Net-Trim from both the algorithmic and sample complexity standpoints, centered on a fast, scalable convex optimization program.  Our analysis includes consistency results between the initial and retrained models before and after Net-Trim application and guarantees on the number of training samples needed to discover a network that can be expressed using a certain number of nonzero terms.  Specifically, if there is a set of weights that uses at most $s$ terms that can re-create the layer outputs from the layer inputs, we can find these weights from $\mathcal{O}(s\log N/s)$ samples, where $N$ is the input size.  These theoretical results are similar to those for sparse regression using the Lasso, and our analysis uses some of the same recently-developed tools (namely recent results on the concentration of measure and convex analysis). Finally, we propose an algorithmic framework based on the alternating direction method of multipliers (ADMM), which allows a fast and simple implementation of Net-Trim for network pruning and compression.
\end{abstract}

\textbf{Keywords:} Pruning Neural Networks, Deep Neural Networks, Compressed Sensing, Bowling Scheme, Rademacher Complexity

\section{Introduction}
Deep neural networks are becoming a prominent tool to learn data structures of arbitrary complexity. This success is mainly thanks to their flexible, yet compact nonlinear formulation, and the development of computational and architectural techniques to improve their training (c.f. \cite{schmidhuber2015deep, goodfellow16de} for a comprehensive review). Increasing the number of layers, and the number of neurons within each layer is generally the most standard way of adding more flexibility to a neural network. While adding such flexibility is capable of improving the fitting of the model to the training data (i.e., reducing the model bias), it makes the models prone to over-parameterization and overfitting (i.e., increasing the model variance), which in turn can degrade the predictive capability of the network.  


To simplify or stabilize neural networks, various regularizing techniques and pruning strategies have been considered. Inspired by the classic regularizers for linear models, such as Ridge \cite{hoerl1970ridge} and Lasso \cite{tibshirani1996regression}, the training of neural networks is also equipped with $\ell_2$ or $\ell_1$ penalties \cite{nowlan1992simplifying, girosi1995regularization} to control their variance and complexity. Adding randomness to the training process is also shown to have regularizing effects, relevant to which we may refer to Dropout \cite{SHKSS2014} and DropConnect \cite{WZZLF2013}, which randomly remove active connections in the training phase and are likely to produce pruned networks. Batch normalization \cite{ioffe2015batch}, associated with stochastic gradient descent-type fitting techniques, can also be considered as a tool of similar nature, where in the training process the updates of the hidden units are weighted by the standard deviation of the random examples included in the mini-batch.

In this paper, we advocate a different approach.  We train the network using standard techniques.  We then extract the internal outputs  (the intermediate features) at each layer and find a sparse set of weights that reproduces these features across all the training data. The philosophy here is that the most important product of training the network is the features that it extracts, not the weights that it settles on to produce those features.  For large networks, there will be many sets of weights that produce exactly the same internal features; of those weights, we choose the simplest.

Our method for finding sparse sets of weights, presented in detail in Section~\ref{sec:prune}, is related to well-known techniques for sparse regression, e.g.\ the Lasso \cite{tibshirani1996regression} in statistics and compressed sensing \cite{candes2006stable} in signal processing.  The main difference is the non-linearity in the mapping of internal features from one layer to another.  If this non-linearity is piecewise linear and convex (as is the rectified linear unit, $\ReLU(\vx)=\max(\vx,\vzero)$, that we use in all of our analysis below), then there is a natural way to recast the condition that the outputs and inputs of a layer match as a set of linear inequality constraints.  There is a similar way to recast an approximate matching as inclusion in a convex set.  Using the $\ell_1$ norm as a proxy for sparsity, the entire program becomes convex.  This opens the door for a thorough analysis of how well and under what conditions we can expect Net-Trim to perform well, and allows us to leverage decades of research in convex optimization to find a scalable algorithm with predictable convergence behavior.

The theory in Section~\ref{sampComplexSec} presents an upper bound on the number of training samples needed to discover a weight matrix that is sparse.  Given a set of layer input vectors $\vx_1,\ldots,\vx_P$ and output vectors $\vy_1,\ldots,\vy_P$,  we solve the program
\begin{equation}
	\label{eq:optintro}
	\minimize_{\mW}~\|\mW\|_1\quad\text{subject to}\quad \ReLU(\mW^\top\vx_p) = \vy_p,~~~~ (p =1,\ldots,P),
\end{equation}
where $\|\mW\|_1 = \sum_{n,m}|w_{n,m}|$ is the sum of the absolute values of the entries in a matrix $\mW\in\mathbb{R}^{N\times M}$.  As the $\ell_1$ norm is convex and the $\ReLU(\cdot)$ function is piecewise linear, meaning that constraints in the program above can be broken into a series of linear equality and inequality constraints, the program above is convex.  We show that if the $\vx_p$ are independent samples of a subgaussian random vector that is non-degenerate (meaning that the correlation matrix is full-rank) and there exists a  $\mW_\star$ with maximally $s$-sparse columns that does indeed satisfy $\vy_p=\ReLU(\mW_\star^\top\vx_p)$ for all $p$, then the solution to \eqref{eq:optintro} is exactly $\mW_\star$ when the number of training samples $P$ is (almost) proportional to the sparsity $s$: we require 
\[
	P~\gtrsim~ s\log(N/s).
\]
We also show that if the $\vx_p$ are subgaussian, then so are the $\vy_p$.  As a results, the theory can be applied layer-by-layer, yielding a sampling result for networks of arbitrary depth.  (When we apply the algorithm in practice, the equality constraints in \eqref{eq:optintro} are relaxed; this is discussed in detail in Section~\ref{sec:layerprune}.) Along with these theoretical guarantees, Net-Trim offers state-of-the-art performance on realistic networks.  In Section~\ref{sec:conc}, we present some numerical experiments that show that compression factors between 10x and 50x (removing 90\% to 98\% of the connections) are possible with very little loss in test accuracy.  


\paragraph{Contributions and relations to previous work}  This paper provides a full description of the Net-Trim method from both a theoretical and algorithmic perspective.  In Section~\ref{sec:prune}, we present our convex formulation for sparsifying the weights in the linear layers of a network; we describe how the procedure can be applied layer-by-layer in a deep network either in parallel or serially (cascading the results), and present consistency bounds for both approaches.  Section~\ref{sampComplexSec} presents our main theoretical result, stated precisely in Theorem~\ref{sampComp}.  This result derives an upper bound on the number of data samples we need to reliably discover a layer that has at most $s$ connections in its linear layer --- we show that if the data samples are random, then these weights can be learned from $\mathcal{O}(s\log N/s)$ samples.  Mathematically, this result is comparable to the sample complexity bounds for the Lasso in performing sparse regression on a linear model (also known as the compressed sensing problem).  Our analysis is based on the bowling scheme \cite{tropp2015convex,mendelson2014learning}; the main technical challenges are adapting this technique to the piecewise linear constraints in the program \eqref{eq:optintro}, and the fact that the input vectors $\{\vx_p\}$ into each layer are non-centered in a way that cannot be accounted for easily.


There are several other examples of techniques for simplifying networks by re-training in the recent literature.  These techniques are typically presented as model compression tools (e.g., \cite{han2015learning, chen2015compressing, han2015deep}) for removing the inherent model redundancies.  In what is perhaps the most closely related work to what we present here, \cite{han2015learning} proposes a pruning scheme that simply truncates small weights of an already trained network, and then re-adjusts the remaining active weights using another round of training.  In contrast, our optimization scheme ensures that the layer inputs and outputs stay consistent as the network is pruned.

The Net-Trim framework was first presented in \cite{Aghasi2017NetTrim}.  This paper provides a far more rigorous and complete analysis (sample complexity bound) of the Net-Trim algorithm for networks with multiple layers (the previous work only considered a single layer of the network).  In addition, we present a scalable (yet relatively simple) implementation of Net-Trim using the alternation direction method of multipliers (ADMM).  This is an iterative method with each iteration requiring a small number of matrix-vector multiplies.  The code, along with all the examples presented in the paper, is available online\footnote{The link to the code and related material: \url{https://dnntoolbox.github.io/Net-Trim/}}.

\paragraph{Notation}
We use lowercase and uppercase boldface for vectors and matrices, respectively. Specifically, the notation $\mI$ is reserved for the identity matrix. For a matrix $\boldsymbol{A}$, $\boldsymbol{A}_{\Gamma_1,:}$ denotes the submatrix formed by restricting the rows of $\boldsymbol{A}$ to the index set $\Gamma_1$.  Similarly, $\boldsymbol{A}_{:,\Gamma_2}$ restricts the columns of $\mA$ to $\Gamma_2$, and $\boldsymbol{A}_{\Gamma_1,\Gamma_2}$ is formed by extracting both rows and columns. 
%
Given a vector $\vx$ (or matrix $\mX$),  $\supp \;\vx$ (or $\supp \;\mX$) is the set of indices with non-zero entries, and $\supp^c~\vx$ (or $\supp^c~\mX$) is the complement set. 

For $\mX=[x_{m,n}]\in\mathbb{R}^{M\times N}$, the matrix trace is denoted by $\mbox{tr}(\mX)$. Furthermore, we use $\|\mX\|_{1}\triangleq\sum_{m=1}^M\sum_{n=1}^N |x_{m,n}|$ as a notation for the sum of absolute entries\footnote{The notation $\|\mX\|_{1}$ should not be confused with the matrix induced $\ell_1$ norm}, and $\|\mX\|_F$ as the Frobenius norm. The neural network activation used throughout the paper is the rectified linear unit (ReLU), which is applied component-wise  to vectors and matrices, 
\[
	\left(\ReLU\left(\mX\right)\right)_{m,n} = \max\left(x_{m,n},0\right).
\]
We will sometimes use the notation $\mX^+$ as shorthand for $\mathtt{ReLU}(\mX)$. 
For an index set $\Omega\subseteq \{1,\cdots,M\}\times\{ 1,\cdots,N\}$, $\mW_\Omega$ represents a matrix of identical size as $\mW=[w_{m,n}]$ with entries
\[\left(\mW_\Omega\right)_{m,n} = \left\{\begin{array}{cc}w_{m,n}&(m,n)\in \Omega\\ 0& (m,n)\notin \Omega\end{array}\right..
\]

Finally, we use $\mathbb{S}^{N}$ to denotes the unit sphere in $\mathbb{R}^{N+1}$;
and the notation $f\gtrsim \hat f$ (or $f\lesssim \hat f$) when there exists an absolute constant $C$ such that $f\geq C\hat f$ (or $f\leq C\hat f$). 

\paragraph{Outline.}
The remainder of the paper is structured as follows. In Section \ref{modelsec}, we briefly overview the neural network architecture considered. Section \ref{sec:prune} presents the pruning idea and the consistency results between the initial and retrained networks. The statistical architecture of the network and the general sample complexity results are presented in Section \ref{sampComplexSec}. To implement the Net-Trim underlying convex program, in Section \ref{app:ADMM} we present an ADMM scheme applicable to the original Net-Trim formulation. Finally, Section \ref{sec:conc} presents some experiments, along with concluding remarks. All the technical proofs of the theorems and results presented in this paper are moved to Section \ref{sec:proof}.

\section{Feedforward Network Model}\label{modelsec}

In this section, we briefly overview the topology of the feedforward network model considered.  The training of the network is performed via $P$ samples $\vx_p$, $p=1,\cdots,P$, where $\vx_p\in \mathbb{R}^N$ is the network input. To compactly represent the training samples, we form a matrix $\mX\in\mathbb{R}^{N\times P}$, structured as $\mX = \left[\vx_1,\cdots,\vx_P\right]$. Considering $L$ layers in the network, the output of the network at the final layer is denoted by $\mX{}^{(L)}\in\mathbb{R}^{N_L\times P}$, where each column in $\mX{}^{(L)}$ is a response to the corresponding training column in $\mX$. 

In a ReLU network, the output of the $\ell$-th layer is $\mX^{(\ell)}\in\mathbb{R}^{N_{\ell}\times P}$, generated by applying the affine transformation $\mW_{ \ell}^\top(\cdot) + \vb^{(\ell)}$ to each column of the previous layer $\mX^{(\ell-1)}$, followed by a ReLU activation:
\begin{equation}
	\label{eqrec0}
	\mX^{(\ell)} = \ReLU\left(\mW_{ \ell}^\top\mX^{(\ell-1)} + \vb^{(\ell)}\boldsymbol{1}^\top \right),\qquad \ell=1,\cdots,L.
\end{equation}
Here $\mW_{\ell}\in\mathbb{R}^{N_{\ell-1}\times N_{\ell}}$, $\mX^{(0)}=\mX$ and $N_0=N$. By adding an additional row to $\mW_{ \ell}$ and $\mX^{(\ell-1)}$, one can absorb the intercept term and compactly rewrite \eqref{eqrec0} as
\begin{equation}
	\label{eqrec}
	\mX^{(\ell)} = \ReLU\left(\mW_{ \ell}^\top\mX^{(\ell-1)}\right),\qquad \ell=1,\cdots,L.
\end{equation}
Often the last layer of a neural network skips an activation by merely going through the affine transformation. As a matter of fact, the results presented in this paper also apply to such architecture (see analysis examples in \cite{Aghasi2017NetTrim}). A neural network that follows the model in \eqref{eqrec} can be fully identified by $\mX$ and $\hat\mW_\ell$, $\ell = 1,\cdots,L$. Throughout the paper, such network will be denoted by $\net(\{\mW_{\ell}\}_{\ell=1}^L;\mX)$.

\section{The Net-Trim Pruning Algorithm}\label{sec:prune}
Net-Trim is a post processing scheme which prunes a neural network after the training phase. Similar to many other regularization techniques, Net-Trim is capable of simplifying trained models at the expense of a controllable increase in the bias. 

After the training phase and learning $\mW_\ell$, Net-Trim retrains the network so that for the same training data the layer outcomes stay more or less close to the initial model, while the redesigned network is sparser, i.e.,
\[\ell = 1,\cdots,L: ~\mbox{nnz}\left( \hat\mW_\ell\right) \ll \mbox{nnz}\left( \mW_\ell \right),~~ \mbox{while}~~ \hat\mX{}^{(\ell)}\approx \mX^{(\ell)}. 
\] 
Here, nnz$(.)$ denotes the number of nonzero entries, and $\hat\mW_\ell$ and $\hat\mX{}^{(\ell)}$ are respectively the redesigned layer matrices and the corresponding layer outcomes. 

Aside from the post-processing nature and some differences in the convex formulations, Net-Trim shares many similarities with the Lasso (least absolute shrinkage and selection operator \cite{tibshirani1996regression}), as they both use an $\ell_1$ proxy to promote model sparsity. In the remainder of this section we overview the Net-Trim formulation and the corresponding pruning schemes.

\subsection{Pruning a Single Layer}
\label{sec:layerprune}

Consider $\mX^{in}\in\mathbb{R}^{N\times P}$ and $\mX^{out}\in\mathbb{R}^{M\times P}$ to be a layer input and output matrices after the training, which based on the model in \eqref{eqrec0} (or \eqref{eqrec}) are connected via
\[	\mX^{out} = \ReLU\left(\mW^\top\mX^{in}\right).
\]
To explore a sparser coefficient matrix, we may consider the minimization 
\begin{equation}\label{eqncx}
\minimize_{\mU}~~\left\|\mU\right\|_{1}\quad \mbox{subject to} \quad \left\|\ReLU\left(\mU^\top\mX^{in}\right)- \mX^{out}\right\|_F\leq\epsilon,
\end{equation}
which may potentially generate a sparser $\mW$-matrix relating $\mX^{in}$ and $\mX^{out}$, at the expense of a (controllable) discrepancy between the layer outcomes before and after the retraining. 

Despite the convex objective, the constraint set in \eqref{eqncx} is non-convex. Using the fact that the entries of $\mX^{out}$ are either zero or strictly positive quantities,  \cite{Aghasi2017NetTrim} propose the following convex proxy to \eqref{eqncx}:    
\begin{equation}\label{eqconv}
\hat\mW = \operatorname*{arg\,min}_{\mU}~\left\|\mU\right\|_{1}\quad \mbox{subject to} \quad \left\{\begin{array}{l}\left\|\left(\mU^\top\mX^{in}- \mX^{out}\right)_\Omega\right\|_F\leq\epsilon\\[.1cm] \left(\mU^\top\mX^{in}\right)_{\Omega^c} \leq \boldsymbol{0} \end{array}\right.,
\end{equation}
where
\begin{equation*}
\Omega = \supp~\mX^{out} = \left\{(m,p): \left[\mX^{out}\right]_{m,p}>0 \right\}.
\end{equation*}
The main idea behind this convex surrogate is imposing similar activation patterns before and after the retraining via the second inequality in \eqref{eqconv}, i.e.,
\[\left(\mW^\top\mX^{in}\right)^+_{\Omega^c} = \left(\hat\mW{}^\top\mX^{in}\right)^+_{\Omega^c} = \boldsymbol{0},
\]
and allowing the $\epsilon$-discrepancy only on the set $\Omega$. 
For a more compact presentation of the convex constraint set, for given matrices $\mX, \mY$ and $\mV$ we use the notation 
\begin{equation}\label{eqdef}\mU\in \mathcal{C}_{\epsilon}\left(\mX,\mY,\mV \right) \iff \left\{\begin{array}{l}\left\|\left(\mU^\top\mX- \mY\right)_\Omega\right\|_F\leq\epsilon\\[.1cm] \left(\mU^\top\mX\right)_{\Omega^c} \leq \mV_{\Omega^c} \end{array}\right., ~\mbox{for}~ \Omega = \supp~\mY.
\end{equation}
Using this notation, the convex program in \eqref{eqconv} may be cast as
\begin{equation}\label{eqconv2}
\hat\mW = \operatorname*{arg\,min}_{\mU}~\left\|\mU\right\|_{1}\quad \mbox{subject to} \quad \mU\in\mathcal{C}_{\epsilon}\left(\mX^{in},\mX^{out},\boldsymbol{0} \right).
\end{equation}

\subsection{Pruning the Network} \label{sec:caspar}

Having access to the tools to retrain any layer within the network, exclusively based on the input and the output, we may consider \emph{parallel} or \emph{cascade} frameworks to retrain the entire network. 

The parallel Net-Trim is a straightforward application of the convex program \eqref{eqconv2} to each layer in the network. Basically, each layer is processed independently based on the initial model input and output, without taking into account the retraining result from the previous layers. Specifically, denoting $\mX^{(\ell-1)}$ and $\mX^{(\ell)}$ as the input and output of the $\ell$-th layer of the initial trained network, we propose to retrain the coefficient matrix $\mW_{\ell}$ via the convex program
\begin{equation}\label{training layer ell}
\hat{\mW}_{\ell}= \operatorname*{arg\,min}_{\mU}~\left\|\mU\right\|_{1} \quad \mbox{subject to} \quad \mU\in \mathcal{C}_{\epsilon_\ell}\left(\mX^{(\ell-1)},\mX^{(\ell)},\boldsymbol{0} \right),\quad \ell = 1,\cdots,L.
\end{equation}
An immediate question would be if each layer of a network is retrained via \eqref{training layer ell} and one replaces $\net(\{\mW_{\ell}\}_{\ell=1}^L;\mX)$ with the retrained network $\net(\{\hat\mW_{\ell}\}_{\ell=1}^L;\mX)$, how do the discrepancies $\epsilon_\ell$ propagate across the network, and how far apart would be the final responses of the two networks to $\mX$? The following result addresses this question.
\begin{theorem}[Parallel Net-Trim]\label{thpar}
Consider a normalized network $\net(\{\mW_{\ell}\}_{\ell=1}^L;\mX)$, such that $\|\mW_\ell\|_1=1$ for $\ell = 1,\cdots, L$.  Solve \eqref{training layer ell} for each layer and form the retrained network $\net(\{\hat\mW_{\ell}\}_{\ell=1}^L;\mX)$. Denoting by $\hat\mX{}^{(\ell)} = \ReLU(\hat \mW_{ \ell}{}^\top\hat\mX{}^{(\ell-1)})$ the outcomes of the retrained network, where $\hat\mX{}^{(0)} = \mX^{(0)} = \mX$, the layer outcomes of the original and retrained networks obey 
\begin{equation}\left\|\hat\mX{}^{(\ell)}-\mX{}^{(\ell)}\right\|_F\leq \sum_{j=1}^\ell \epsilon_j,\quad \ell = 1,\cdots,L.
\end{equation}

\end{theorem}
It is noteworthy that the normalization assumption $\|\mW_\ell\|_1=1$ in Theorem \ref{thpar} is made with no loss in generality, and is only a way of presenting the result in a standard form. This is simply because $\ReLU(|\alpha| x)=|\alpha|\ReLU(x)$, and a scaling of any of the weight matrices $\mW_\ell$ would scale $\mX^{(L)}$ (or $\mX^{(\ell')}$ where $\ell'\geq \ell$) by the same amount. Specifically, the outcomes of the network before and after the process obey
\[\left\|\hat\mX{}^{(L)}-\mX{}^{(L)}\right\|_F\leq \sum_{j=1}^L \epsilon_j,
\]
which makes parallel Net-Trim a stable process, producing  a controllable overall discrepancy.

A more adaptive way of retraining a network, which we would refer to as the cascade Net-Trim, incorporates the outcome of the previously pruned layers to retrain a target layer. Basically, in a cascade Net-Trim, retraining $\mW_\ell$ takes place by exploring a path between the input/output pairs $(\hat\mX{}^{(\ell-1)},\mX_\ell)$ instead of $(\mX{}^{(\ell-1)},\mX_\ell)$. Due to some feasibility concerns, that will be detailed in the sequel, a cascade formulation does not simply happen by replacing $\mX{}^{(\ell-1)}$ with $\hat\mX{}^{(\ell-1)}$ in \eqref{training layer ell}, and the formulation requires some modifications. 

To derive the cascade formulation, consider starting the process by retraining the first layer via
\begin{equation}\label{eqfirst}
\hat{\mW}_{1}= \operatorname*{arg\,min}_{\mU}~\left\|\mU\right\|_{1} \quad \mbox{subject to} \quad \mU\in \mathcal{C}_{\epsilon_1}\left(\mX,\mX^{(1)},\boldsymbol{0} \right).
\end{equation}
Setting $\hat\mX{}^{(1)} = \ReLU(\hat\mW_{ 1}{}^\top\mX)$, to adaptively prune the second layer, one would ideally consider the program 
\begin{equation}\label{eqsecond}
\minimize_{\mU}\;\;\left\|\mU\right\|_{1}\quad \mbox{subject to} \quad \mU\in \mathcal{C}_{\epsilon_2}\left(\hat\mX{}^{(1)},\mX^{(2)},\boldsymbol{0}\right).
\end{equation}
It is not hard to see that the simple generalization in \eqref{eqsecond} is not guaranteed to be feasible, that is, there exists a matrix $\mW$ such that for $\Omega = \supp \mX^{(2)}$:
\begin{equation}\label{eqthird}
\left\{\begin{array}{l}\left\|\left(\mW^\top\hat\mX{}^{(1)}- \mX^{(2)}\right)_\Omega\right\|_F\leq\epsilon_2\\[.1cm] \left(\mW^\top\hat\mX{}^{(1)}\right)_{\Omega^c} \leq \boldsymbol{0} \end{array}\right. .
\end{equation}
If instead of $\hat\mX{}^{(1)}$ the constraint set \eqref{eqthird} was parameterized by $\mX^{(1)}$, a natural feasible point would have been $\mW = \mW_{2}$. Now that $\hat\mX{}^{(1)}$ is a perturbed version of $\mX^{(1)}$, the constraint set needs to be properly slacked to maintain the feasibility of $\mW_{2}$. In this context, one may easily verify that $\mW_{2}$ is feasible for the slacked program
\begin{equation}\label{eqcass2}
\minimize_{\mU}~\left\|\mU\right\|_{1}\quad \mbox{subject to} \quad \mU\in \mathcal{C}_{\epsilon_2}\left(\hat\mX{}^{(1)},\mX^{(2)},\mW_{2}^\top\hat\mX{}^{(1)}\right),
\end{equation}
as long as for some $\gamma\geq 1$,
\[\epsilon_2 = \gamma \left\|\left(\mW_2^\top\hat\mX{}^{(1)}- \mX^{(2)}\right)_\Omega\right\|_F.
\]
The $\gamma$-coefficient is a free parameter, which we refer to as the \emph{inflation rate}. When $\gamma=1$, the matrix $\mW_{2}$ is only tightly feasible for \eqref{eqcass2} and the feasible set can at the very least become a singleton. However, increasing the inflation rate would expand the set of permissible matrices and makes \eqref{eqcass2} capable of producing sparser solutions. 

The process applied to the second layer may be generalized to the subsequent layers and form a cascade paradigm to prune the network layer by layer. The pseudocode in Algorithm 1 summarizes the Net-Trim cascade scheme, where we set $\epsilon_1 = \epsilon$ for the first layer, and consider the inflation rates $\gamma_\ell$, $\ell = 2,\cdots,L$, for the subsequent layers.

\begin{algorithm}
\centerline{\caption{Cascade Net-Trim}}
\begin{algorithmic}
\STATE{$\hat{\mW}_{1}\gets \operatorname*{arg\,min}_{\mU}~\left\|\mU\right\|_{1} \quad \mbox{subject to} \quad \mU\in \mathcal{C}_{\epsilon}\left(\mX,\mX^{(1)},\boldsymbol{0} \right)$}

\STATE{$\hat\mX{}^{(1)} \gets \ReLU\left (\hat\mW_{ 1}{}^\top\mX\right )$}

\FOR{$\ell=2,\cdots,L$}
\STATE{
    $\Omega \gets \supp \mX^{(\ell)}$;
     $~~\epsilon_\ell \gets \gamma_\ell \left\|\left(\mW_\ell ^\top\hat\mX{}^{(\ell-1)}- \mX^{(\ell)}\right)_\Omega\right\|_F$}

\STATE{$\hat{\mW}_{\ell}\gets \operatorname*{arg\,min}_{\mU}~\left\|\mU\right\|_{1}\quad \mbox{subject to} \quad \mU\in \mathcal{C}_{\epsilon_\ell}\left(\hat\mX{}^{(\ell-1)},\mX^{(\ell)},\mW_{\ell}^\top\hat\mX{}^{(\ell-1)}\right)$}

 \STATE{$\hat\mX{}^{(\ell)} \gets \ReLU\left (\hat\mW_{ \ell}{}^\top\hat\mX{}^{(\ell-1)}\right)$  } 
\ENDFOR
    \end{algorithmic}
\end{algorithm}

Similar to the parallel scheme, we can show a bounded discrepancy between the outcomes of the initial network $\net(\{\mW_{\ell}\}_{\ell=1}^L;\mX)$ and the retrained network $\net(\{\hat\mW_{\ell}\}_{\ell=1}^L;\mX)$, as follows. 
 \begin{theorem}[Cascade Net-Trim]\label{thcas}
Consider a normalized network $\net(\{\mW_{\ell}\}_{\ell=1}^L;\mX)$, such that $\|\mW_\ell\|_1=1$ for $\ell = 1,\cdots, L$.  If the network is retrained according to Algorithm 1, the layer outcomes of the original and retrained networks will obey 
\begin{equation}\label{eqcasthstat}
\left \|\hat\mX{}^{(\ell)}-\mX{}^{(\ell)}\right \|_F\leq \epsilon {\prod_{j=2}^\ell \gamma_j}.\end{equation}
\end{theorem}
Specifically, when an identical inflation rate is used across all the layers, one would have $\|\hat\mX{}^{(L)}-\mX^{(L)}\|_F\leq \gamma^{{(L-1)}}\epsilon$, which is a controllably small quantity, given that $\gamma$ can be selected arbitrarily close to 1. For instance when $\gamma=1.01$ and $L=10$, the total network discrepancy would be still less than $1.1\epsilon$. As will be demonstrated in the experiments section, for the same level of total network discrepancy, the cascade Net-Trim is capable of producing sparser networks. However, such reduction is achieved  at the expense of the loss in distributability, which makes the parallel scheme computationally more attractive for big data problems.

\section{Sample Complexity Bounds Using Subgaussian Random Flow}\label{sampComplexSec}
In the previous section we discussed and analyzed the convex retraining scheme and its consistency with the reference model. In this section we analyze the sample complexity of the proposed retraining framework. Basically, the goal of this section is addressing the following question: \textit{if there exists a sparse transformation matrix relating the input and output of a layer, how many random samples are sufficient to recover it via the proposed retraining scheme?}

As will be detailed in the sequel, we will show that retraining each neuron within the network is possible with fewer samples than the neuron degrees of freedom. More specifically, for a trained neuron with $N$ input ports, if generating an identical response is possible with $s\ll N$ nonzero weights, Net-Trim is able to recover such model with only $\mathcal{O}(s\log(N/s))$ random samples. This result is valid for the neurons of any layer within the network, as long as some standard statistical properties can be established for the input samples. 
%
%

Unlike the previous work \cite{Aghasi2017NetTrim}, which establishes a similar result for only the neurons within the first layer, here, due to some favorable tail properties of subgaussian random vectors, we are able to generalize the result to the entire network. Basically, we will show that when the network input samples are independently drawn from a standard normal (or any other subgaussian) distribution, the input samples at all subsequent layers remain independent and subgaussian (what we refer to as a subgaussian flow). By carefully using some technical tools from the structured signal recovery literature \cite{tropp2015convex, mendelson2014learning}, we are able to present the main sample complexity result in a general form. 

To present the results, we first start with a brief overview of subgaussian random variables. For a more comprehensive overview, the reader is referred to \cite{vershynin_2012} and \S 2.2 of \cite{van1996weak}. 
\begin{definition}[subgaussian random variable]
A random variable $\varphi$ is subgaussian\footnote{In general, the right-hand expression in \eqref{subg1} can be replaced with $c\exp\left( - {t^2}/{\kappa^2}\right)$ using two absolute constants $c$ and $\kappa$} if there exists a constant $\kappa$, such that for all $t\geq 0$,
\begin{equation}\label{subg1}
\mathbb{P}\left\{ |\varphi| > t\right\} \leq \exp\left(1 - \frac{t^2}{\kappa^2}\right).
\end{equation}
Equivalently, $\varphi$ is subgaussian if there exists a constant $\hat\kappa$ such that  
\begin{equation}\label{subg2}
\mathbb{E}\exp\left( \frac{\varphi^2}{{\hat\kappa}^2}\right) \leq e.
\end{equation}
\end{definition}
The subgaussian norm of $\varphi$, also referred to as the Orlicz norm, is denoted by $\|\varphi\|_{\psi_2}$, and defined as
\[\|\varphi\|_{\psi_2} \triangleq \sup_{p\geq 1} p^{-\frac{1}{2}}\left ( \mathbb{E}|\varphi|^p\right)^{\frac{1}{p}}.
\]
While calculating the exact Orlicz norm can be challenging, if either one of the properties \eqref{subg1} or \eqref{subg2} hold, $\|\varphi\|_{\psi_2}$ is the smallest possible number ($\kappa$ or $\hat\kappa$) in either one of these inequalities, up to an absolute constant.
\begin{definition}[subgaussian random vector]
A random vector $\bphi\in\mathbb{R}^N$ is subgaussian if for all $\balpha\in \mathbb{R}^N$ (or equivalently all $\balpha\in \mathbb{S}^{N-1}$), the one-dimensional marginals $\balpha^\top\bphi$ are subgaussian. 
\end{definition}
The notion of Orlicz norm also generalizes to the vector case as
\begin{equation}\label{b7}
\|\bphi\|_{\psi_2} \triangleq \sup_{\balpha\in\mathbb{S}^{N-1}}\|\balpha^\top\bphi\|_{\psi_2}=\sup_{\balpha\in\mathbb{S}^{N-1}}\sup_{p\geq 1} p^{-\frac{1}{2}}\left ( \mathbb{E}\left|\balpha^\top\bphi \right|^p\right)^{\frac{1}{p}}.
\end{equation}

\begin{figure}[t]
\centering \begin{overpic}[trim={0 -.25cm  0 0},clip,width=6.1in,height=1.75in]{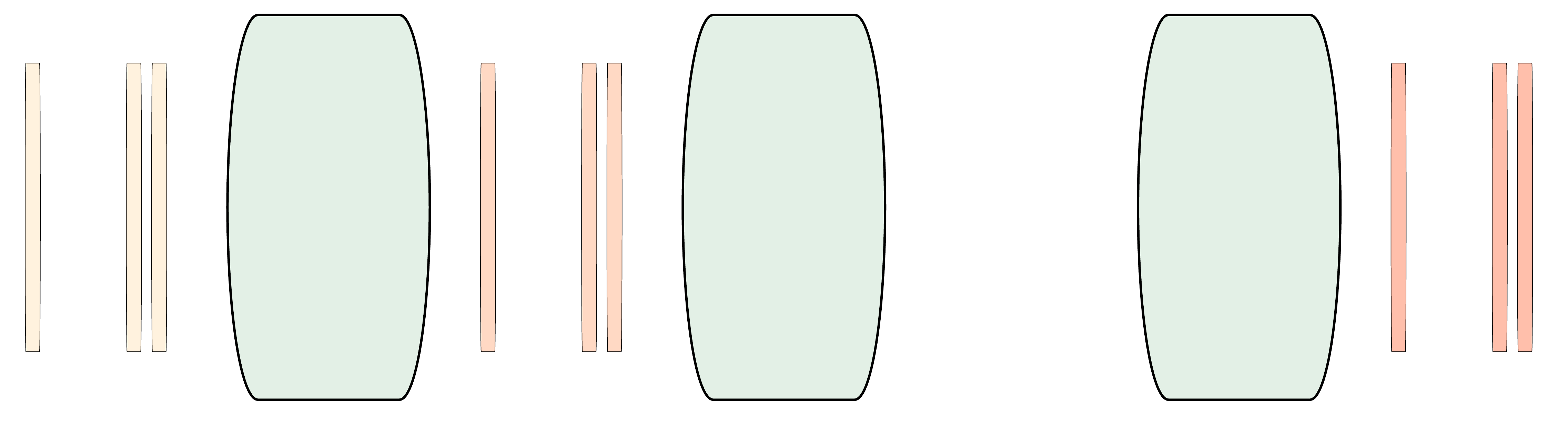}
\put (1,26) {\scalebox{.8}{$\vx_P$}} 
\put (7,26) {\scalebox{.8}{$\vx_2$}} 
\put (9.8,26) {\scalebox{.8}{$\vx_1$}} 
\put (4.3,14.5) {\scalebox{.8}{$\cdots$}}
\put (30,26) {\scalebox{.8}{$\vx^{(1)}_P$}} 
\put (35.5,26) {\scalebox{.8}{$\vx^{(1)}_2$}} 
\put (38.8,26) {\scalebox{.8}{$\vx^{(1)}_1$}} 
\put (33.3,14.5) {\scalebox{.8}{$\cdots$}}
\put (88,26) {\scalebox{.8}{$\vx^{(L)}_P$}} 
\put (93.5,26) {\scalebox{.8}{$\vx^{(L)}_2$}} 
\put (96.8,26) {\scalebox{.8}{$\vx^{(L)}_1$}} 
\put (91.3,14.5) {\scalebox{.8}{$\cdots$}}
\put (11.5,15.8) {\scalebox{1}{$\rightarrow$}} 
\put (11.5,14.5) {\scalebox{1}{$\rightarrow$}} 
\put (11.5,13.2) {\scalebox{1}{$\rightarrow$}}  
\put (28,15.8) {\scalebox{1}{$\rightarrow$}} 
\put (28,14.5) {\scalebox{1}{$\rightarrow$}} 
\put (28,13.2) {\scalebox{1}{$\rightarrow$}} 
\put (40.5,15.8) {\scalebox{1}{$\rightarrow$}} 
\put (40.5,14.5) {\scalebox{1}{$\rightarrow$}} 
\put (40.5,13.2) {\scalebox{1}{$\rightarrow$}} 
\put (62.5,14.5) {\scalebox{1.4}{$\cdots$}}
\put (86,15.8) {\scalebox{1}{$\rightarrow$}} 
\put (86,14.5) {\scalebox{1}{$\rightarrow$}} 
\put (86,13.2) {\scalebox{1}{$\rightarrow$}} 
%
%
\put (19,8) {\scalebox{.8}{\rotatebox{90}{$\left(\mW_{\!\! 1}^\top\vx+\vb^{(1)}\right)^+$}}} 
\put (48,7) {\scalebox{.8}{\rotatebox{90}{$\left(\mW_{\!\! 2}^\top\vx^{(1)}+\vb^{(2)}\right)^+$}}} 
\put (74.5,4.5) {\scalebox{.8}{\rotatebox{90}{$\left(\mW_{\!\! L}^\top\vx^{(L-1)}+\vb^{(L-1)}\right)^+$}}} 
\put (79.5,14) {\scalebox{.8}{\rotatebox{90}{or}}}
\put (82,6) {\scalebox{.8}{\rotatebox{90}{$\mW_{\!\! L}^\top\vx^{(L-1)}+\vb^{(L-1)}$}}} 
\put (0,5) {\scalebox{1}{$\underbrace{{\white ----- }}_{i.i.d ~ SG}$}} 
\put (29,5) {\scalebox{1}{$\underbrace{{\white ----- }}_{i.i.d ~ SG}$}} 
\put (87,5) {\scalebox{1}{$\underbrace{{\white ----- }}_{i.i.d ~ SG}$}} 
\end{overpic}
 \caption{When $\vx_1,\vx_2,\cdots,\vx_P$, the input samples to the proposed neural network,  are independently drawn from a subgaussian distribution, the input/output vectors of every subsequent layer remain i.i.d and subgaussian}\label{figSG}\vspace{-.5cm}
\end{figure}
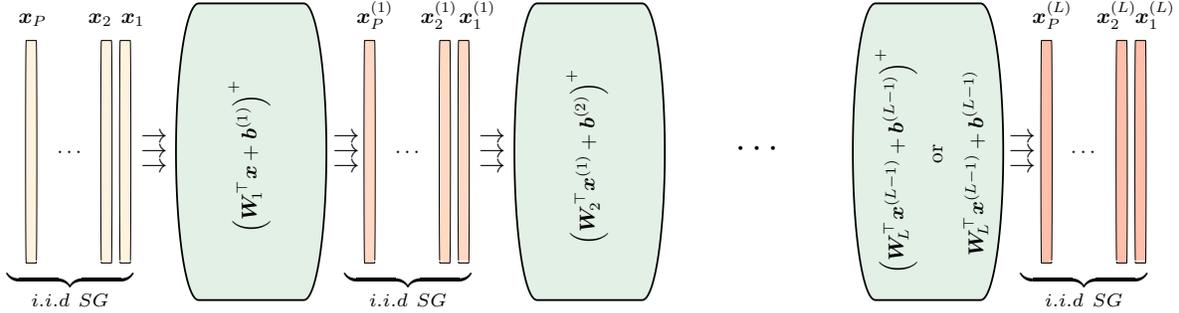

We are now ready to state the first result, which warrants a subgaussian random flow across the network, as long as the network input samples are independently drawn from a standard Gaussian (or subgaussian) distribution. 
\begin{theorem}\label{subGth}
Consider a network with fixed parameters $\mW_{\!\! \ell}$, $\vb^{(\ell)}$, where the input and output to each layer are related via 
\begin{equation}\label{eqrecvec}
\vx^{(\ell)} = \ReLU\left(\mW_{\!\! \ell}^\top\vx^{(\ell-1)}+\vb^{(\ell)}\right),\qquad \ell=1,\cdots,L.
\end{equation}
If the network is fed with i.i.d sample vectors $\vx_1^{(0)},\cdots,\vx_P^{(0)}\sim\mathcal{N}\left(\boldsymbol{0},\boldsymbol{I}\right)$, the response samples at each layer output remain i.i.d subgaussian. 
\end{theorem}
As shown in the proof, the result of Theorem \ref{subGth} still holds when the network input samples are independently drawn from a subgaussian distribution instead of a standard normal, and/or when the last layer skips a ReLU activation. Specifically, when the network is fed with $\vx_1^{(0)},\vx_2^{(0)},\cdots,\vx_P^{(0)}$, independently drawn from a subgaussian distribution, the resulting responses $\vx_1^{(\ell)},\vx_2^{(\ell)},\cdots,\vx_P^{(\ell)}$ at any layer $\ell$ remain independent and subgaussian. The diagram in Figure \ref{figSG} demonstrates such statistical structure among the layer inputs across the network. 

Having independent subgaussian samples at any layer input port allows us to relate the number of samples to the recovery of a reduced model. Exchanging the layer index with the general input/output notation, when $\mX^{in}\in\mathbb{R}^{N\times P}$ and $\mX^{out}\in\mathbb{R}^{M\times P}$ are respectively the input and output to a layer, related via $\mX^{out} = \ReLU(\mW^\top\mX^{in})$, obtaining the pruned layer matrix $\hat\mW\in\mathbb{R}^{N\times M}$ is performed via 
\begin{equation}\label{retrainALayer}
\hat\mW = \operatorname*{arg\,min}_{\mW}~\left\|\mW\right\|_{1} \quad \mbox{subject to} \quad \mW\in \mathcal{C}_{\epsilon}\left(\mX^{in},\mX^{out},\boldsymbol{0} \right).
\end{equation}
When $\epsilon=0$, the program in \eqref{retrainALayer} decouples into $M$ individual convex programs each retraining a column in $\mW$. Basically, instead of solving \eqref{retrainALayer} for $\hat\mW$, if $\vx{{}^{out}}^\top\in\mathbb{R}^P$ is a row in $\mX^{out}$, the corresponding column in $\hat\mW$ can be calculated via
\begin{equation}\label{retrainANeuron}
\minimize_{\vw}~~\left\|\vw\right\|_{1} \quad \mbox{subject to} \quad \vw\in \mathcal{C}_{0}\left(\mX^{in},{\vx^{out}}^\top,\boldsymbol{0} \right),
\end{equation}
reducing \eqref{retrainALayer} to retraining each of the $M$ output neurons, individually. Our focus on the  case of $\epsilon=0$ (which also makes the cascade and parallel schemes equivalent) is working in an underdetermined regime, where the required samples are shown to be much less than the layer (neuron) degrees of freedom. In this case, the relationship between $\mX^{in}$ and $\mX^{out}$ can be established via infinitely many $\mW$ matrices and one seeks a unique sparse solution via \eqref{retrainALayer}.

Before stating the main technical result, we would like to introduce some notions used in the presentation. When a neuron is initially trained via a vector $\vw_0\in\mathbb{R}^N$ and fed with i.i.d instances of $\vx$, the activation pattern of the neuron is fully controlled by the sign of $\vw_0^\top\vx$. In this case, one expects to gain the main retraining information from the cases when ReLU is in the linear mode (i.e., $\vw_0^\top\vx>0$). In this regard, corresponding to the random input $\vx$, we define the random \emph{virtual input} as
\[\boldsymbol{\upsilon} = \vx 1_{\vw_0^\top\vx>0}=\left\{ \begin{array}{lc} \vx & \vw_0^\top\vx >0\\[.1cm] \boldsymbol{0} & \vw_0^\top\vx\leq 0 \end{array}\right. .
\]
The virtual random vector plays a key role in our presentation. Our presentation also depends on the smallest eigenvalue of the virtual covariance matrix, which follows the standard definition:
\[\lambda_{\min}\left(\operatorname{cov}\left(\boldsymbol{\upsilon}\right)\right) = \inf_{\balpha\in\mathbb{S}^{N-1}}\balpha^\top \operatorname{cov}\left(\boldsymbol{\upsilon}\right)\balpha, \quad \mbox{where}\quad \operatorname{cov}\left(\boldsymbol{\upsilon}\right)=\mathbb{E}\left(\boldsymbol{\upsilon}\boldsymbol{\upsilon}^\top\right) -\mathbb{E}\left(\boldsymbol{\upsilon}\right)\mathbb{E}\left(\boldsymbol{\upsilon}\right)^\top.
\]
\begin{theorem}\label{sampComp}  
For the model \eqref{eqrec}, consider a trained neuron obeying $\vx^{out} = \mathtt{ReLU}({\mX^{in}}^\top\!\vw_0)$, where $\mX^{in} = [\vx_1,\cdots,\vx_P]\in\mathbb{R}^{N\times P}$ and $\vx,\;\! \vx_1,\cdots,\vx_P$ are independent samples of a subgaussian distribution. Assume, an $s$-sparse vector $\vw^*\in\mathbb{R}^N$ is capable of generating an identical response to $\mX^{in}$ as $\vx^{out}$. Fix $\beta\geq 1/2$ and $t\geq 0$, then if \vspace{-.2cm}
\begin{equation}\label{PsampleCompRate}
P \gtrsim C_{\beta,\upsilon} \left(s\log\left( \frac{N}{s}\right)+s+1+t \right),\vspace{-.2cm}
\end{equation}
retraining the neuron via \eqref{retrainANeuron} recovers $\vw^*$ with probability exceeding $1-e^{-ct}$. The absolute constant $c$ is universal and the constant $C_{\beta,\upsilon}$ depends on the statistics of the virtual input $\boldsymbol{\upsilon} = \vx 1_{\vw_0^\top\vx>0}$ via \vspace{-.2cm}
\begin{equation}\label{eqSampCons}
C_{\beta,\upsilon} = \left(1+\beta\right)^2\left(\frac{{\|\boldsymbol{\upsilon}- \mathbb{E} \boldsymbol{\upsilon}\|_{\psi_2}^2}}{\lambda_{\min}(\operatorname{cov}(\boldsymbol{\upsilon}))}\right)^{3+\frac{1}{\beta}}.
\end{equation}
\end{theorem}
%

We would like to highlight some technical details related to Theorem \ref{sampComp}. To establish the result we use the \emph{bowling scheme} proposed by \cite{tropp2015convex}, which discusses the recovery of a structured (e.g., sparse) signal from independent linear measurements. Below, we make a connection between our problem with nonlinear constraints to the problem with linear constraints described there. While we used the compact model \eqref{eqrec} for a more concise presentation, the model in \eqref{eqrec0} is still covered by Theorem \ref{sampComp}, treating the intercept as a constant feature appended to the neuron input.

It is important to note that due to the application of the ReLU at each layer, the random samples entering the next layer are non-centered and this requires a careful analysis of the problem. In fact, the majority of the measurement systems in the structured recovery literature work with centered random measurements, as some of the powerful analysis tools, such as the restricted isometry property \cite{candes2008restricted, candes2006stable}, the certificate of duality \cite{candes2011probabilistic, gross2011recovering}, and the Mendelson's small ball method -- which stands as the backbone for the bowling scheme \cite{koltchinskii2015bounding, mendelson2014learning, Mendelson2017} rely critically on the random vectors being centered. In the presentation of Theorem \ref{sampComp}, the constant is related to the statistics of the centered virtual input, regardless of the mean shift that the previous activation units have caused to the input\footnote{This is important because the Orlicz norm of a noncentered random vector can easily become dimension-dependent. For instance, if the components of $\vx\in\mathbb{R}^N$ are i.i.d standard Gaussians, one can easily verify that $\left\| \vx^+ \right\|_{\psi_2}=\mathcal{O}(\sqrt{N})$, while $\left\| \vx^+ - \mathbb{E}\vx^+ \right\|_{\psi_2}=\mathcal{O}(1)$.}.

Finally, Theorem \ref{sampComp} can be used as a general and powerful tool to estimate the retraining sample complexity for any layer within the network. To establish the $\mathcal{O}(s\log(N/s))$ rate for a given layer, we only need to show that for the corresponding input $\vx$ and initially trained weights $\vw_0$, the virtual input $\boldsymbol{\upsilon} = \vx 1_{\vw_0^\top\vx>0}$ satisfies the following two conditions:
\begin{equation}\label{sampleconds}
\lambda_{\min}\left(\operatorname{cov}\left(\boldsymbol{\upsilon}\right)\right)\gtrsim 1, \qquad \mbox{and}\qquad \left\| \boldsymbol{\upsilon}- \mathbb{E} \boldsymbol{\upsilon} \right\|_{\psi_2}\lesssim 1.
\end{equation}
As an insightful example, we go through the exercise of establishing the bounds in \eqref{sampleconds} for a layer fed with i.i.d Gaussian samples; this is not an unreasonable scenario for the first layer of a neural network. As will be detailed in Section \ref{sec:probiid} below, using standard tools to verify the conditions in \eqref{sampleconds}, conveniently proves the $\mathcal{O}(s\log(N/s))$ rate for such layer.

For a network fed with i.i.d Gaussian samples, going through a similar exercise for the subsequent layers (say layer $\ell>1$, with independent copies of the random input $\vx^{(\ell)}$), requires tracing the statistics of $\boldsymbol{\upsilon}^{(\ell)} = \vx^{(\ell)} 1_{\vw_0^\top\vx^{(\ell)}>0}$ down to the Gaussian input $\vx^{(0)}$. In such case, warranting the conditions in \eqref{sampleconds} would require stating realistic conditions on the initially trained $\mW_{\!\! j}$ for $j=1,\cdots,\ell$. Such generalization could be application specific and beyond the current load of the paper, which is left as a potential future work.

\subsection{Feeding a Neuron with i.i.d Gaussian Samples}\label{sec:probiid}
In this section we go through the exercise of establishing the conditions in \eqref{sampleconds} for a neuron fed with independent copies of $\vx \sim \mathcal{N}(\boldsymbol{0},\mI)$, $\vx\in\mathbb{R}^N$. Below, we go through each bound in \eqref{sampleconds}, separately. In all the calculations, $\vw_0\neq \boldsymbol{0}$ is a fixed vector that corresponds to the initially trained model. In \cite{Aghasi2017NetTrim}, the authors go through a chain of techniques to prove an $\mathcal{O}(s\log N)$ sample complexity by carefully constructing a dual certificate for the convex program. Here we will see that thanks to Theorem \ref{sampComp}, such process is markedly reduced to establishing the conditions in \eqref{sampleconds}, which is conveniently fulfilled using standard tools.

\subsubsection{Step 1: Bounding the Covariance Matrix}
To evaluate the virtual input covariance matrix we have 
\begin{align}\nonumber 
\lambda_{\min}\left(\operatorname{cov}\left(\vx 1_{\vw_0^\top\vx>0}\right)\right) &= \lambda_{\min}\left( \mathbb{E} \vx\vx^\top 1_{\vw_0^\top\vx>0} - \left(\mathbb{E} \vx 1_{\vw_0^\top\vx>0}\right)\left(\mathbb{E} \vx 1_{\vw_0^\top\vx>0}\right)^\top  \right) \\ \nonumber &\geq  \lambda_{\min}\left( \mathbb{E} \vx\vx^\top 1_{\vw_0^\top\vx>0}\right) +\lambda_{\min}\left(- \left(\mathbb{E} \vx 1_{\vw_0^\top\vx>0}\right)\left(\mathbb{E} \vx 1_{\vw_0^\top\vx>0}\right)^\top  \right)\\ & = \lambda_{\min}\left( \mathbb{E} \vx\vx^\top 1_{\vw_0^\top\vx>0}\right) -\left\| \mathbb{E} \vx 1_{\vw_0^\top\vx>0}  \right\|^2, \label{covvirtual}
\end{align}
where the second line follows from Weyl's inequality. To conveniently calculate the required moments, we can make use of the following lemma, which reduces the calculations to the bivariate case.  
\begin{lemma}\label{LemInt}
Consider $\vx=(x_1,\cdots,x_N)^\top\sim \mathcal{N}(\boldsymbol{0},\mI)$ and let $g(.):\mathbb{R}\to\mathbb{R}$ be a real-valued function. Then, for any fixed vectors $\balpha,\bbeta\in\mathbb{S}^{N-1}$:
\begin{equation}
\mathbb{E}_{\vx}~ g\left(\balpha^\top\vx\right)1_{\bbeta^\top\vx>0}=\mathbb{E}_{x_1,x_2}~  g\left(  \left(\balpha^\top\bbeta\right) x_1 +  \sqrt{1- \left(\balpha^\top\bbeta\right)^2} x_2\right)1_{x_1>0}.\label{bivarred}
\end{equation} 
\end{lemma}
With no loss of generality we can assume $\vw_0\in\mathbb{S}^{N-1}$, and apply \eqref{bivarred} to the first right-hand side term in \eqref{covvirtual} to get
\begin{align*}
\lambda_{\min}\left( \mathbb{E} \vx\vx^\top 1_{\vw_0^\top\vx>0}\right) &= \inf_{\balpha\in\mathbb{S}^{N-1}}\mathbb{E} \left(\balpha^\top\vx\right)^2 1_{\vw_0^\top\vx>0}\\& = \inf_{\balpha\in\mathbb{S}^{N-1}}\mathbb{E}_{x_1,x_2}\left(  \left(\balpha^\top\vw_0\right) x_1 +  \sqrt{1- \left(\balpha^\top\vw_0\right)^2} x_2\right)^2 1_{x_1>0}\\ & = \inf_{\balpha\in\mathbb{S}^{N-1}} \frac{1}{2} \left(\balpha^\top\vw_0\right)^2 + \frac{1}{2}\left( 1 - \left(\balpha^\top\vw_0\right)^2\right) \\ &=\frac{1}{2}
\end{align*}
For the second term in \eqref{covvirtual} we have
\begin{align*}
\left\| \mathbb{E} \vx 1_{\vw_0^\top\vx>0}  \right\| &= \sup_{\balpha\in\mathbb{S}^{N-1}} \mathbb{E} \left(\balpha^\top\vx\right) 1_{\vw_0^\top\vx>0}\\ & = \sup_{\balpha\in\mathbb{S}^{N-1}}\mathbb{E}_{x_1,x_2}\left(  \left(\balpha^\top\vw_0\right) x_1 +  \sqrt{1- \left(\balpha^\top\vw_0\right)^2} x_2\right) 1_{x_1>0}\\ & = \sup_{\balpha\in\mathbb{S}^{N-1}} \frac{1}{\sqrt{2\pi}}\left(\balpha^\top\vw_0\right)\\ &= \frac{1}{\sqrt{2\pi}},
\end{align*}
as a result of which one has $\lambda_{\min}\left(\operatorname{cov}\left(\vx 1_{\vw_0^\top\vx>0}\right)\right)\geq 1/2-1/(2\pi)$.

\subsubsection{Step 2: Bounding the Orlicz Norm}
To bound the Orlicz norm of the centered virtual input by a constant, we only need to introduce a constant $\kappa$ such that for all $\balpha\in\mathbb{S}^{N-1}$  the marginals $\balpha^\top ( \vx 1_{\vw_0^\top\vx>0}- \mathbb{E} \vx 1_{\vw_0^\top\vx>0})$ obey \eqref{subg1}. To this end, one has
\begin{align*}
\forall \balpha\in\mathbb{S}^{N-1}: \left| \balpha^\top \left( \vx 1_{\vw_0^\top\vx>0}- \mathbb{E} \vx 1_{\vw_0^\top\vx>0}\right) \right| &\leq \left| \balpha^\top  \vx 1_{\vw_0^\top\vx>0}\right| + \left\|\mathbb{E} \vx 1_{\vw_0^\top\vx>0}\right\|\\ &\leq \left| \balpha^\top  \vx \right| + \frac{1}{\sqrt{2\pi}}.
\end{align*}
As a result, for $\vx\sim \mathcal{N}(\boldsymbol{0},\mI)$ and any fixed $\balpha\in\mathbb{S}^{N-1}$:
\begin{align*}
\forall t\geq 0: ~~\mathbb{P}\left\{ \left| \balpha^\top \left( \vx 1_{\vw_0^\top\vx>0}- \mathbb{E} \vx 1_{\vw_0^\top\vx>0}\right) \right| > t\right\} &\leq \mathbb{P}\left\{ \left| \balpha^\top  \vx \right| + \frac{1}{\sqrt{2\pi}} > t \right\}\\ & = \mathbb{P}\left\{ \left| \balpha^\top  \vx \right| > \max\left(t- \frac{1}{\sqrt{2\pi}},0\right) \right\}
\\ &\leq \exp \left( -\frac{1}{2}\max\left(t- \frac{1}{\sqrt{2\pi}},0\right)^2 \right),
\end{align*}
where in the last inequality we used the fact that $\balpha^\top\vx\sim\mathcal{N}(0,1)$ and for a standard normal variable $z$, $\mathbb{P}\{ |z|\geq t\}\leq \exp(-t^2/2)$ for all $t\geq 0$. Finally we can use the basic inequality stated in Lemma \ref{lemx1} of the proofs section to get
\[
\forall t\geq 0: ~~\mathbb{P}\left\{ \left| \balpha^\top \left( \vx 1_{\vw_0^\top\vx>0}- \mathbb{E} \vx 1_{\vw_0^\top\vx>0}\right) \right| > t\right\} \leq \exp \left( 1 -\frac{t^2}{2+\frac{1}{2\pi}} \right),
\]
which implies that $\left\| \vx 1_{\vw_0^\top\vx>0}- \mathbb{E} \vx 1_{\vw_0^\top\vx>0} \right\|_{\psi_2}\lesssim 1$.


\section{Net-Trim Implementation}
\label{app:ADMM}
In this section we discuss details of an ADMM implementation for the Net-Trim convex program. The approach that we suggest here is based on the \emph{global variable consensus} (see \S7.1 of \cite{boyd2011distributed}). This technique is useful in addressing convex optimizations with additively separable objectives. 


For $\mW \in \mathbb{R}^{N\times M}$, $\mX^{in}\in \mathbb{R}^{N\times P}$, and $\Omega\subseteq \{1,\cdots,M\}\times\{1,\cdots,P\}$ the Net-Trim central program  
\begin{equation}
\minimize_{\mW}~\left\|\mW\right\|_1\quad \mbox{subject to} \quad  \left\{\begin{array}{l}\left\|\left(\mW^\top\mX{}^{in} - \mX{}^{out} \right)_\Omega\right \|_F\leq \epsilon  \\[.1cm] \left(\mW^\top\mX{}^{in}\right )_{\Omega^c} \leq \mV_{\Omega^c} \end{array}\right. ,
\end{equation}
can be cast as the equivalent form
\begin{equation}\label{equivNetTrim}
\minimize_{\substack{\mW^{(1)}\in\mathbb{R}^{M\times P} \\ \mW^{(2)}, \mW^{(3)} \in\mathbb{R}^{N\times M }}}  f_1\left(\mW^{(1)}\right) + f_2\left(\mW^{(2)}\right) ~~ \mbox{subject to} ~~  \left\{\!\begin{array}{l}{\mW^{(1)}} = {\mW^{(3)}}^\top \mX^{in}  \\ \mW^{(2)} = \mW^{(3)}\end{array}\right. ,
\end{equation}
where
\[f_1\left(\mW\right) = \mathcal{I}_{\left\| \mW_\Omega - \mX_\Omega^{out}\right\|_F\leq \epsilon}\left(\mW\right)+ \mathcal{I}_{ \mW_{\Omega^c} \leq \mV_{\Omega^c}}\left(\mW\right), ~\mbox{and}~~ f_2\left(\mW\right) = \left\| \mW\right\|_1.
\]
Here $\mathcal{I}_C(\cdot)$ represents the indicator function of the set $C$,
\[\mathcal{I}_C(\mW) = \left\{\begin{array}{cc}0&\mW\in C\\ +\infty& \mW\notin C\end{array}\right..
\]
For the convex program \eqref{equivNetTrim}, the ADMM update for each variable at the $k$-th iteration follows the standard forms
\begin{align}\label{admmprog1}
\mW^{(1)}_{k+1} &= \argmin_{\mW}~ f_1\left(\mW\right) + \frac{\rho}{2}\left\|\mW + \mU^{(1)}_k - {\mW^{(3)}_{k}}^\top\mX^{in} \right\|^2_F,\\ \label{admmprog2} \mW^{(2)}_{k+1} &= \argmin_{\mW}~ f_2\left(\mW\right) + \frac{\rho}{2}\left\|\mW + \mU^{(2)}_k -  \mW^{(3)}_k \right\|^2_F,\\ \label{admmprog3} \mW^{(3)}_{k+1} &= \left(\mX^{in}\mX^{in}{}^\top + \mI\right)^{-1}\left(\mX^{in}\left( {\mW^{(1)}_{k+1}}+{\mU^{(1)}_k}\right)^\top + \mW^{(2)}_{k+1}+\mU^{(2)}_k\right),
\end{align}
and the dual updates are performed via
\begin{align*}
\mU^{(1)}_{k+1}= \mU^{(1)}_{k} + \mW^{(1)}_{k+1} - {\mW^{(3)}_{k+1}}^\top\mX^{in} ,~~~ \mU^{(2)}_{k+1} = \mU^{(2)}_{k} + \mW^{(2)}_{k+1} - \mW^{(3)}_{k+1}.
\end{align*}
The update stated in \eqref{admmprog3} is derived by finding the minimizer of the augmented Lagrangian with respect to $\mW^{(3)}$, which amounts to the minimization
\[\minimize_{\mW}~\frac{\rho}{2}\left\|\mW^{(1)}_{k+1} + \mU^{(1)}_k - \mW^\top\mX^{in} \right\|^2_F + \frac{\rho}{2}\left\|\mW^{(2)}_{k+1} + \mU^{(2)}_k -  \mW \right\|^2_F.
\] 
While the updates for $\mW^{(1)}$ and $\mW^{(2)}$, as in \eqref{admmprog1} and \eqref{admmprog2}, are stated in the general form, they can be further simplified and presented in closed form. To this end, a first observation is that \eqref{admmprog1} can be decoupled into independent minimizations in terms of $\mW_\Omega$ and $\mW_{\Omega^c}$, i.e., 
\begin{align}\nonumber\hspace{-.25cm} \mW^{(1)}_{k+1} = \argmin_{ \mW_\Omega:~\! \left\| \mW_\Omega - \mX^{out}_\Omega\right\|_F\leq \epsilon} &\frac{\rho}{2}\left\|\mW_\Omega + \left( \mU^{(1)}_k - {\mW^{(3)}_{k}}^\top\mX^{in}\right)_\Omega  \right\|_F^2 \\ &+ \argmin_{ \mW_{\Omega^c}:~\! \mW_{\Omega^c} \leq \mV_{\Omega^c} }~ \frac{\rho}{2}\left\|\mW_{\Omega^c} + \left( \mU^{(1)}_k - {\mW^{(3)}_{k}}^\top\mX^{in}\right)_{\Omega^c}  \right\|_F^2. \label{closestBall}
\end{align}
The first minimization on the right-hand side of \eqref{closestBall} is basically the problem of finding the closest point of an $\epsilon$-radius Euclidean ball to a given point. For the non-trivial case that the given point is outside the ball, the solution is the intersection of the ball surface with the line connecting the point to the center of the ball. More specifically, for fixed $\mY$ and $\mZ$,  
\[ \argmin_{ \mW_\Omega:~ \left\| \mW_\Omega - \mZ_\Omega\right\|_F\leq \epsilon}~ \frac{\rho}{2}\left\|\mW_\Omega - \mY_\Omega  \right\|_F^2  = \left\{ \begin{array}{cc} \mY_\Omega & \mbox{if} ~ \left\|  \mY_\Omega -  \mZ_\Omega \right\|_F\leq \epsilon\\[.05cm] \mZ_\Omega + \epsilon\frac{ \mY_\Omega -  \mZ_\Omega}{\left\|  \mY_\Omega -  \mZ_\Omega \right\|_F} & \mbox{else} \end{array}\right..
\]
The second term in \eqref{closestBall} is an instance of a projection onto an orthant and can be delivered in closed form as 
\[ \argmin_{ \mW_{\Omega^c}:~\mW_{\Omega^c} \leq \mV_{\Omega^c} }~ \frac{\rho}{2}\left\|\mW_{\Omega^c} -  \mY_{\Omega^c}  \right\|_F^2  =  \mY_{\Omega^c} - \left(  \mY_{\Omega^c} -  \mV_{\Omega^c} \right)^+.
\]
Finally, the solution to \eqref{admmprog2} is the standard soft thresholding operator (e.g., see \S4.4.3 of \cite{boyd2011distributed}), which reduces the update to
\[\left(\mW^{(2)}_{k+1}\right)_{n,m} = S_{1/\rho}\left( \left( \mW^{(3)}_k - \mU^{(2)}_k \right)_{n,m}\right), ~\mbox{where}~~S_{c}\left(w\right)= \left\{ \begin{array}{cc} w-c & w>c\\ 0 & |w|\leq c\\ w+c & w<-c\end{array}\right.. 
\]
After combining the steps above, we propose Algorithm 2 as a computational scheme to address the Net-Trim central program. The only computational load of the proposed scheme is the linear solve \eqref{admmprog3}, for which the coefficient matrix $\mX^{in}\mX^{in}{}^\top + \mI$ only needs to be calculated once. As observable, the processing time for each ADMM step is relatively low, and only involves few matrix multiplications.

\begin{algorithm}[H]\small 
\centerline{\caption{Implementation of the Net-Trim Central Program}}

\begin{algorithmic}
\STATE{\textbf{input:} $\mX^{in}\in\mathbb{R}^{N\times P}$, $\mX^{out}\in\mathbb{R}^{M\times P}$, $\Omega$, $\mV_\Omega$, $\epsilon$, $\rho$}

\STATE{\textbf{initialize}: $\mU^{(1)}, \mU^{(2)}$ and $\mW^{(3)}$} \qquad\qquad\qquad \qquad   \texttt{\% all initializations can be with $\boldsymbol{0}$}\vspace{.04cm}

\STATE{$\mC\leftarrow \mX^{in}\mX^{in}{}^\top + \mI$\vspace{.04cm}}

\WHILE{not converged}
\STATE{$\mY\leftarrow \mW^{(3)}{}^\top\mX^{in}-\mU^{(1)}$}
 \IF{$\left \|\mY_\Omega - \mX^{out}_\Omega\right \|_F\leq \epsilon$} 
 \STATE {$\mW^{(1)}_\Omega\leftarrow \mY_\Omega$} 
 \ELSE
 \STATE{$\mW^{(1)}_\Omega\leftarrow \mX^{out}_\Omega+\epsilon\left\|\mY_\Omega - \mX^{out}_\Omega\right\|_F^{-1}\left(\mY_\Omega - \mX^{out}_\Omega\right)$} 
 \ENDIF
 \STATE{$\mW^{(1)}_{\Omega^c}\leftarrow  \mY_{\Omega^c} - (  \mY_{\Omega^c} -  \mV_{\Omega^c} )^+$}
\STATE{ $\mW^{(2)}\leftarrow S_{1/\rho}( \mW^{(3)} - \mU^{(2)} )$\quad  \texttt{~~~ \% $S_{1/\rho}$  applies to each element of the matrix}}\vspace{-.0cm}
\STATE{ $\mW^{(3)}\leftarrow \mC^{-1}(\mX^{in}( {\mW^{(1)}}+{\mU^{(1)}})^\top + \mW^{(2)}+\mU^{(2)})$}\vspace{-.0cm}
\STATE{$\mU^{(1)} \leftarrow \mU^{(1)} + \mW^{(1)} - \mW^{(3)}{}^\top\mX^{in}$}
\STATE{$\mU^{(2)} \leftarrow \mU^{(2)} + \mW^{(2)} - \mW^{(3)}$}
\ENDWHILE
\RETURN $\mW^{(3)}$

\end{algorithmic}
\end{algorithm}
\subsection{Net-Trim for Convolutional Layers} Since the convolution operator is linear, similar steps as the ones above can be taken to implement a version of Net-Trim for convolutional layers and inputs in the form of tensors. The main difference is addressing the least-squares update in \eqref{admmprog3}, which can be performed by incorporating the adjoint operator. The details of implementing Net-Trim for convolutional layers are presented in Section \ref{sec:convimp} of the Supplementary Materials.

\section{Experiments and Remarks }\label{sec:conc}
While the main purpose of this paper is introducing a theoretical framework for a class of pruning techniques in deep learning, we briefly present some experiments which highlight the performance of Net-Trim in real-world problems. Due to space limitation, most details of the simulations along with additional experiments are presented in Section \ref{sec:exp} of the Supplementary Materials. Also Net-Trim implementation is made publicly available online\footnote{To access the algorithm implementation, visit: \url{https://dnntoolbox.github.io/Net-Trim/}}.

Our first set of experiments corresponds to a comparison between the cascade and parallel frameworks. For this experiment we use a fully connected (FC) neural network of size $784\times 300 \times 1000 \times 100 \times 10$ (composed of four layers), trained to classify the MNIST dataset. Throughout the section we refer to this network as the FC model. While the theory supports retraining the network with new samples, in practice Net-Trim can be applied to the dataset used to train the original network. In this experiment we also assess the possibility of applying Net-Trim to only a portion of the training data (i.e., working with a subset of columns in $\mX$). Clearly, working with smaller $\mX$ matrices is computationally more desirable. 

\begin{figure}[!htb]\vspace{.5cm}\hspace{-.4cm}
\centering \begin{tabular}{c}
\begin{overpic}[ width=0.25\textwidth,tics=10]{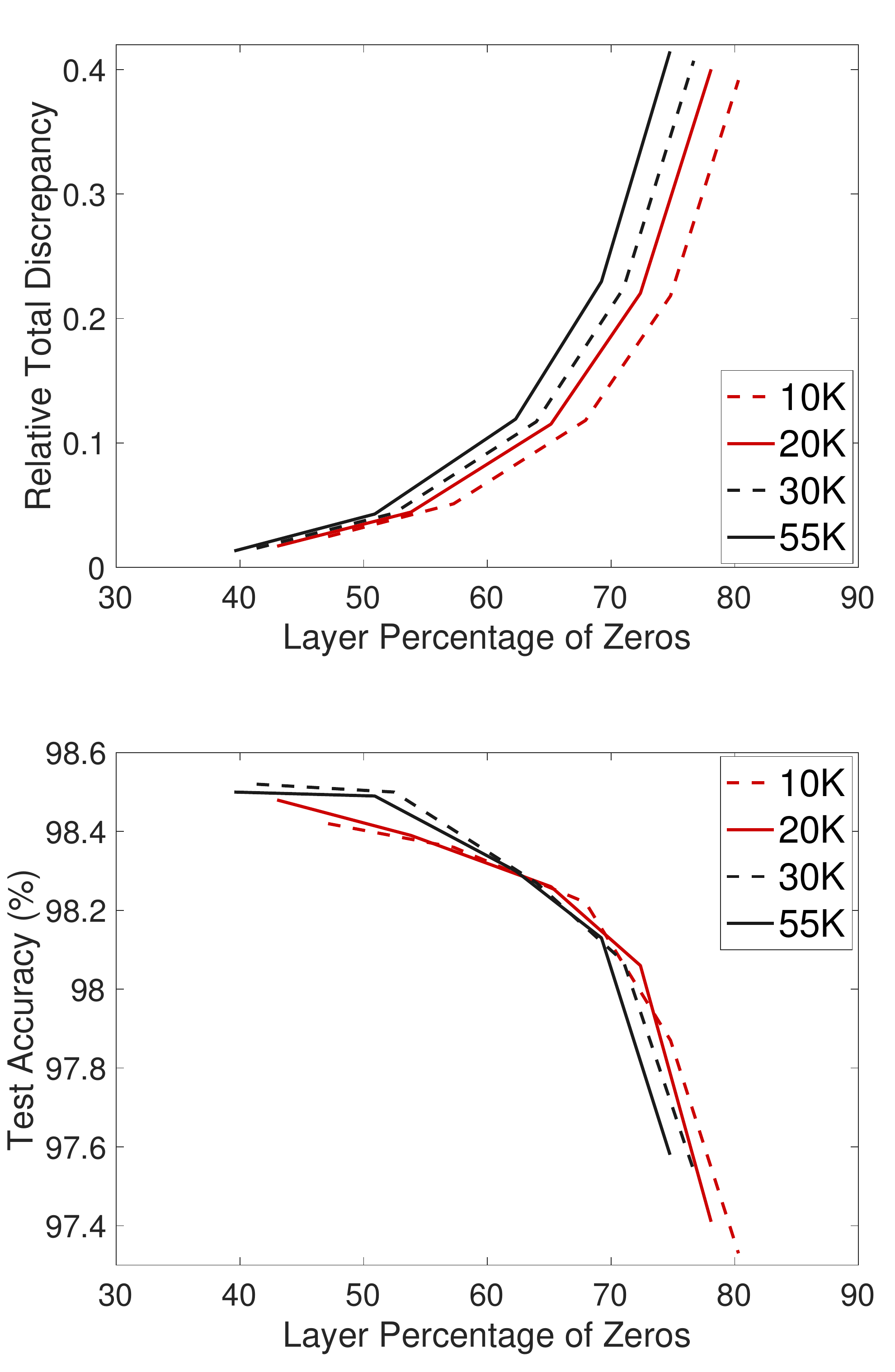}
\put (27,100) {\scalebox{1}{Layer 1}}
\end{overpic}\hspace{-.17cm}
\begin{overpic}[ width=0.25\textwidth,tics=10]{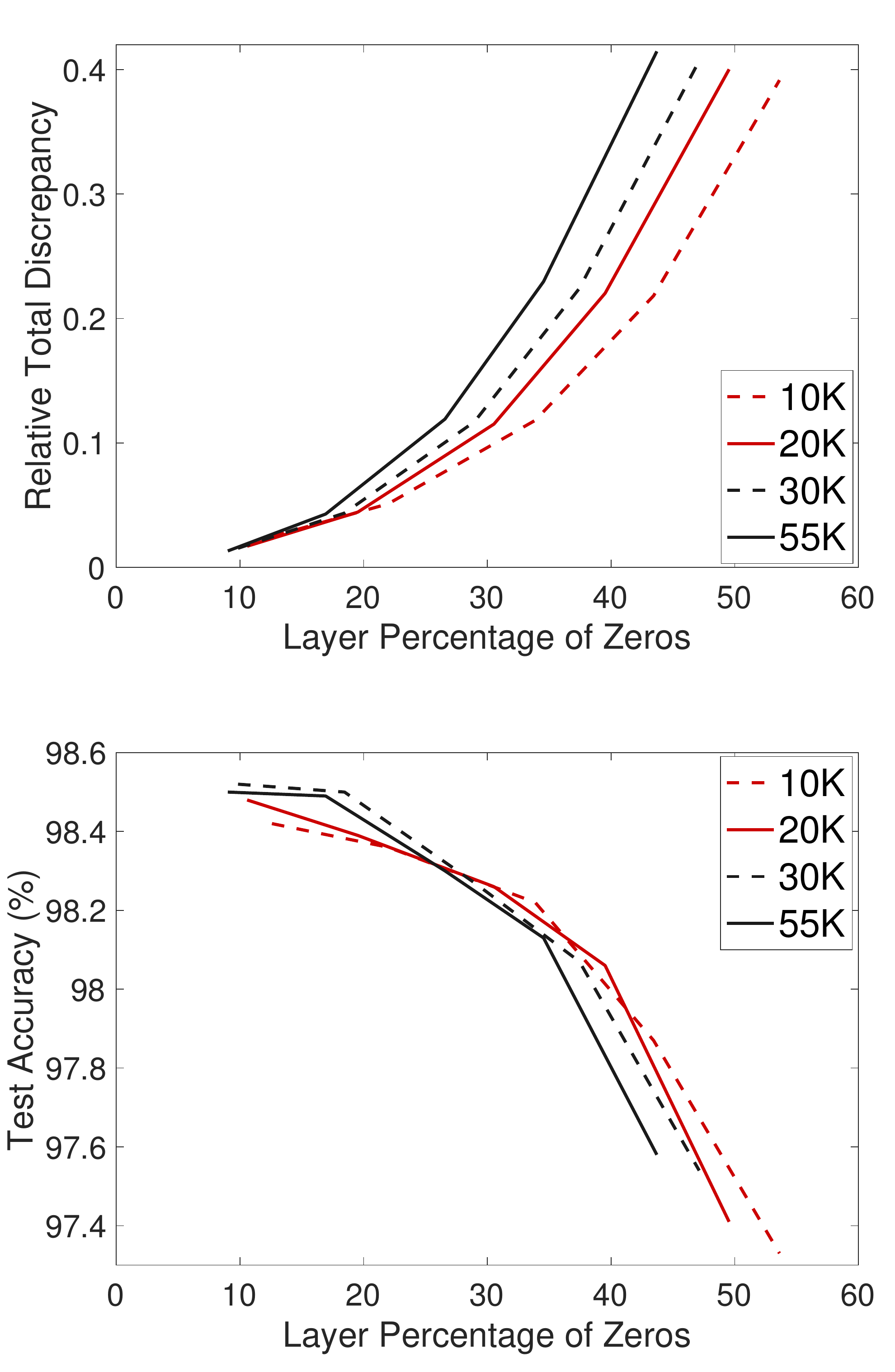}
\put (27,100) {\scalebox{1}{Layer 2}}
\end{overpic}\hspace{-.16cm}
\begin{overpic}[ width=0.25\textwidth,tics=10]{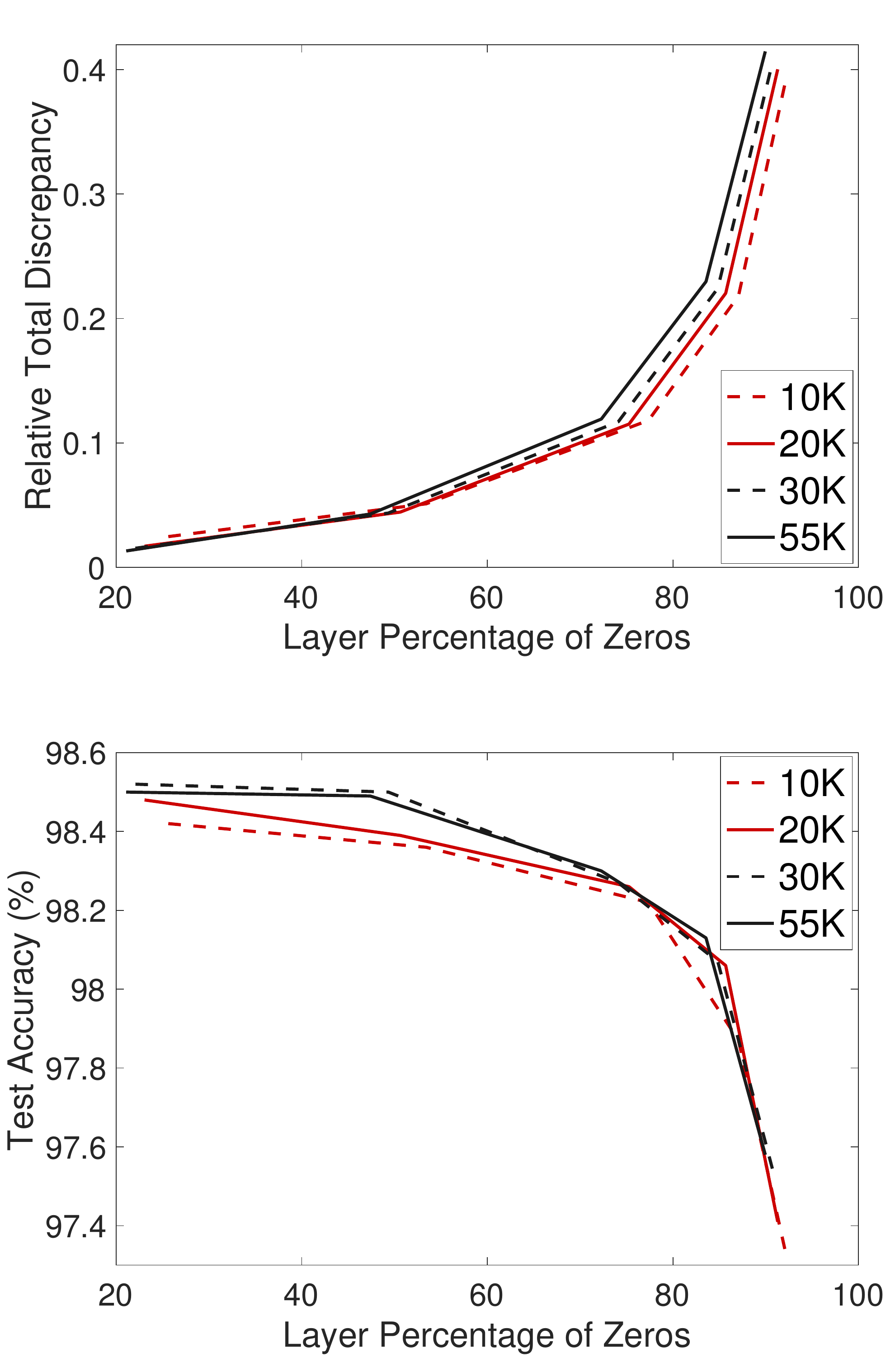}
\put (27,100) {\scalebox{1}{Layer 3}}
\put (1,-1) {\scalebox{.85}{(b)}}
\put (1,50) {\scalebox{.85}{(a)}}
\end{overpic}\hspace{-.17cm}
\begin{overpic}[ width=0.25\textwidth,tics=10]{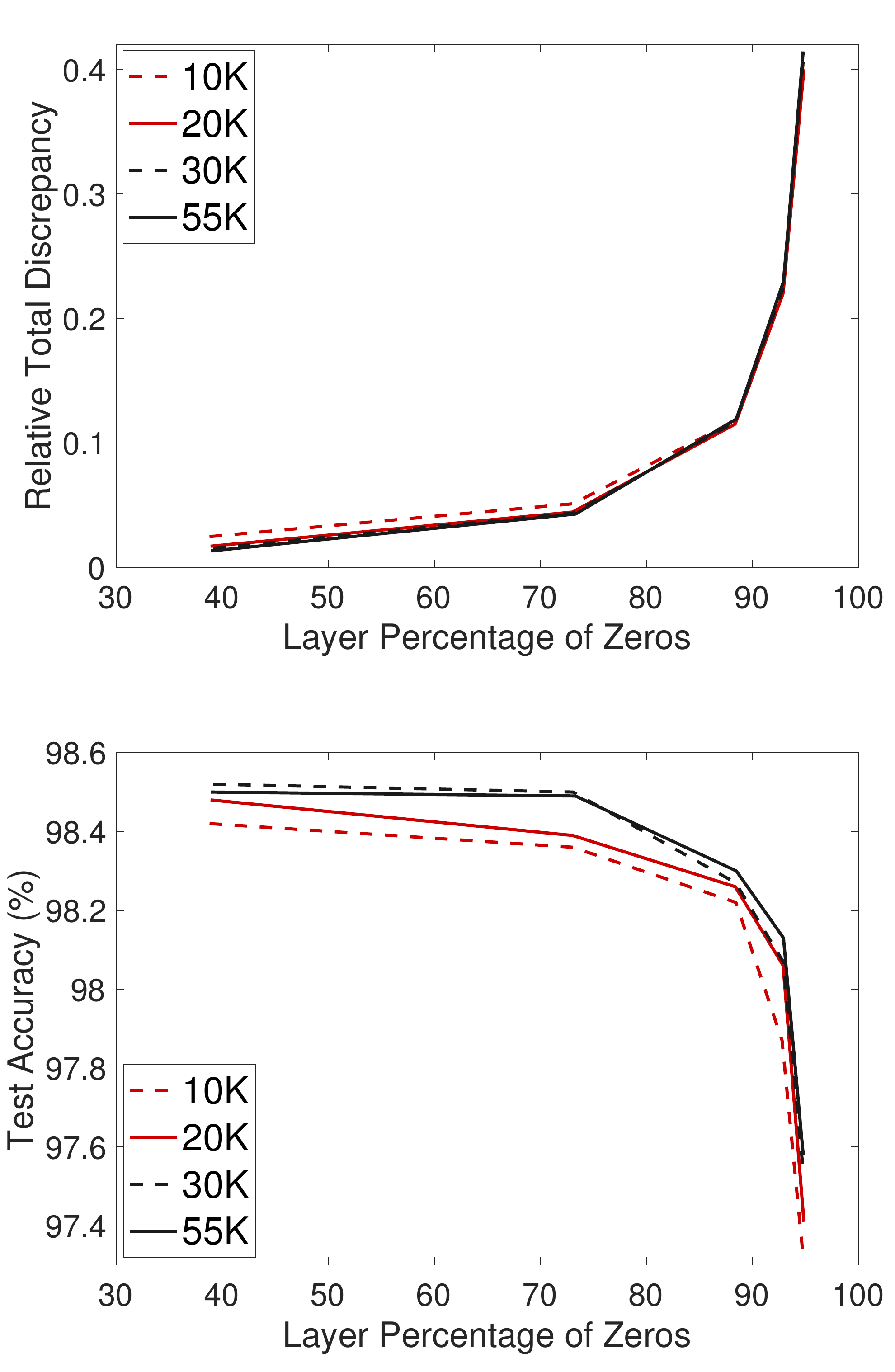}
\put (27,100) {\scalebox{1}{Layer 4}}
\end{overpic}
\\
\begin{overpic}[ width=0.25\textwidth,tics=10]{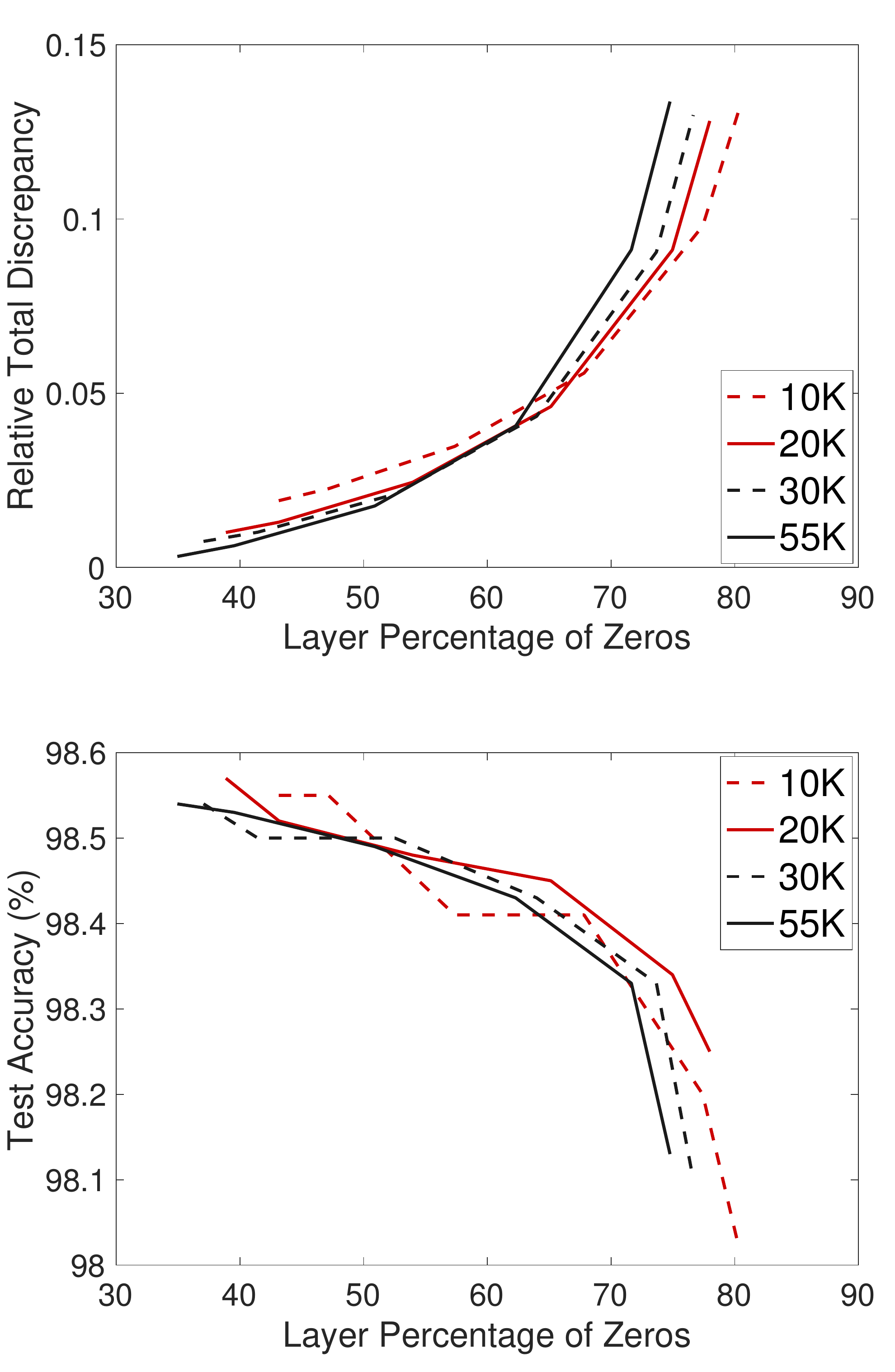}
\end{overpic}\hspace{-.17cm}
\begin{overpic}[ width=0.25\textwidth,tics=10]{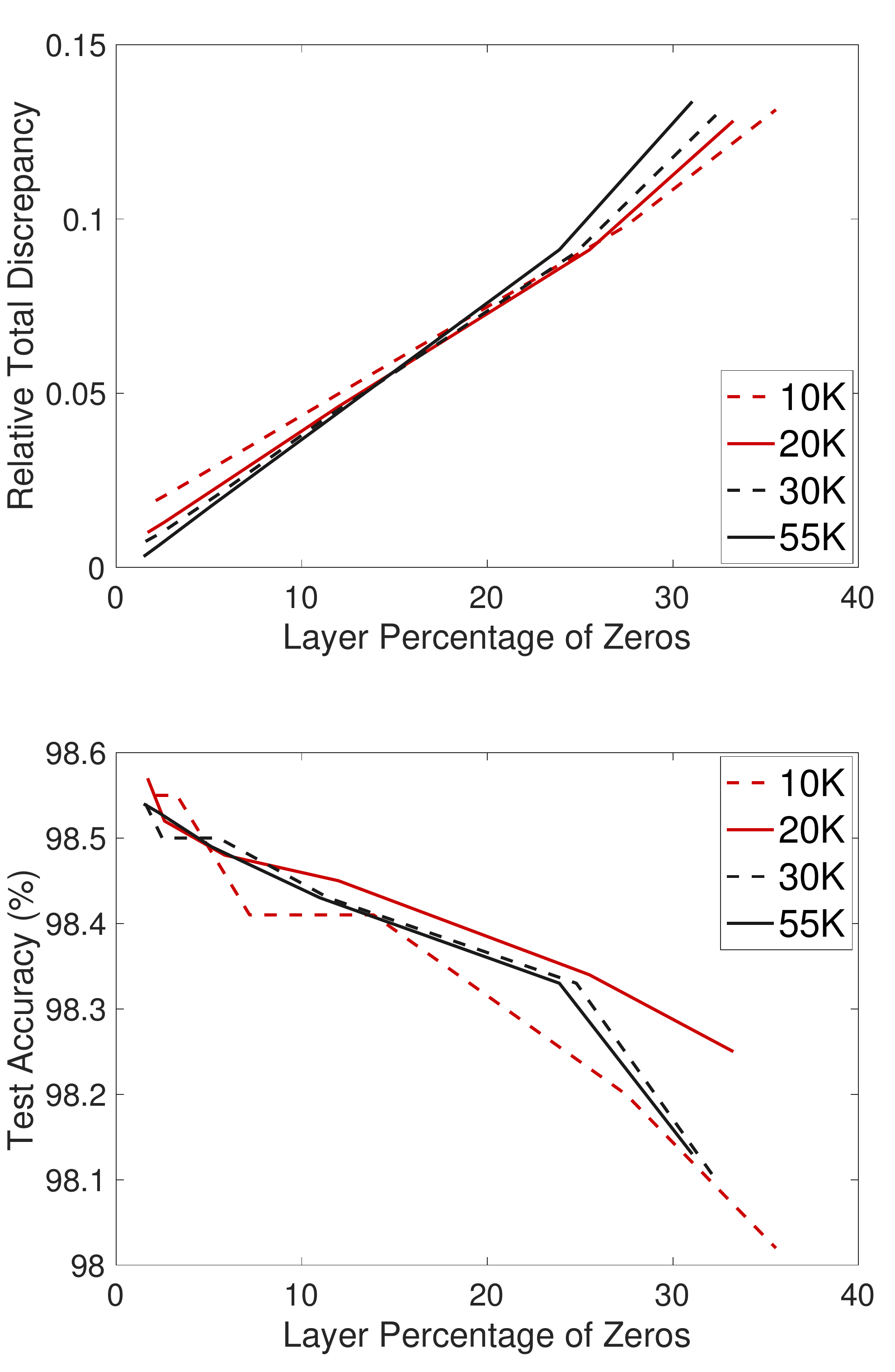}
\end{overpic}\hspace{-.17cm}
\begin{overpic}[ width=0.25\textwidth,tics=10]{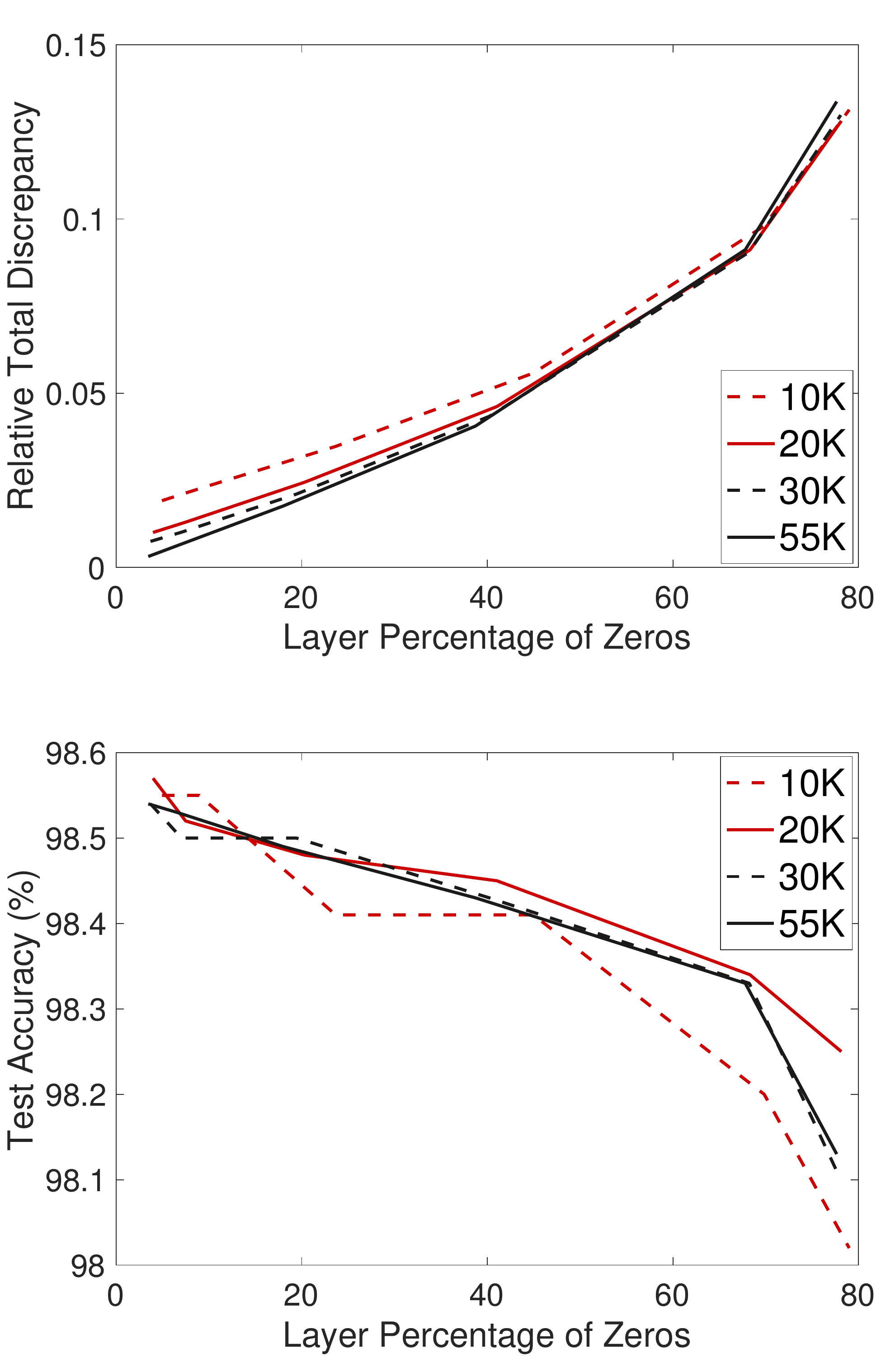}
\put (1,-2) {\scalebox{.85}{(d)}}
\put (1,50) {\scalebox{.85}{(c)}}
\end{overpic}\hspace{-.16cm}
\begin{overpic}[ width=0.25\textwidth,tics=10]{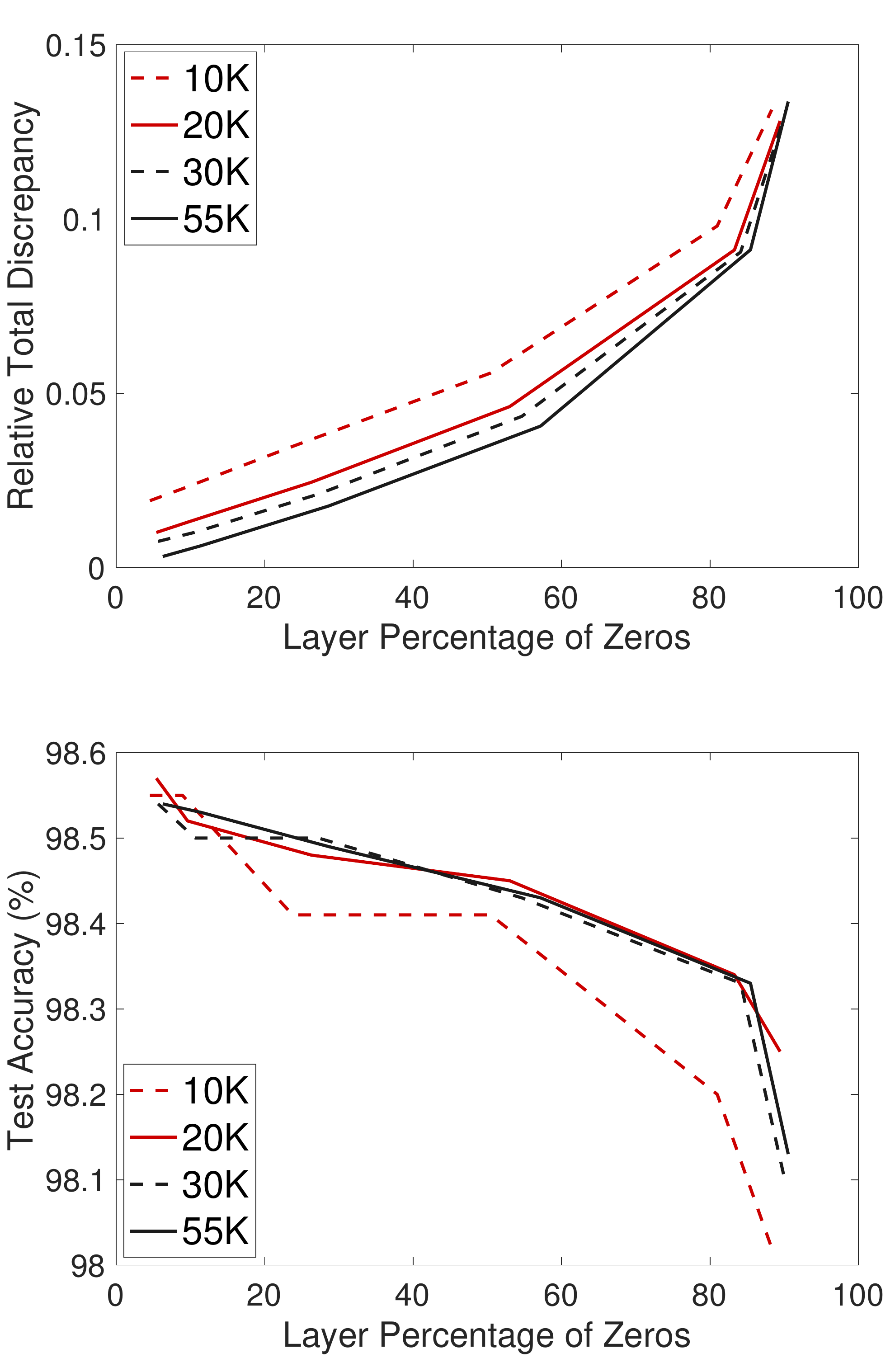}
\end{overpic}
\end{tabular}
 \caption{Retraining the FC network that is initially trained with the full MNIST training set and retrained with 10K, 20K, 30K and 55K samples (each column corresponds to a layer); (a) network relative total discrepancy (RTD) vs the layer percentage of zeros (LPZ) after a parallel scheme; (b) test accuracy vs LPZ after a parallel scheme; (c) RTD vs LPZ after a cascade scheme; (d) test accuracy vs LPZ after a cascade scheme;}\label{fig2m}
\end{figure}

Figure \ref{fig2m} summarizes the parallel and cascade pruning results. A quick comparison between the range of relative discrepancies in panels (a) and (c) (calculated as $\| \hat\mX{}^{(L)}-\mX{}^{(L)}\|_F/\| \mX{}^{(L)}\|_F$) reveals that  for more or less similar sparsity rates, cascade Net-Trim produces a smaller overall discrepancy compared to the parallel scheme (note the axis ranges). This may be considered as the return for going through a non-distributable scheme. However, a comparison of the test accuracies in panels (b) and (d), and especially for larger values of the sparsity ratio, shows a less significant difference between the test accuracies of the two schemes; specifically that using the parallel scheme and its distributable nature is more desirable for big data. 

\begin{figure}[!htb]\vspace{.5cm}\hspace{-.5cm}
\begin{tabular}{c}
\begin{overpic}[height=.314\textwidth,tics=10]{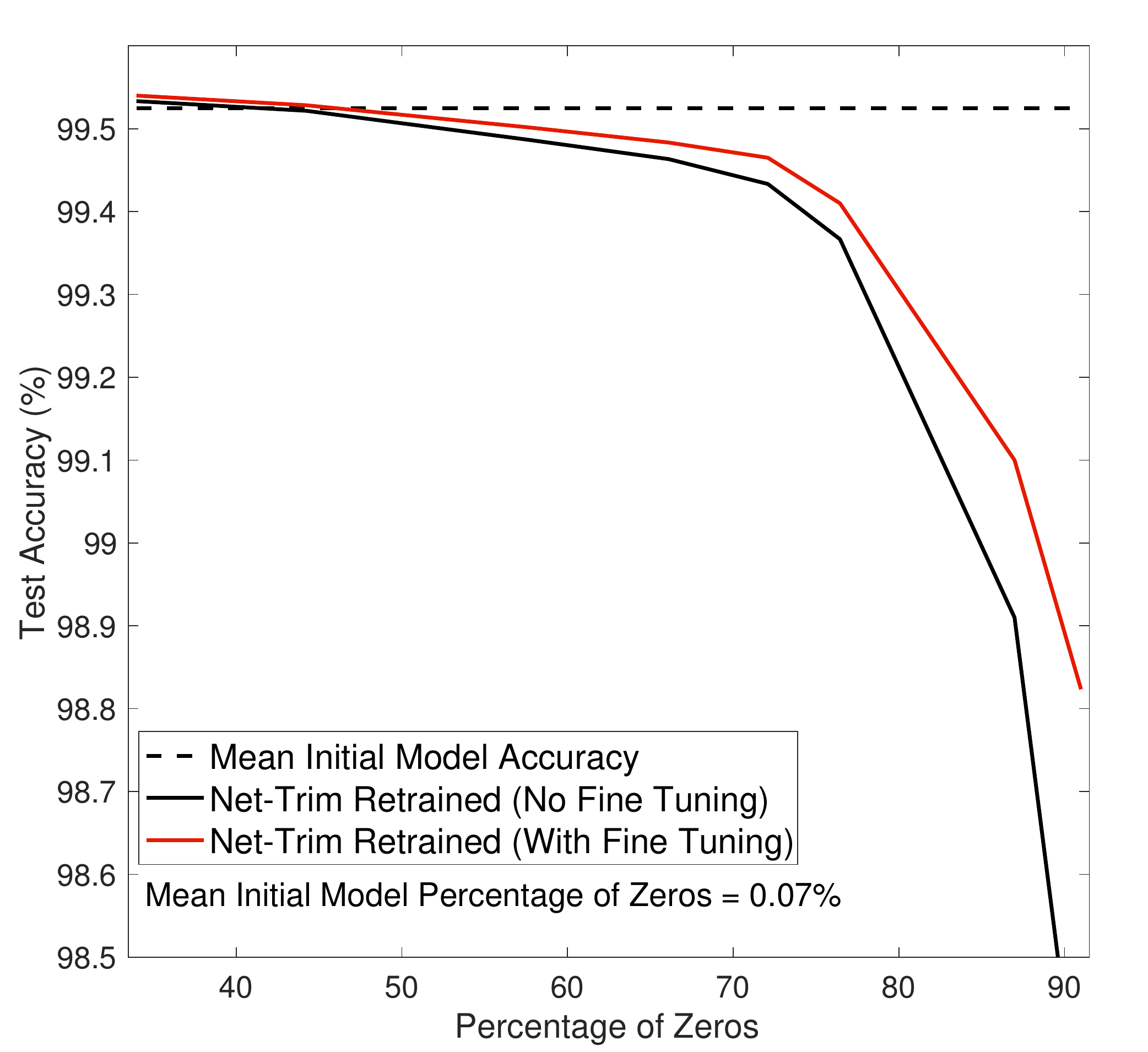}
\put (49,-5.5) {\scalebox{.85}{(a)}}
\end{overpic}\hspace{-.35cm}
\begin{overpic}[ height=.314\textwidth,tics=10]{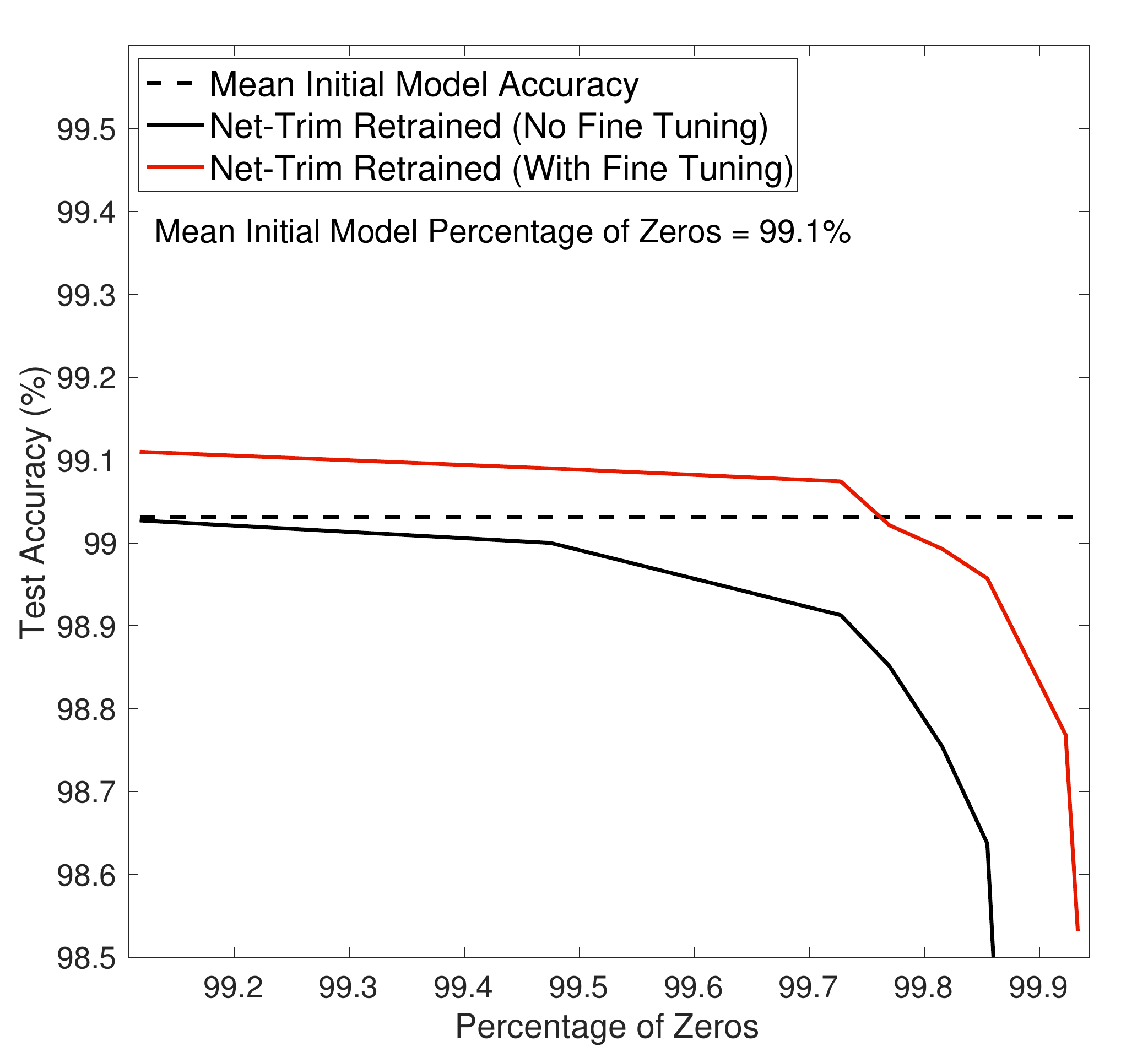}
\put (49,-5.5) {\scalebox{.85}{(b)}}
\end{overpic}\hspace{-.35cm}
\begin{overpic}[ height=.314\textwidth,tics=10]{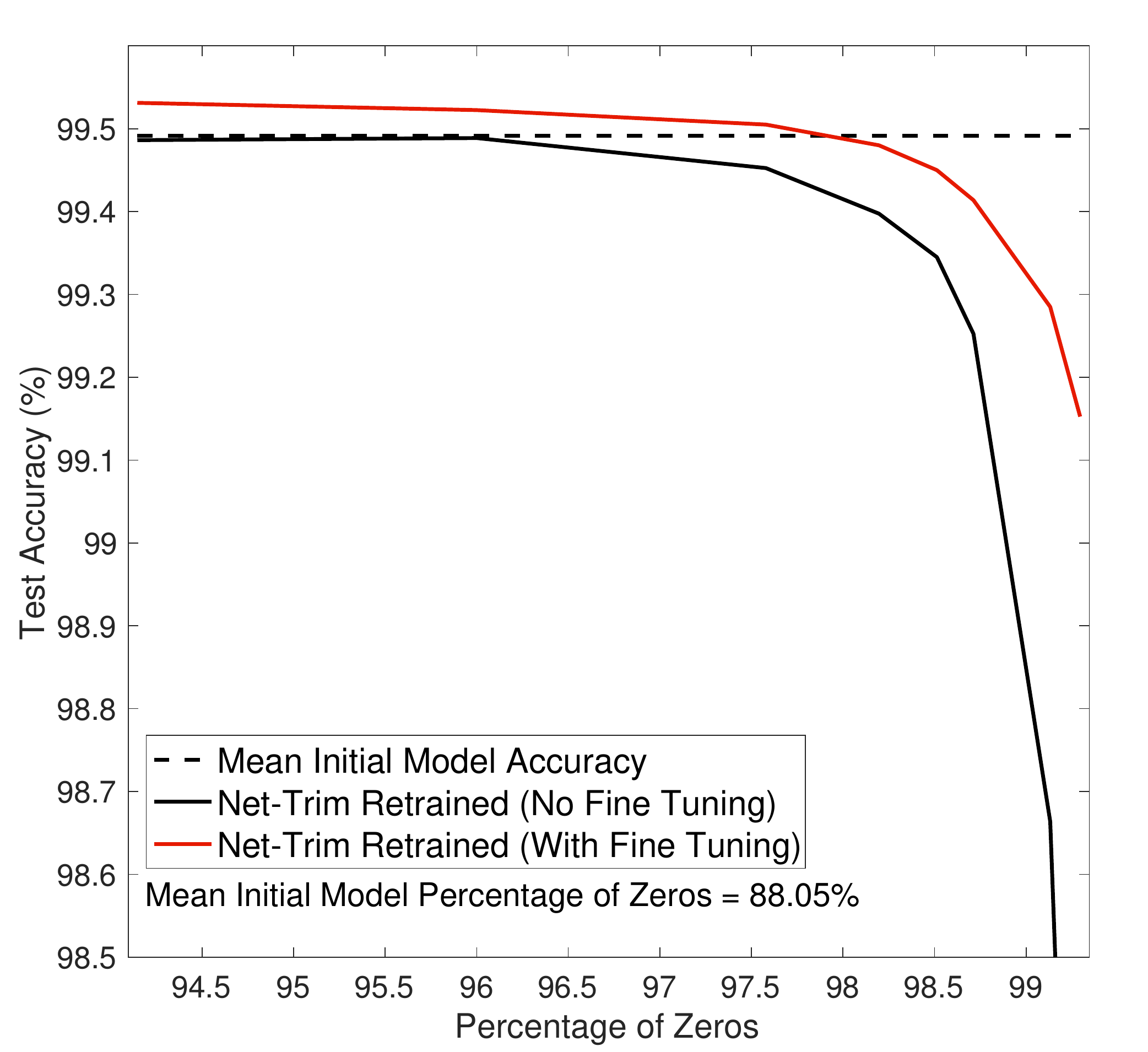}
\put (49,-5.5) {\scalebox{.85}{(c)}}
\end{overpic}
\end{tabular}
 \caption{Mean test accuracy vs mean model sparsity after the application of Net-Trim to the LeNet network initially regularized via $\ell_1$ penalty, Dropout, or both (the regularization parameter and Dropout probability are picked from a range of values and the mean accuracy and sparsity are reported); (a) a model trained with Dropout only: $0.3\leq p \leq 0.8$; (b) a model trained with $\ell_1$ penalty only: $10^{-5}\leq \lambda \leq 5\times 10^{-3}$; (c) a model trained with Dropout and $\ell_1$: $10^{-5}\leq \lambda \leq 2\times 10^{-4}$, $0.5\leq p \leq 0.75$;}\label{fig3m}
\end{figure}

Our next set of experiments corresponds to the application of Net-Trim to the LeNet convolutional network \cite{LecunBBH98} to highlight its performance against well-established methods of Dropout and $\ell_1$ regularization. In these experiments the mean test accuracy and initial model sparsity are reported for the cases of Dropout, $\ell_1$ penalty, and a combination of both. For each run, the tuning parameters ($\lambda$: the coefficient of $\ell_1$-penalty, $p$: the Dropout probability, or both) are varied in a range of values and  the mean quantities are reported. It is noteworthy that Net-Trim can always be followed by an optional fine-tuning step (FT), which performs few training iterations on the weights that Net-Trim has left nonzero. The plots in Figure \ref{fig3m} show how the application of Net-Trim can further contribute to the sparsity and accuracy of the network. For instance panel (c) indicates that without a loss in the accuracy, applying Net-Trim to a network, where almost 88\% of the weights are pruned via Dropout and $\ell_1$ regularization, can elevate the sparsity to almost 98\%.

\begin{figure}[!htb]\vspace{.5cm}\hspace{-.17cm}
\centering \begin{tabular}{c}
\begin{overpic}[ trim={1.5cm .8cm 0 0},clip, height=.29\textwidth,tics=10]{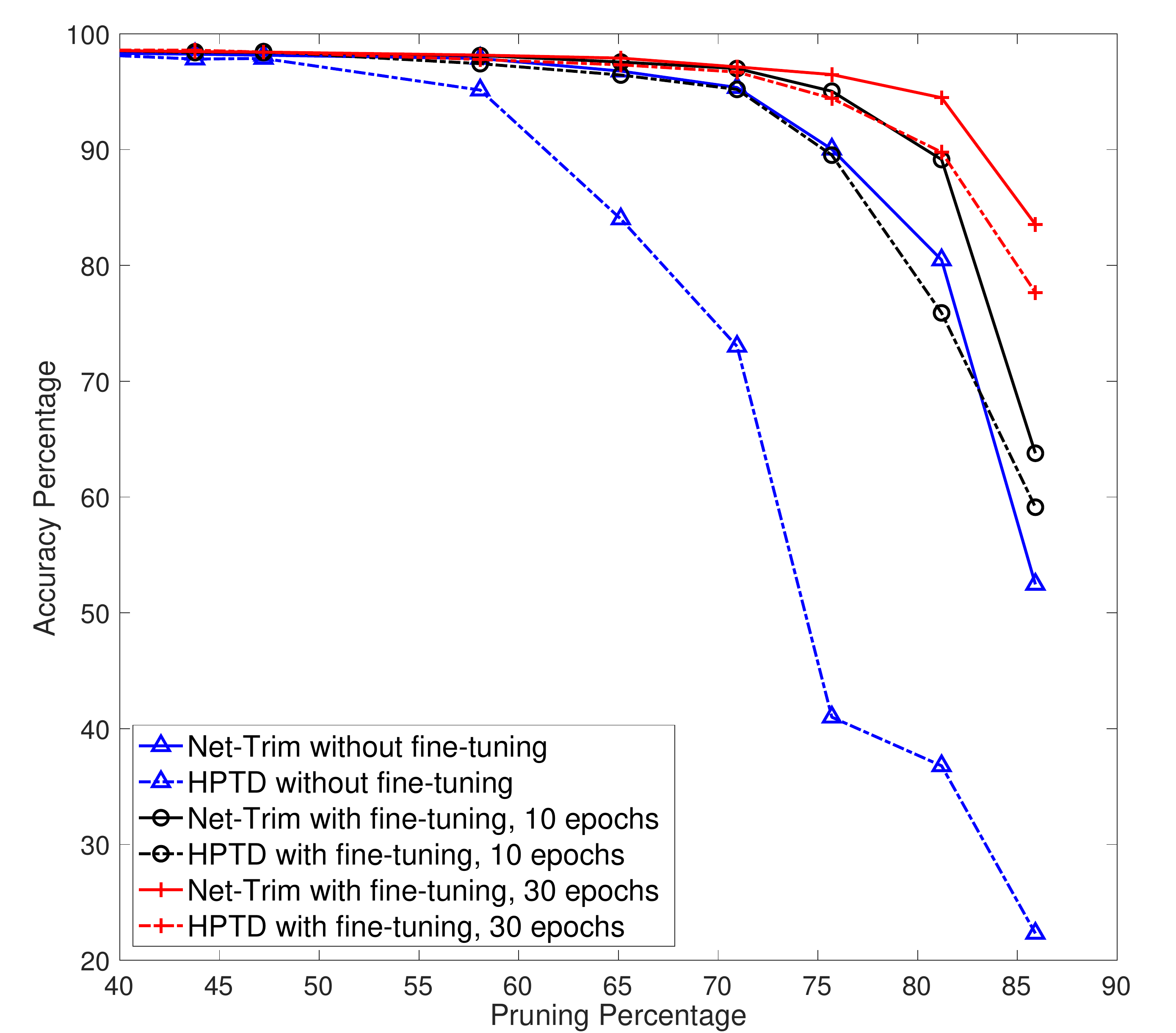}
\put (-2,35){\rotatebox{90}{\scalebox{.5}{Test Accuracy (\%)}}}
\put (35,-3){\rotatebox{0}{\scalebox{.5}{Percentage of Zeros}}}
\end{overpic}\hspace{-.1cm}
\begin{overpic}[trim={1.4cm .8cm 0 0},clip,height=.29\textwidth,tics=10]{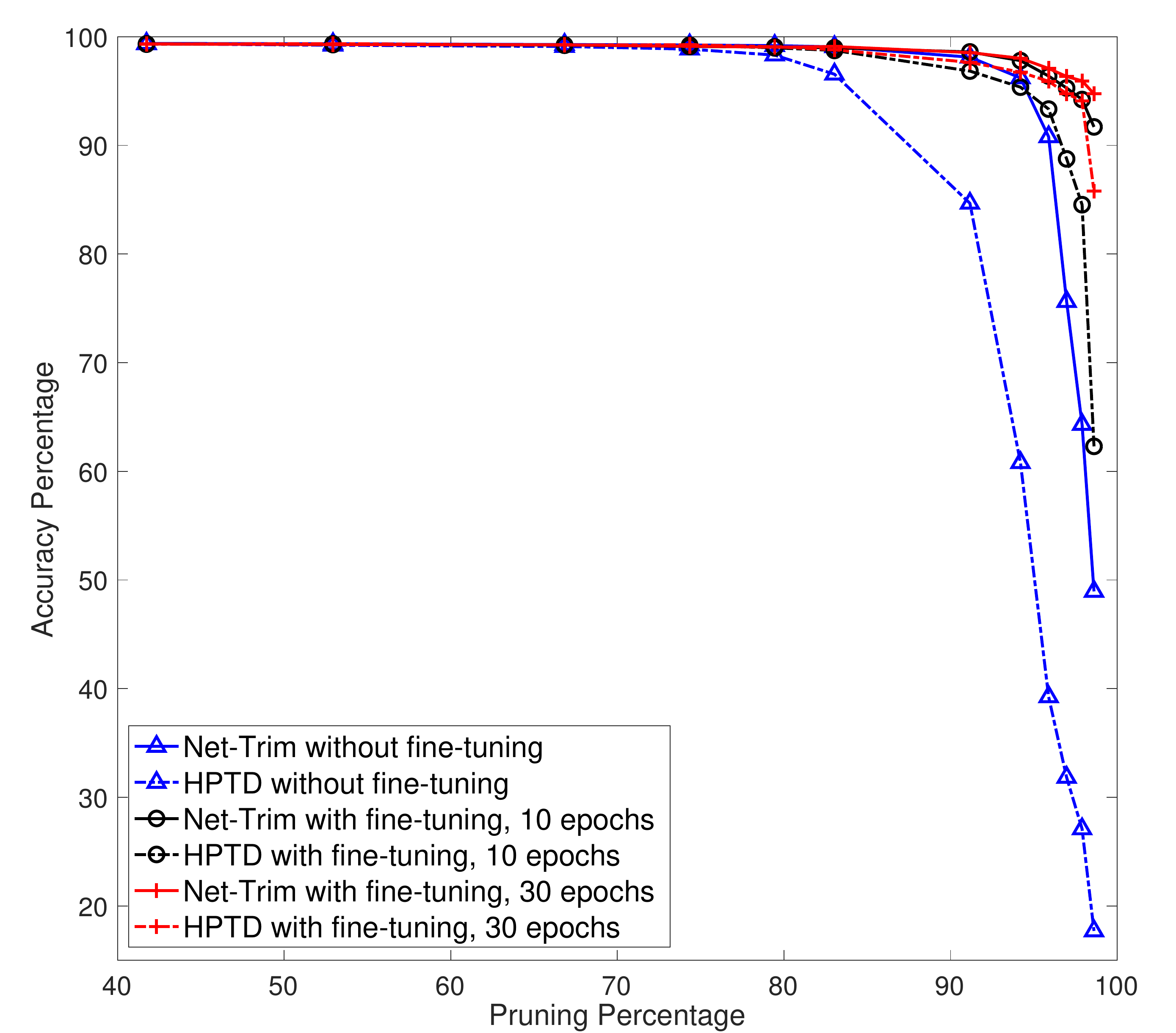}
\put (-2,35){\rotatebox{90}{\scalebox{.5}{Test Accuracy (\%)}}}
\put (35,-3){\rotatebox{0}{\scalebox{.5}{Percentage of Zeros}}}
\end{overpic}\hspace{-.1cm}
\hspace{.1cm}\begin{overpic}[ trim={1.8cm .8cm 0 0},clip,height=.29\textwidth,tics=10]{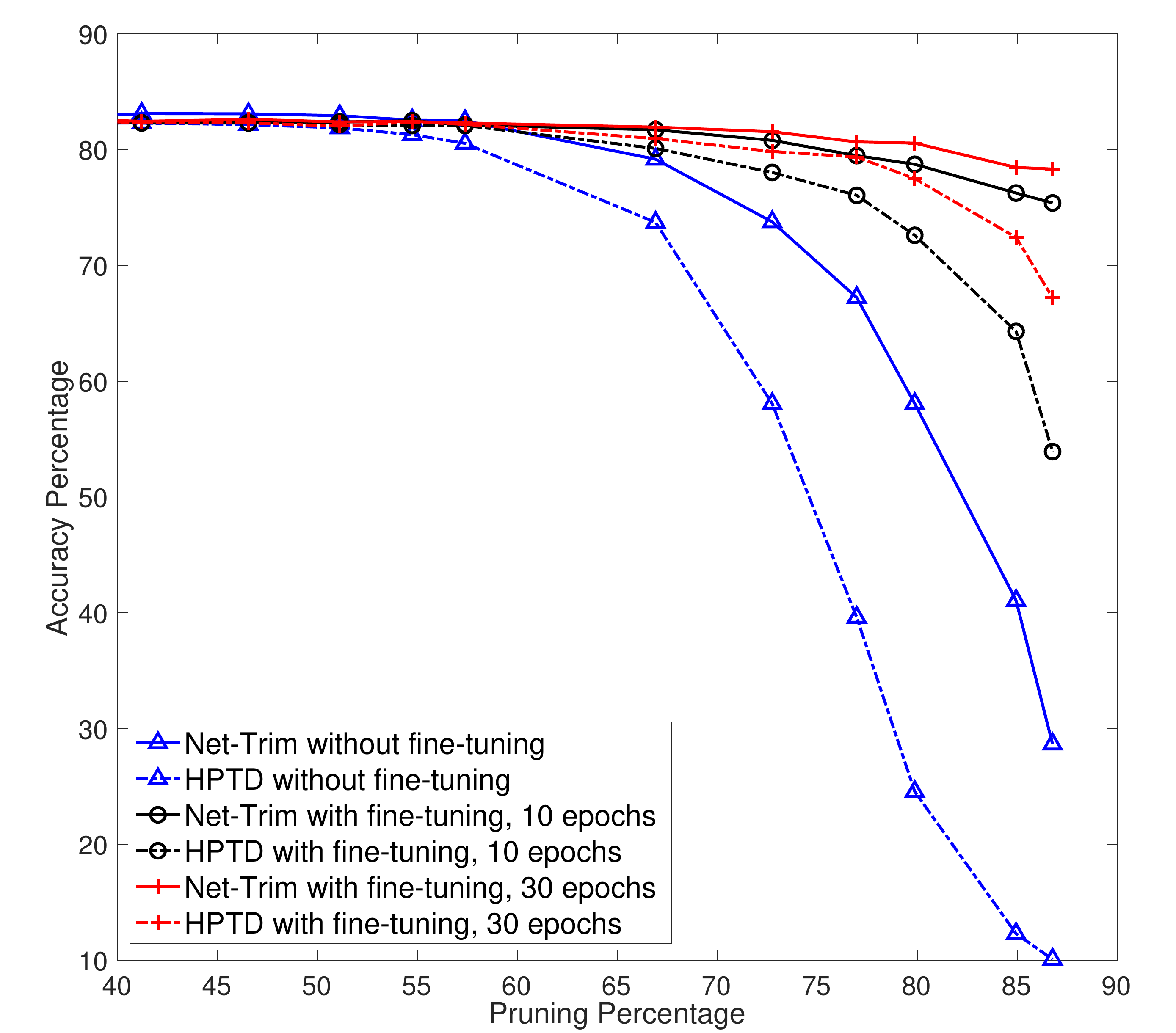}
\put (-4,35){\rotatebox{90}{\scalebox{.5}{Test Accuracy (\%)}}}
\put (35,-3){\rotatebox{0}{\scalebox{.5}{Percentage of Zeros}}}
\end{overpic}\hspace{-.06cm}
\end{tabular}
 \caption{Comparison of Net-Trim and HPTD test accuracy vs percentage of zeros, without fine-tuning, and with fine-tuning using 10 and 30 epochs (a) FC model, (b) LeNet model; (c) CIFAR-10 model;}\label{fig4m}
\end{figure}

Another well-known scheme in model pruning is the algorithm by Han, Pool, Tran and Dally (HPTD: \cite{han2015learning}). The HPTD algorithm is a heuristic tool used for network compression, which truncates the small weights across a trained network and performs another round of training on the active weights (same as the fine-tuning scheme explained above). 
Figure \ref{fig4m} presents a comprehensive comparison between the Net-Trim and HPTD on the FC, LeNet, and a CIFAR-10 model. The initial CIFAR-10 model uses an augmented training set of size 6.4M samples, to retrain which Net-Trim uses 50K samples. 
One of the main drawbacks with the HPTD is the truncation based on the magnitude of the weights, which in many cases may discard connections to the important features and variables in the network. That is mainly the reason that Net-Trim consistently outperforms this method. In fact, Net-Trim can also present vital information about the data structure and important features that are not immediately available using other techniques. 

\begin{figure}[htb!]
\centering\begin{overpic}[width=5.1in,height=2.3in]{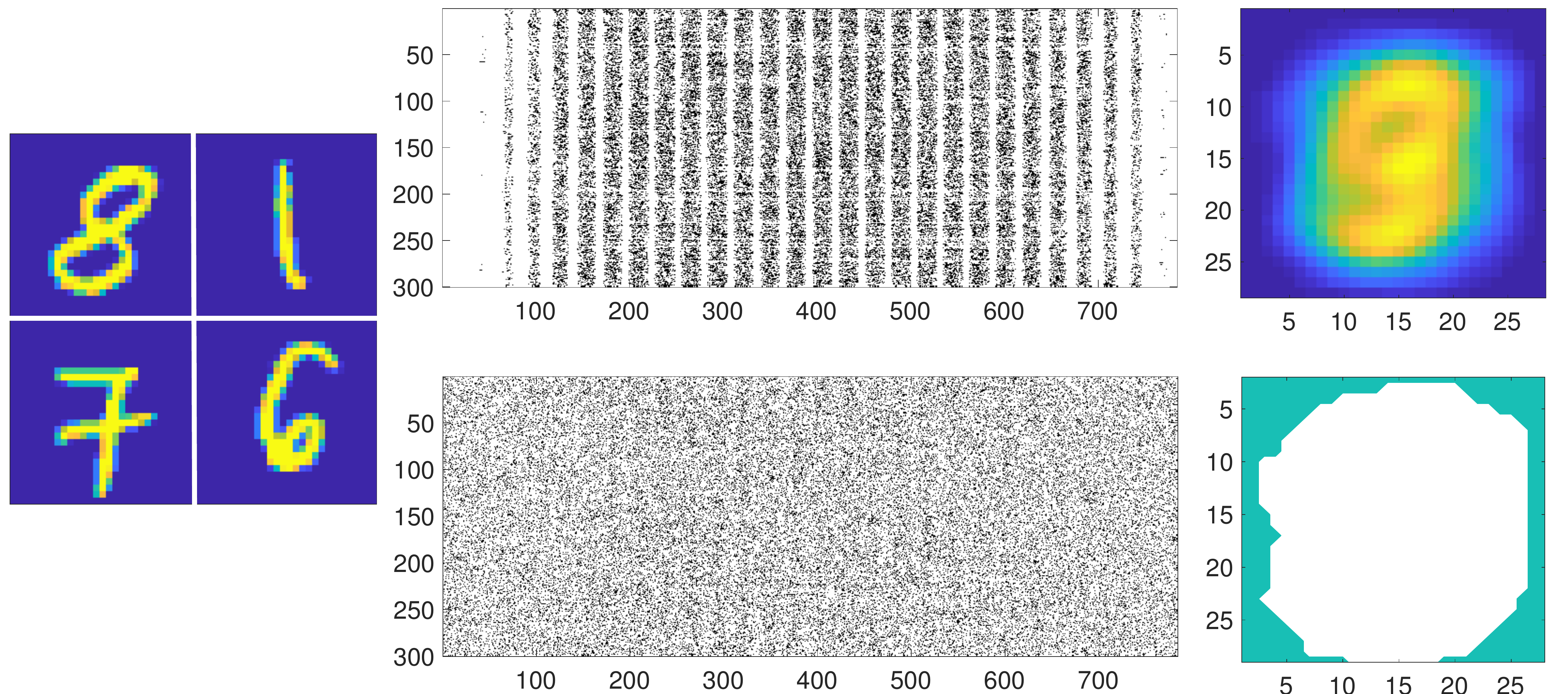}
\put (11.2,9) {\scalebox{.75}{(a)}} 
\put (51,21.8) {\scalebox{.75}{(b)}} 
\put (51,-2.5) {\scalebox{.75}{(c)}} 
\put (87.5,21.5) {\scalebox{.75}{(d)}} 
\put (87.5,-2.5) {\scalebox{.75}{(e)}} 
\end{overpic}\\[.2cm]
 \caption{Instant identification of important features using Net-Trim; (a) samples from MNIST dataset; (b) visualization of $\hat\mW_1{}^\top$ in the FC retrained model using Net-Trim; (c) similar visualization of $\hat\mW_1{}^\top$ in the FC retrained model, using HPTD with a single fine-tuning step; (d) histogram of the pixel values in the MNIST data set; (e) the green mask corresponding to the zero columns in panel (b);}\label{fig5m}\vspace{-.5cm}
\end{figure}

In Figure \ref{fig5m} we have depicted the retrained $\hat\mW_1$ matrix of the FC model after applying Net-Trim and HPTD. In panel (b) we can see many columns that are fully zero. After plotting the histogram of the MNIST samples (as in panel (d)), one would immediately observe that the zero columns in $\hat\mW_1$ correspond to the boundary pixels with the least level of information. As HPTD only relies on the truncation based on the weight magnitudes, despite the similar number of zeros in panels (b) and (c), the latter does not highlight such data structure. To obtain a similar pattern as in panel (b), the authors in \cite{han2015learning} suggest an iterative pruning path with a fine-tuning after truncating a portion of the network weights. However, this is not a computationally efficient path as it requires retraining the network multiple times, which can take a lot of time for large data sets and is not guaranteed to identify the right structures. 

\subsection{Concluding Remarks} Net-Trim can be generalized to a large class of problems, where the architecture of each layer in a trained network is restructured via a program of the type
\begin{equation}\label{eqgen}
\minimize_{\mU\in\mathbb{R}^{N\times M}}~~\mathcal{R}\left(\mU\right)\qquad \mbox{subject to} \qquad \sigma\left(\mU^\top\mX^{in}\right)\approx \mX^{out}.
\end{equation}
The objective $\mathcal{R}(\cdot)$ aims to promote a desired structure, and the constraint enforces a consistency between the initial and retrained models. While in this paper we merely emphasized on $\mathcal{R}(\mU)=\|\mU\|_1$, a variety of other structures may be explored by adaptively selecting the objective. For instance, other than the Ridge and the elastic net penalties as regularizing tools, choosing $\mathcal{R}(\mU)=\|\mU\|_{2,1} = \sum_{n=1}^N\|\mU_{n,:}\|$ can promote selection of a subset of the rows in $\mX^{in}$, and act as a feature selection or node-dropping tool for each layer. Total variation, or rank penalizing objectives may also directly apply to network compression problems. 

While in this paper we specifically focused on $ \sigma = \ReLU(\cdot)$ to exploit the convex formulation, in principal other forms of activation may be explored. Even if a convex (re)formulation is suboptimal or not possible, powerful tools from non-convex analysis would still allow us to have an understanding of when and how well programs of type \eqref{eqgen} work. Clearly, the techniques used for such type of analysis might be initialization-sensitive, and different than those used in this paper.

Net-Trim can specifically become a useful tool when the number of training samples is limited. While overfitting is likely to happen in this situation, Net-Trim allows reducing the complexity of the models, yet maintaining the consistency with the original model. From a different perspective, Net-Trim may simplify the process of determining the network size. For large networks that are trained with insufficient samples, employing Net-Trim can reduce the size of the models to an order matching the data.

\section{Proofs}\label{sec:proof}
Before we start a detailed proof of the results, we would like to state two inequalities that will be frequently used throughout this section:
\begin{align}\label{oq1}
\forall ~\mX\in\mathbb{R}^{d_1\times d_2}:\qquad &\left\| \mX^+\right\|_F \leq \left\| \mX\right\|_F,\\ \forall ~\mX,\mY\in\mathbb{R}^{d_1\times d_2}: \qquad &\left\| \mX^+-\mY^+\right\|_F \leq \left\| \mX-\mY\right\|_F.\label{oq2}
\end{align}
The first inequality is straightforward to verify. To verify \eqref{oq2} we note that for all $x,y\in\mathbb{R}$:
\[x^+ = (x-y + y)^+\leq (x-y)^+ + y^+ \leq |x-y| + y^+,
\]
which is interchangeable in $x$ and $y$, and yields $|x^+-y^+|\leq |x-y|$.

\subsection{Proof of Theorem \ref{thpar}}
The central convex program \eqref{eqconv2} requires that for $\Omega = \supp ~\mX^{out}$:
\begin{align}\label{oq3}
\left\|\left(\hat\mW^\top\mX^{in} - \mX^{out} \right)_\Omega\right \|_F\leq \epsilon, ~~~\mbox{and}~~~  
\left(\hat\mW^\top\mX^{in}\right )_{\Omega^c} \leq 0.
\end{align}
As the first step, notice that for $\tilde\mX{}^{out} = (\hat\mW^\top \mX^{in})^+$ one has
\begin{align}\notag 
\left\| \tilde\mX{}^{out} - \mX{}^{out} \right\|_F^2 &= \left\| \left(\tilde\mX{}^{out} - \mX{}^{out} \right)_\Omega\right\|_F^2 + \left\| \left(  \tilde\mX{}^{out}- \mX{}^{out}\right)_{\Omega^c}\right\|_F^2 \\
\notag &=\left\|  \left(\hat\mW^\top\mX^{in} \right)^+_\Omega-\mX^{out}_\Omega\right\|^2_F\\ 
\notag & = \left\|  \left(\hat\mW^\top\mX^{in} \right)^+_\Omega - \left(\mX^{out}\right)^+_\Omega\right\|^2_F\\ 
\notag & \leq \left\|\left(  \hat\mW^\top\mX^{in} - \mX^{out}\right)_\Omega\right\|^2_F\\ 
\label{oq4} &\leq \epsilon^2,
\end{align}
where the first inequality is thanks to \eqref{oq2}. Now consider $\hat\mX{}^{in}$ be any matrix such that $\|\hat\mX{}^{in} - \mX{}^{in} \|_F \leq \epsilon_{in}$, 
then for $\hat\mX{}^{out} = (\hat\mW^\top \hat \mX{}^{in})^+$ one has
\begin{align}\notag
\left\| \hat\mX{}^{out} - \mX{}^{out} \right\|_F &\leq \left\| \hat\mX{}^{out} - \tilde \mX{}^{out} \right\|_F + \left\| \tilde \mX{}^{out}-\mX{}^{out} \right\|_F\\
\notag & \leq \left\| \left(\hat\mW^\top \hat \mX{}^{in}\right )^+ - \left(\hat\mW^\top  \mX{}^{in}\right)^+\right\|_F + \epsilon \\
\notag & \leq \left\| \hat\mW^\top \left(\hat \mX{}^{in} -  \mX{}^{in}\right)\right\|_F + \epsilon \\
\notag & \leq \left\| \hat\mW\right\|_F\left\| \hat \mX{}^{in} -  \mX{}^{in}\right\|_F + \epsilon\\
&\leq \epsilon_{in} + \epsilon.\label{oq5}
\end{align}
To present the last inequality we used the fact that 
\[\left\|\hat\mW\right \|_F \leq \left\|\hat\mW\right \|_1 \leq \left\|\mW\right \|_1 = 1.
\]
We may now complete the proof via a simple induction. For the parallel scheme sketched in \eqref{training layer ell}, inequality \eqref{oq4} implies that $\|\hat\mX{}^{(1)} - \mX{}^{(1)} \|_F\leq \epsilon_1$. Also, \eqref{oq4} requires that $\|\hat\mX{}^{(\ell)} - \mX{}^{(\ell)} \|_F\leq \epsilon_\ell$, and assuming that $\|\hat\mX{}^{(\ell-1)} - \mX{}^{(\ell-1)} \|_F\leq \sum_{j=1}^{\ell-1} \epsilon_j$, \eqref{oq5} yields 
\[\left\|\hat\mX{}^{(\ell)} - \mX{}^{(\ell)} \right\|_F\leq \sum_{j=1}^{\ell} \epsilon_j.
\]

\subsection{Proof of Theorem \ref{thcas}}
For the cascade scheme outlined in Algorithm 1, replacing the $\ell$ indexing with the $in/out$ notation, the layer retraining takes place by addressing the convex program 
\begin{equation}\label{casinout}
\hat{\mW}= \operatorname*{arg\,min}_{\mU}~\left\|\mU\right\|_{1}\quad \mbox{subject to} \quad \mU\in \mathcal{C}_{\epsilon}\left(\hat\mX{}^{in},\mX^{out},\mW^\top\hat\mX{}^{in}\right),
\end{equation}
where $\hat\mX{}^{in}$ is the retrained model input, $\mX^{out} = (\mW^\top\mX^{in})^+$ is the initially trained model output, and for $\Omega = \supp ~\mX^{out}$,
\begin{equation*}
\epsilon= \gamma \left\|\left(\mW ^\top\hat\mX{}^{in}- \mX^{out}\right)_\Omega\right\|_F.
\end{equation*}
The central convex program \eqref{casinout}, hence requires that
\begin{align}\label{oq6}
\left\|\left(\hat\mW^\top\hat\mX{}^{in} - \mX{}^{out} \right)_\Omega\right \|_F &\leq \gamma\left\| \left( \mW^\top \hat\mX{}^{in} -  \mX^{out} \right)_\Omega\right\|_F, \\
\left(\hat\mW^\top\hat\mX{}^{in}\right )_{\Omega^c} &\leq \left(\mW^\top\hat\mX{}^{in}\right )_{\Omega^c}.\label{oq7}
\end{align}
For the output of the initial and retrained models, one has
\begin{align}\label{oq8}
\left\| \hat\mX{}^{out} - \mX{}^{out} \right\|_F^2 &= \left\| \left( \hat \mW^\top \hat \mX{}^{in} \right)_\Omega^+ - \mX{}^{out}_\Omega  \right\|_F^2 + \left\| \left( \hat \mW^\top \hat \mX{}^{in} \right)_{\Omega^c}^+ \right\|_F^2.
\end{align}
For the first term in \eqref{oq8} thanks to \eqref{oq2} and \eqref{oq6}, one has
\begin{align}\notag 
\left\| \left( \hat \mW^\top \hat \mX{}^{in} \right)_\Omega^+ - \mX{}^{out}_\Omega  \right\|_F^2 &= \left\| \left( \hat \mW^\top \hat \mX{}^{in} \right)_\Omega^+ - \left(\mX{}^{out}\right)_\Omega^+  \right\|_F^2\\ \notag 
& \leq \left\| \left( \hat \mW^\top \hat \mX{}^{in}  - \mX{}^{out}\right)_\Omega  \right\|_F^2\\
&\leq \gamma^2\left\| \left( \mW^\top \hat\mX{}^{in} -  \mX^{out} \right)_\Omega\right\|_F^2.    \label{oq9}
\end{align}
The second term in \eqref{oq8} can also be bounded by
\begin{align}\notag 
\left\| \left( \hat \mW^\top \hat \mX{}^{in} \right)_{\Omega^c}^+ \right\|_F^2\! \leq \left\| \left(  \mW^\top \hat \mX{}^{in} \right)_{\Omega^c}^+ \right\|_F^2 &= \left\| \left(  \mW^\top \hat \mX{}^{in} \right)_{\Omega^c}^+ - \left(  \mW^\top \mX{}^{in} \right)_{\Omega^c}^+ \right\|_F^2 \\&\leq  \left\| \left(  \mW^\top \hat \mX{}^{in} -  \mW^\top  \mX{}^{in}\right)_{\Omega^c} \right\|_F^2. \label{oq10}
\end{align}
Using $\mX^{out}_\Omega  = (\mW^\top\mX^{in})_\Omega$, and applying the results of \eqref{oq9} and \eqref{oq10} to \eqref{oq8} yields 
\begin{align}\notag
\left\| \hat\mX{}^{out} - \mX{}^{out} \right\|_F^2 &\leq \gamma^2\left\| \left( \mW^\top \hat\mX{}^{in} -  \mW^\top\mX^{in} \right)_\Omega\right\|_F^2 + \left\| \left(  \mW^\top \hat \mX{}^{in} -  \mW^\top \mX{}^{in} \right)_{\Omega^c} \right\|_F^2\\ \notag  &\leq   \gamma^2 \left\|   \mW^\top \left(\hat \mX{}^{in} -  \mX^{in} \right)\right\|_F^2\\ &\leq \gamma^2 \left\|   \hat \mX{}^{in} -  \mX^{in} \right\|_F^2.\label{eqcasfinal}
\end{align}
In a cascade Net-Trim, the first layer goes through the standard retraining \eqref{eqfirst} with $\epsilon_1=\epsilon$, for which Theorem \ref{thpar} warrants $\|\hat\mX{}^{(1)} - \mX^{(1)}\|_F\leq \epsilon$. On the other hand, for $\ell\geq 2$, \eqref{eqcasfinal} warrants $\|\hat\mX{}^{(\ell)} - \mX^{(\ell)}\|_F\leq \gamma_\ell \|\hat\mX{}^{(\ell-1)} - \mX^{(\ell-1)}\|_F$, which together with the discrepancy of the first layer yield the advertised result in \eqref{eqcasthstat}.

\subsection{Proof of Theorem \ref{subGth}}
It suffices to show the following statements:
\begin{itemize}
\item[--] If $\vx\in \mathbb{R}^{N}$  is a subgaussian vector, then for given $\mW\in \mathbb{R}^{N\times M}$ and $\vb\in \mathbb{R}^{M}$, the random vector $\vy = \mW^\top \vx+\vb$ is subgaussian.
\item[--] If $\vx\in \mathbb{R}^{N}$ is a subgaussian vector, $\vy = \vx^+$ is also subgaussian. 
\end{itemize}
We start by proving the first statement. The subgaussianity of $\vx$ implies that there exists a constant $\kappa$ such that for any given $\balpha\in\mathbb{S}^{N-1}$: 
\begin{equation}\label{expr1.5}\forall t\geq 0 \ : \ \mathbb{P}\left\{ \left|\balpha^\top\vx\right| > t\right\} \leq c\exp\left(  - \frac{t^2}{\kappa^2}\right).
\end{equation}
Now considering $\balpha\in\mathbb{S}^{N-1}$ we have
\begin{align*}
\left| \balpha^\top\vy\right| & \leq \left| \left(\mW\balpha\right)^\top\vx\right|
+\left| \balpha^\top\vb\right|\\ &=  \left\| \mW\balpha\right\|\left| \left(\frac{\mW\balpha}{\left\| \mW\balpha\right\|}\right)^\top\vx\right|
+\left| \balpha^\top\vb\right|\\ &\leq  \left\| \mW\right\|\left| \left(\frac{\mW\balpha}{\left\| \mW\balpha\right\|}\right)^\top\vx\right|
+\left\| \vb\right\|,
\end{align*}
which immediately implies that 
\[\forall \balpha\in\mathbb{S}^{N_\ell-1}: \left\{\vx :  \left|\balpha^\top\left( \mW^\top \vx+\vb\right)\right|>t \right\} \subseteq \left\{ \vx :  \left\| \mW\right\|\left| \left(\frac{\mW\balpha}{\left\| \mW\balpha\right\|}\right)^\top\vx\right|
+\left\| \vb\right\| > t\right\}. 
\]
By the measure comparison we get
\begin{align*}
\mathbb{P}\left\{\left|\balpha^\top\left( \mW^\top \vx+\vb\right)\right|>t \right\} &\leq \mathbb{P}\left\{  \left\| \mW\right\|\left| \left(\frac{\mW\balpha}{\left\| \mW\balpha\right\|}\right)^\top\vx\right|
+\left\| \vb\right\| > t\right\}\\& = \mathbb{P}\left\{ \left| \left(\frac{\mW\balpha}{\left\| \mW\balpha\right\|}\right)^\top\vx\right|
 >  \max\left(\frac{t-\left\| \vb\right\|}{\left\| \mW\right\|},0\right) \right\}\\ &\leq c\exp\left(  - \max\left(\frac{t-\left\| \vb\right\|}{\kappa \left\| \mW\right\|},0\right)^2\right).
\end{align*}
Using Lemma \ref{lemx1} below, for ${\kappa'}^2 \geq \kappa^2 \left\| \mW\right\|^2 + \left\| \vb\right\|^2$ and $c' = ce$, the following should hold:
\[\forall t\geq 0: ~~\mathbb{P}\left\{\left|\balpha^\top\left( \mW^\top \vx+\vb\right)\right|>t \right\} \leq
c'\exp\left(  - \frac{t^2}{{\kappa'}^2}\right), 
\]
which completes the first part of the proof.
\begin{lemma}\label{lemx1}
Fix $a>0$ and $b\geq 0$. Then, for $c^2\geq a^2 + b^2$,
\begin{equation}\label{expr1}\forall t\geq 0 \ : \ \exp\left( -\max\left( \frac{t-b}{a},0\right)^2\right)\leq \exp\left( 1 - \frac{t^2}{c^2}\right).
\end{equation}
\end{lemma}

\textit{Proof:}\\
For $t\leq b$, the proposed conditions require $1 - t^2/c^2>0$, for which \eqref{expr1} automatically holds. In the case of $t>b$, to establish \eqref{expr1} it suffices to show that 
\begin{equation*} \left( \frac{t-b}{a}\right)^2\geq   \frac{t^2}{c^2}-1,
\end{equation*}
or in a simplified form 
\begin{equation}\label{expr2}\left( 1 - \frac{a^2}{c^2}\right)t^2 -2bt+a^2+b^2\geq 0. 
\end{equation}
The discriminant of the quadratic expression in \eqref{expr2} is $4a^2\left( (a^2+b^2)/c^2 - 1\right)$, which is never positive and the expression always takes an identical sign to $1-a^2/c^2$.$\square$

We next show the subgaussianity of $\vx^+$ for a subgaussian random vector $\vx\in\mathbb{R}^N$. For this purpose we introduce a constant $\kappa'$ such that $\mathbb{E}\exp( (\balpha^\top\vx^+)^2/{\kappa'}^2) \leq e$ for all $\balpha\in\mathbb{S}^{N-1}$. To this end, we first bound the magnitude of the marginals as
\begin{equation}\label{recti}(\balpha^\top\vx^+)^2 \leq \|\balpha\|^2\|\vx^+\|^2\leq \|\vx\|^2.
\end{equation}
We now make use of the following lemma borrowed from \cite{hsu2012tail} (see Theorem 2.1 and Remark 2.3 therein).
\begin{lemma}\label{lemkaka}
Let $\mA\in\mathbb{R}^{N\times N}$ be a matrix and $\bsigma = \mA^\top \mA$. Suppose $\vx\in\mathbb{R}^N$ is a random vector such that for some $\bmu\in\mathbb{R}^N$ and $\kappa\geq 0$ 
\begin{align}\label{sg1}
\forall \balpha\in \mathbb{R}^{N}:~~\mathbb{E}\exp\left(\balpha^\top \left( \vx - \bmu \right) \right)\leq \exp\left( \|\balpha\|^2\kappa^2 / 2\right),
\end{align}
then for ${\kappa'}^2\geq 2\kappa^2\|\bsigma\|$,
\[\mathbb{E}\exp\left( \frac{\left\|\mA\vx \right\|^2}{{\kappa'}^2}\right) \leq \exp\left( \kappa^2\mbox{tr}\left(\bsigma\right){\kappa'}^{-2} + \frac{\kappa^4\mbox{tr}\left(\bsigma^2\right){\kappa'}^{-4} + \|\mA\bmu\|^2{\kappa'}^{-2}}{1-2\kappa^2\|\bsigma\|{\kappa'}^{-2}}\right).
\]
\end{lemma}
Condition \eqref{sg1} is technically a certificate of the subgaussianity of $\vx$ \cite{hsu2012tail,vershynin_2012}. Setting $\mA$ in Lemma \ref{lemkaka} to the identity matrix and making use of \eqref{recti} verify that for $\balpha\in\mathbb{S}^{N-1}$ and ${\kappa'}^2\geq 2\kappa^2$, 
\begin{equation}\label{sg2}
\mathbb{E}\exp\left( \frac{(\balpha^\top\vx^+)^2}{{\kappa'}^2}\right)\leq \mathbb{E}\exp\left( \frac{\left\|\vx \right\|^2}{{\kappa'}^2}\right) \leq \exp\left( N\kappa^2{\kappa'}^{-2} + \frac{N\kappa^4{\kappa'}^{-4} + \|\bmu\|^2{\kappa'}^{-2}}{1-2\kappa^2{\kappa'}^{-2}}\right).
\end{equation}
By selecting $\kappa'$ sufficiently large, specifically $\kappa' \gtrsim \max(\sqrt{N}\kappa,\|\bmu\|)$, one can upper bound the right-hand side expression of \eqref{sg2} by $e$.

\subsection{Proof of Theorem \ref{sampComp}}
With reference to \eqref{retrainANeuron}, for $\mX^{in} = [\vx_1,\cdots,\vx_P]\in\mathbb{R}^{N\times P}$ and
\[\Omega = \left\{p:x^{out}_p>0\right\} = \left\{p:\vx_p^\top\vw_0>0\right\},
\]
we need to derive the conditions that $\vw^*$ is the unique solution to
\begin{equation}\label{e1}
\operatorname*{min}_{\vw}\;\;\left\|\vw\right\|_{1}\qquad \mbox{subject to} \qquad \left \{ \begin{array}{l} {\mX^{in}_{:,\Omega}}^\top\vw = \vx^{out}_\Omega \\[.05cm] {\mX^{in}_{:,\Omega^c}}^\top\vw\preceq \boldsymbol{0}\end{array}\right..
\end{equation}
%
%
%
%
For general $\mX,\mX_0\in \mathbb{R}^{N\times P}$ and $\Omega\subseteq\{1,\cdots,P\}$, consider the operator
\begin{equation*}
\mathcal{T}_\Omega^{\mX_{\!0}}\mX \triangleq \mX\mbox{diag}\left( \mathds{1}_\Omega\right)+\mX_{\! 0}, 
\end{equation*}
where $\mathds{1}_\Omega\in\mathbb{R}^P$ is the indicator of the set $\Omega$. Simply, $\mathcal{T}_\Omega^{\boldsymbol{0}}\mX$ replaces columns of $\mX$ indexed by $\Omega^c$ with zero vectors.  Exploiting the notion of minimum conic singular value, we first state a unique optimality result for \eqref{e1}, which generally holds regardless of the specific structure of $\mX^{in}$.

\begin{lemma}\label{lemmOpt}
Fix $\bmu\in\mathbb{R}^N$ and $\sigma\in\mathbb{R}- \{0\}$. Consider $\vw^*\in\mathbb{R}^N$ to be a (sparse) feasible vector for \eqref{e1}, and define the descent cone
\[ \mathcal{D} = \bigcup_{\tau>0}\left\{  \begin{pmatrix}\vy\\ z\end{pmatrix} \in\mathbb{R}^{N+1}: \|\vw^*+\tau\vy\|_1\leq \| \vw^*\|_1\right\}.
\]
For $\boldsymbol{\Upphi} = \mathcal{T}_\Omega^{\;-\bmu\boldsymbol{1}^\top}\!\mX^{in}$, if 
\begin{equation}\label{eqq3.1}
\inf~ \left\{  \left\|\begin{pmatrix} \boldsymbol{\Upphi}^\top & \sigma\boldsymbol{1}\end{pmatrix}\vv\right\|:~ \vv\in \mathcal{D}\cap \mathbb{S}^N\right\} >0,
\end{equation}
then $\vw^*$ is the unique solution to \eqref{e1}.
\end{lemma}

\textit{Proof:}\\
Showing the following three statements would complete the proof:
\begin{enumerate}
\item[(S.1)] If $\vw^*$ is feasible for \eqref{e1}, then the pair $(\vw^*,\sigma^{-1}\bmu^\top\vw^*)$ is feasible for the convex program:\vspace{-.2cm}
\begin{equation}
\label{e4}
\minimize_{(\vw,u)} \;\;\|\vw\|_1\qquad \mbox{subject to} \qquad  \begin{pmatrix} \boldsymbol{\Upphi}^\top & \sigma\boldsymbol{1}\end{pmatrix}\begin{pmatrix}  \vw\\ u \end{pmatrix}= \begin{pmatrix}  \vx^{out}_\Omega\\ \boldsymbol{0} \end{pmatrix}.\vspace{-.2cm}
\end{equation}
\item[(S.2)] For any pair $(\vw^*,u^*)$ that is feasible for \eqref{e4}, if condition \eqref{eqq3.1} holds, then $(\vw^*,u^*)$ is the unique solution to \eqref{e4}.
\item[(S.3)] If $(\vw^*,\sigma^{-1}\bmu^\top\vw^*)$ is the unique solution to \eqref{e4}, then $\vw^*$ is the unique solution to \eqref{e1}.
\end{enumerate}

Based on the definition $\boldsymbol{\Upphi} = \mathcal{T}_\Omega^{\;-\bmu\boldsymbol{1}^\top}\!\mX^{in}$, verifying (S.1) is trivial. Claim (S.2) is a direct application of the minimum conic singular value result (e.g., see Prop. 2.2 of \cite{chandrasekaran2012convex}, or Prop. 2.6 of \cite{tropp2015convex}). To prove (S.3), suppose under the proposed assumption, \eqref{e1} has a different solution $\hat\vw$, where $\|\hat\vw\|_1\leq \|\vw^*\|_1$. Then (S.1) requires $(\hat\vw, \sigma^{-1}\bmu^\top\hat\vw)$ to be feasible for \eqref{e4}. However the objective for this feasible point is less than $\|\vw^*\|_1$, which is in contradiction with $(\vw^*,\sigma^{-1}\bmu^\top\vw^*)$ being the unique solution to \eqref{e4}.$\square$

Using Lemma \ref{lemmOpt} and the bowling scheme sketched in \cite{tropp2015convex}, we continue with lower-bounding the minimum conic singular value away from zero, and relating the conditions to the number of samples, $P$.

To this end, we may look into the structure of the matrix $\boldsymbol{\Upphi}$ in Lemma \ref{lemmOpt} as being populated with independent copies of $\vx 1_{\vw_0^\top \vx>0}-\bmu$ as the columns, and exploit the independence required for the bowling scheme. To assure centered columns, we choose $\bmu = \mathbb{E}\vx 1_{\vw_0^\top \vx>0}$, making columns of $\boldsymbol{\Upphi}=[\bphi_1,\cdots,\bphi_P]$ independent copies of the centered subgaussian\footnote{Since $| \balpha^\top \vx 1_{\vw_0^\top \vx>0} |\leq | \balpha^\top \vx |$ and for $t\geq0$, $\mathbb{P}\{| \balpha^\top \vx 1_{\vw_0^\top \vx>0} |>t\}\leq \mathbb{P}\{| \balpha^\top \vx |>t\}$, \eqref{subg1} confirms that $\vx$ being subgaussian implies $\vx 1_{\vw_0^\top \vx>0}$ to be subgaussian.} random vector
\[\bphi \triangleq  \vx 1_{\vw_0^\top \vx>0} - \mathbb{E}\vx 1_{\vw_0^\top \vx>0}.
\]
For reasons that become apparent later in the proof, our arbitrary choice of $\sigma$ in Lemma \ref{lemmOpt} is narrowed to 
\[\sigma_0 \triangleq \sqrt{2}\|\bphi\|_{\psi_2}.
\]
In a random setting, to lower-bound the minimum conic singular value, we adapt the following result from \cite{tropp2015convex}, Prop. 5.1 (or see Theorem 5.4 of \cite{mendelson2014learning} for the original statement). 
\begin{theorem}\label{MainBowl}
Fix a set $E\subset\mathbb{R}^d$. Let $\vphi$ be a random vector on $\mathbb{R}^d$, and let $\vphi_1,\cdots,\vphi_P$ be independent copies of $\vphi$. For $\xi\geq 0$, suppose the marginal tail relation below holds:
\[\inf_{ \vv\in E} \mathbb{P} \left\{\left| \vphi^\top \vv \right|\geq \xi\right\} \geq C_\xi >0.
\]
Let $\varepsilon_1,\cdots,\varepsilon_P$ be independent Rademacher random variables,
independent from everything else, and define the
mean empirical width of the set $E$:
\begin{equation}\label{meanEwidth}
\mathcal{W}_P\left( E;\vphi\right) \triangleq \mathbb{E}~\sup_{\vv\in E} \langle \vh,\vv\rangle, \quad\mbox{where}\quad \vh = \frac{1}{\sqrt{P}}\sum_{p=1}^P \varepsilon_p\vphi_p.
\end{equation}
Then, for any $\xi>0$ and $t>0$, with probability at least $1-\exp(-t^2/2)$:
\begin{equation}\label{MCSV}
\inf_{\vv\in E}~ \left(  \sum_{p=1}^P\left(\vphi_p^\top\vv\right)^2\right)^{\frac{1}{2}} \geq \frac{\xi}{2} C_{\xi} \sqrt{P}-2\mathcal{W}_P(E;\boldsymbol{\phi})-\frac{\xi}{2} t.
\end{equation}
\end{theorem}

For a more compact (and inline) notation, we use the following notation for the concatenation of a vector $\vw$ and a scalar $u$, 
\[\vw^\frown u \triangleq \begin{pmatrix} \vw \\ u \end{pmatrix}.
\]
Also, for a given objective and point $\vv_0\in\mathbb{R}^d$, we denote the descent cone by
\[\mathpzc{D}_{\vv}\left( f(\vv);\vv_0\right) = \bigcup_{\tau>0}\left\{ \vy\in\mathbb{R}^{d}: f(\vv_0+\tau \vy)\leq f(\vv_0)\right\}.
\]
To show that condition \eqref{eqq3.1} holds for the prescribed $s$-sparse vector $\vw^*$, we will show that for sufficiently large $P$, the right-hand side expression in \eqref{MCSV} can be bounded away from zero. To apply Theorem \ref{MainBowl} to our problem in \eqref{eqq3.1}, the random vector $\vphi$ and the set $E$ to consider are
\[\vphi = \bphi^\frown\sigma_0, \quad\mbox{and} \quad E = \mathpzc{D}_{\vw^\frown u}\left( \|\vw\|_1; {\vw^*}^\frown u^* \right)\cap \mathbb{S}^{N},
\]
where
\begin{align*}
\mathpzc{D}_{\vw^\frown u}\left( \|\vw\|_1; {\vw^*}^\frown u^* \right) &= \bigcup_{\tau>0}\left\{  \vy^\frown z \in\mathbb{R}^{N+1}: \|\vw^*+\tau\vy\|_1\leq \| \vw^*\|_1\right\}\\ &= \mathpzc{D}_{\vw}\left( \|\vw\|_1; \vw^*\right) \times \mathbb{R},
\end{align*}
and $u^* = \sigma_0^{-1}\left(\mathbb{E}\vx 1_{\vw_0^\top \vx>0}\right)^\top\vw^*$. 
Note that in the formulation above, $\mathpzc{D}_{\vw^\frown u}(.;.)\subset \mathbb{R}^{N+1}$, while $\mathpzc{D}_{\vw}(.;.)\subset \mathbb{R}^{N}$. The remainder of the proof focuses on bounding the contributing terms on the right-hand side expression of \eqref{MCSV}. 

\subsubsection{Bounding the Mean Empirical Width} 
In this section of the proof, we aim to upper-bound 
\[\mathcal{W}_P\left( \mathpzc{D}_{\vw^\frown u}\left( \|\vw\|_1; {\vw^*}^\frown u^* \right)\cap \mathbb{S}^{N};\bphi^\frown\sigma_0\right),
\]
where following the formulation in \eqref{meanEwidth} we have
\[\vh = \frac{1}{\sqrt{P}}\sum_{p=1}^P \varepsilon_p \boldsymbol{\varphi}_p^\frown \sigma_0  =  \underbrace{\begin{pmatrix}  \boldsymbol{0}\\ \frac{\sigma_0}{\sqrt{P}}\sum_{p=1}^P \varepsilon_p \end{pmatrix}}_{\vh_u}  + \underbrace{\begin{pmatrix}  \frac{1}{\sqrt{P}}\sum_{p=1}^P \varepsilon_p\boldsymbol{\varphi}_p\\ 0 \end{pmatrix}}_{\vh_{\vw}^\frown 0}.
\]
Using the compact notations
\[\mathcal{K}_{\vw^*, u^*} = \mathpzc{D}_{\vw^\frown u}\left( \|\vw\|_1;  {\vw^*}^\frown u^* \right), \quad\mbox{and}\quad \mathcal{K}_{\vw^*} = \mathpzc{D}_{\vw}\left( \|\vw\|_1; \vw^*\right),
\]
ones has
\begin{align}\nonumber
\mathcal{W}_P\left( \mathcal{K}_{\vw^*, u^*}\cap \mathbb{S}^{N};\bphi^\frown\sigma_0\right) &= \mathbb{E}\sup_{\vv\in \mathcal{K}_{{\vw^*}, u^*}\cap \mathbb{S}^N} \langle \vh,\vv\rangle\\ & \leq \mathbb{E}\sup_{\vv\in \mathcal{K}_{{\vw^*}, u^*}\cap \mathbb{S}^N} \langle \vh_u,\vv\rangle + \mathbb{E}\sup_{\vv\in \mathcal{K}_{{\vw^*}, u^*}\cap \mathbb{S}^N} \left \langle \vh_{\vw}^\frown 0,\vv\right\rangle.\label{b2}
\end{align}
For the first term in \eqref{b2} note that
\begin{align}\nonumber 
\mathbb{E}\sup_{\vw^\frown u\in \mathcal{K}_{\vw^*,u^*}\cap \mathbb{S}^n} \langle \vh_u,\vw^\frown u\rangle &= \mathbb{E}\sup_{\vw^\frown u\in \mathcal{K}_{\vw^*,u^*}\cap \mathbb{S}^N} \left(\frac{\sigma_0}{\sqrt{P}}\sum_{p=1}^P \varepsilon_p  \right)u\\ \nonumber & =  \mathbb{E} \left| \frac{\sigma_0}{\sqrt{P}}\sum_{p=1}^P \varepsilon_p \right|\\ \nonumber &\leq \frac{\sigma_0}{\sqrt{P}} \left( \mathbb{E} \left(\sum_{p=1}^P\varepsilon_p\right)^2\right)^{\frac{1}{2}}\\&=\sigma_0.\label{b2.5}
\end{align}
To bound the second term in \eqref{b2}, we proceed by first showing that for any fixed $\vh_{\vw}\in\mathbb{R}^N$,
\begin{equation}\label{b3}
\sup_{\vw^\frown u \in \mathcal{K}_{\vw^*,u^*}\cap \mathbb{S}^N} \left \langle \vh_{\vw}^\frown 0,\vw^\frown u \right\rangle = \sup_{\vw\in \mathcal{K}_{\vw^*}\cap \mathbb{S}^{N-1}} \langle \vh_{\vw},\vw\rangle.
\end{equation}
For this purpose only the following two cases need to be considered:

-- \textbf{case 1:} $\langle \vh_{\vw},\vw\rangle\leq 0, \forall \vw\in \mathcal{K}_{\vw^*}.$\\
In this case the supremum value for both sides of \eqref{b3} is zero, which may be attained by picking $\vw=\boldsymbol{0}$ and $u=1$.  

-- \textbf{case 2:} $\exists \vw\in \mathcal{K}_{\vw^*}$, such that $ \langle\vh_{\vw},\vw\rangle> 0.$\\
To show the equality in this case, we only need to show that if $\hat\vv = \hat\vw^\frown \hat u$ is a point at which the (positive) supremum is attained, i.e.,
\[\sup_{\vv\in \mathcal{K}_{\vw^*,u^*}\cap \mathbb{S}^N} \left \langle \vh_{\vw}^\frown 0,\vv\right\rangle = \left \langle \vh_{\vw}^\frown 0,\hat \vv\right\rangle = \left \langle \vh_{\vw},\hat \vw\right\rangle,
\]
then we must have $\hat u=0$. If $\hat u\neq 0$, then the condition $\hat\vv\in \mathbb{S}^N$ requires that $\|\hat\vw\|<1$. In this case the alternative feasible point $\tilde \vv = \|\hat\vw\|^{-1}{\hat\vw}^\frown 0$ produces a greater inner product:
\[\left \langle \vh_{\vw}^\frown 0,\tilde \vv\right\rangle = \frac{1}{\|\hat\vw\|}\left \langle \vh_{\vw},\hat \vw\right\rangle > \left \langle \vh_{\vw},\hat \vw\right\rangle,
\]
which cannot be possible. Therefore $\hat u=0$, and for both sides of \eqref{b3} the supremum value is $\left \langle \vh_{\vw},\hat \vw\right\rangle$. Combining cases 1 and 2 establishes the claim in \eqref{b3}. 

Now, employing \eqref{b3} and \eqref{b2.5} in \eqref{b2} certifies that for 
\[\vh_{\vw} = \frac{1}{\sqrt{P}}\sum_{p=1}^P \varepsilon_p\boldsymbol{\varphi}_p,
\]
and some absolute constant $C$, one has
\begin{align}\nonumber
\mathcal{W}_P\left( \mathcal{K}_{\vw^*, u^*}\cap \mathbb{S}^{N};\bphi^\frown\sigma_0\right) &\leq \sigma_0 + \mathbb{E}\sup_{\vw\in \mathcal{K}_{\vw^*}\cap \mathbb{S}^{N-1}} \langle \vh_{\vw},\vw\rangle\\ &\leq C\|\bphi\|_{\psi_2}\left(\sqrt{s\log\left( \frac{N}{s}\right)+s}+1\right).\label{WpmeanWidth}
\end{align}
The last line in \eqref{WpmeanWidth} is thanks to the following inequality (see \S 6.6 of \cite{tropp2015convex}), which relates the mean empirical width of a centered subgaussian random vector $\bphi$ to the Gaussian width:
\[\mathbb{E}\sup_{\vw\in \mathcal{K}_{\vw^*}\cap \mathbb{S}^{N-1}} \langle \vh_{\vw},\vw\rangle \lesssim \|\bphi\|_{\psi_2} \mathbb{E}\sup_{ \substack{\vw\in \mathcal{K}_{\vw^*}\cap \mathbb{S}^{N-1}\\ \vg\sim \mathcal{N}(\boldsymbol{0},\mI)}   } \langle \vg,\vw\rangle\lesssim \|\bphi\|_{\psi_2}\sqrt{s\log\left( \frac{N}{s}\right)+s}.
\]

\subsubsection{Relating the Marginal Tail Bound and the Virtual Covariance}
As the next step in lower-bounding the right-hand side expression in \eqref{MCSV}, noting that
\begin{equation}\label{eqinftriv}\inf_{ \vv\in \mathcal{K}_{\vw^*, u^*}\cap\mathbb{S}^N} \mathbb{P} \left\{\left| \vv^\top \bphi^\frown \sigma_0 \right|\geq \xi\right\}\geq \inf_{ \vv\in \mathbb{S}^N} \mathbb{P} \left\{\left| \vv^\top \bphi^\frown \sigma_0 \right|\geq \xi\right\},
\end{equation}
in this section we focus on lower bounding the right-hand side expression in \eqref{eqinftriv} in terms of $\|\bphi\|_{\psi_2}$ and the minimum eigenvalue of the virtual covariance matrix. To this end, using the notation 
\[\tilde\lambda_{\min} \triangleq \lambda_{\min}\left(\operatorname{cov}\left(\boldsymbol{\upsilon}\right)\right) = \lambda_{\min}\left(\mathbb{E}\bphi\bphi^\top\right),
\]
one has
\begin{equation}\label{eqsig0}\mathbb{E} \bphi^\frown \sigma_0 {\bphi^\frown \sigma_0 }^\top = \begin{pmatrix} \mathbb{E}\bphi\bphi^\top &\boldsymbol{0}\\ \boldsymbol{0}^\top& \sigma_0^2 \end{pmatrix}\succeq \min\left( \tilde\lambda_{\min},\sigma_0^2\right)\mI.
\end{equation}
On the other hand, from the subgaussian properties of $\bphi$ we have
\begin{align*}
\|\bphi\|_{\psi_2} \geq \sup_{\vw\in\mathbb{S}^{N-1}} 2^{-\frac{1}{2}}\left ( \mathbb{E}\left|\vw^\top\bphi \right|^2\right)^{\frac{1}{2}} \geq \inf_{\vw\in\mathbb{S}^{N-1}} 2^{-\frac{1}{2}}\left ( \mathbb{E}\left|\vw^\top\bphi \right|^2\right)^{\frac{1}{2}} = \sqrt{\frac{\tilde\lambda_{\min}}{2}},
\end{align*}
which simply implies that $\sigma_0^2\geq\tilde\lambda_{\min}$ and combining with \eqref{eqsig0} yields
\begin{equation}\label{b6}\inf_{\vv\in \mathbb{S}^N}\mathbb{E} \left| \vv^\top \bphi^\frown \sigma_0 \right|^2\geq \min\left( \tilde\lambda_{\min},\sigma_0^2\right) = \tilde\lambda_{\min}.
\end{equation}
Considering a positive random variable $\chi$ and a fixed $\xi\geq 0$, we can derive a variant of the Paley-Zygmund inequality by writing $\chi^2 = \chi^2 1_{\left\{\chi^2<\xi^2\right\}}+\chi^2 1_{\left\{\chi^2\geq \xi^2\right\}}$, which using the H\"{o}lder's inequality naturally yields
\begin{align*}
\mathbb{E}\chi^2 \leq \xi^2 + \left(\mathbb{P}\left\{ \chi\geq \xi \right\}\right)^{\frac{\beta}{1+\beta}}\left(\mathbb{E}\chi^{2(1+\beta)}\right)^\frac{1}{1+\beta},\quad \beta>0.
\end{align*}
Subsequently, selecting $\xi\in [0,\sqrt{ \tilde\lambda_{\min}}]$ warrants that
\begin{align*}
\forall \vv\in \mathbb{S}^N: \qquad \mathbb{P} \left\{\left| \vv^\top \bphi^\frown \sigma_0 \right|\geq \xi\right\}\geq \left(\frac{\tilde\lambda_{\min}-\xi^2 }{\left(\mathbb{E} \left| \vv^\top \bphi^\frown \sigma_0 \right|^{2(1+\beta)}\right)^{\frac{1}{\beta+1}}}\right)^{1+\frac{1}{\beta}}.
\end{align*}
We can also use the subgaussian properties of $\bphi$ to bound the denominator as follows
\begin{align*}
\forall \vw^\frown u\in \mathbb{S}^N, ~\alpha\geq 1: \quad \left(\mathbb{E} \left| {\vw^\frown u}^\top \bphi^\frown \sigma_0 \right|^{\alpha}\right)^{\frac{1}{\alpha}} &= \left(\mathbb{E} \left| \vw^\top \bphi + \sigma_0 u \right|^{\alpha}\right)^{\frac{1}{\alpha}}\\ & \leq \left(\mathbb{E} \left| \vw^\top \bphi  \right|^{\alpha}\right)^{\frac{1}{\alpha}} + \sigma_0 |u|\\& =  \|\vw\|\left(\mathbb{E} \left| \frac{\vw^\top}{\|\vw\|} \bphi  \right|^{\alpha}\right)^{\frac{1}{\alpha}} + \sigma_0 |u|\\&\leq \sqrt{\alpha}\|\bphi\|_{\psi_2} \|\vw\| + \sqrt{2}\|\bphi\|_{\psi_2}|u|\\[.1cm] &\leq \sqrt{\alpha+2}\|\bphi\|_{\psi_2},
\end{align*}
where the first inequality is a direct application of the Minkowski inequality, the second inequality uses the subgaussian definition \eqref{b7} and the last bound is thanks to the Cauchy-Schwarz inequality. As a result for $\xi\in [0,\sqrt{\tilde \lambda_{\min}}]$
\begin{equation}\label{b8}
\inf_{ \vv\in \mathbb{S}^N} \mathbb{P} \left\{\left| \vv^\top \bphi^\frown \sigma_0 \right|\geq \xi\right\}\geq \left(\frac{\tilde\lambda_{\min}-\xi^2 }{2(2+\beta)\|\bphi\|_{\psi_2}^2 }\right)^{1+\frac{1}{\beta}}.
\end{equation}
\subsubsection{Combining the Bounds}
We can now combine the bounds \eqref{WpmeanWidth} and \eqref{b8}, and use Theorem \ref{MainBowl} to state that with probability at least $1-\exp(-t^2/2)$:
\begin{align*}\inf_{ \vv\in \mathcal{K}_{\vw^*, u^*}\cap\mathbb{S}^N} ~ \left(  \sum_{p=1}^P\left(\bphi_p^\frown\sigma_0^\top\vv\right)^2\right)^{\frac{1}{2}}  &\geq \frac{\xi}{2}  \sqrt{P}\left(\frac{\tilde\lambda_{\min}-\xi^2 }{2(2+\beta)\|\bphi\|_{\psi_2}^2}\right)^{1+\frac{1}{\beta}} \\& ~~~~ - 2C\|\bphi\|_{\psi_2}\left(\sqrt{s\log\left( \frac{N}{s}\right)+s}+1\right)-\frac{\xi}{2} t.
\end{align*}
Selecting $\xi = \sqrt{\tilde\lambda_{\min}}/3$ would bound the expression above away from zero, as long as
\begin{equation}\label{b9}
P\geq \frac{36\left(9\left(1+\frac{\beta}{2} \right)\|\bphi\|_{\psi_2}^2\right)^{2+\frac{2}{\beta}}}{\tilde\lambda_{\min}\left( 2\tilde\lambda_{\min} \right)^{2+\frac{2}{\beta}} }\left( 2C\|\bphi\|_{\psi_2}\left(\sqrt{s\log\left( \frac{N}{s}\right)+s}+1\right)+\frac{\sqrt{\tilde\lambda_{\min}}}{6} t\right)^2.
\end{equation}
Noting that $\tilde \lambda_{\min}\leq 2\|\bphi\|_{\psi_2}^2$ and using the basic inequality $(a+b)^2\leq 2a^2+2b^2$ twice yields 
\[\left( 2C\|\bphi\|_{\psi_2}\left(\sqrt{s\log\left( \frac{N}{s}\right)+s}+1\right)+\frac{\sqrt{\tilde\lambda_{\min}}}{6} t\right)^2 \lesssim \|\bphi\|_{\psi_2}^2\left(s\log\left( \frac{N}{s}\right)+s+1+\frac{t^2}{72C^2} \right).
\]
Also, since $(1+\frac{\beta}{2})^{\frac{2}{\beta}}\lesssim 1$ for $\beta\geq 1$, one has
\[\frac{36\left(9\left(1+\frac{\beta}{2} \right)\|\bphi\|_{\psi_2}^2\right)^{2+\frac{2}{\beta}}}{\tilde\lambda_{\min}\left( 2\tilde\lambda_{\min} \right)^{2+\frac{2}{\beta}} }\lesssim \left (1+\frac{\beta}{2}\right)^2 \frac{\|\bphi\|_{\psi_2}^{4+\frac{4}{\beta}}}{\tilde\lambda_{\min}^{3+\frac{2}{\beta}}}.
\]
Therefore, the desired condition in \eqref{eqq3.1} holds, as long as
\[P \gtrsim \left (1+\frac{\beta}{2}\right)^2 \frac{\|\bphi\|_{\psi_2}^{6+\frac{4}{\beta}}}{\tilde\lambda_{\min}^{3+\frac{2}{\beta}}}\left(s\log\left( \frac{N}{s}\right)+s+1+\frac{t^2}{72C^2} \right).
\]
Finally, setting $\beta' = \beta/2$ and $t' = t^2/(72C^2)$ yields the advertised claim in \eqref{PsampleCompRate}.

\subsection{Proof of Lemma \ref{LemInt}}
We follow a similar line of argument as \S 5.3.1 of \cite{louart2017random}. To evaluate 
\[I = \mathbb{E}_{\vx}~ g\left(\balpha^\top\vx\right)1_{\bbeta^\top\vx>0} = \frac{1}{(2\pi)^{\frac{N}{2}}}\int_{\bbeta^\top\vx>0} g\left(\balpha^\top\vx\right)  \exp\left( -\frac{\left\|\vx\right\|^2}{2}\right) \mbox{d}\vx,
\]
we assume that $\balpha$ and $\bbeta$ are not aligned (for the aligned case a similar procedure applies to merely $\balpha$). We consider the unitary matrix $\mE = [\ve_1,\cdots,\ve_N]$, where  
\[\ve_1 = \bbeta, \qquad \ve_2 = \frac{\balpha - \left(\bbeta^\top\balpha\right)\bbeta}{\sqrt{1-\left(\bbeta^\top\balpha\right)^2}},
\]  
and $\ve_3,\cdots \ve_N$ are any completion of the ortho-basis. Setting $\vx = \mE\vz$ yields $\bbeta^\top\vx=z_1$ and 
$$\balpha^\top\vx =\left(\bbeta^\top\balpha\right) z_1 +  \sqrt{1- \left(\bbeta^\top\balpha\right)^2} z_2. $$
Taking into account the injectivity of the linear map $\mE$, we can reformulate the integral in the $\vz$-domain as (see Theorem 263D of \cite{fremlin2000measure} for the formal statement)
\[I = \frac{1}{2\pi}\int_{z_1>0}  g\left(  \left(\bbeta^\top\balpha\right) z_1 +  \sqrt{1- \left(\bbeta^\top\balpha\right)^2} z_2\right)\exp\left( -\frac{z_1^2+z_2^2}{2}\right) \mbox{d}z_1\mbox{d}z_2.
\]

\noindent \textbf{Acknowledgement:} A. Aghasi would like to thank Roman Vershynin and Richard Kueng for the insightful suggestions and communications.

\section{Supplementary Materials}
\subsection{Net-Trim for Convolutional Layers}\label{sec:convimp}
When $\mX^{in}\in \mathbb{T}_{in}$, $\mW\in \mathbb{T}_{w}$ and $\mX^{out}\in \mathbb{T}_{out}$ are tensors, and $\Omega$ indicates a subset of the tensor elements, our central program takes the following form:
\begin{equation}\label{eqop1}
\minimize_{\mW}~\left\|\mW\right\|_1\quad \mbox{subject to} \quad  \left\{\begin{array}{l}\left\|\left(\AX(\mW) - \mX{}^{out} \right)_\Omega\right \|_F\leq \epsilon  \\[.1cm] \left(\AX(\mW)\right )_{\Omega^c} \leq \mV_{\Omega^c} \end{array}\right.,
\end{equation}
where $\|\cdot\|_1$ and $\|\cdot\|_F$ naturally apply to the vectorized tensors. As before, for a given tensor $\mZ$, $\mZ_\Omega$ is a tensor of similar size, which takes identical values as $\mZ$ on $\Omega$ and zero values on $\Omega^c$.

The operator $\AX:\mathbb{T}_w\to\mathbb{T}_{out}$ is a linear operator that is parameterized by $\mX^{in}$. For instance in convolutional layers it is a tensor convolution operator with one of the operands being $\mX^{in}$. Throughout the text we assume that $\AX^*:\mathbb{T}_{out}\to \mathbb{T}_w$ is the adjoint operator. The adjoint operator needs to satisfy the following property:
\[\forall \mW\in\mathbb{T}_w, \forall \mZ\in\mathbb{T}_{out}:~~~\langle\AX(\mW),\mZ \rangle_{\mathbb{T}_{out}} = \langle\mW ,\AX^*(\mZ)\rangle_{\mathbb{T}_{w}}.
\]

Going through an identical line of argument as Section \ref{app:ADMM} yields similar ADMM steps, only different in the way that $\mX^{in}$ interacts with $\mW^{(3)}$. More specifically,  
\begin{align}\label{admmprog1op}
\mW^{(1)}_{k+1} &= \argmin_{\mW}~ f_1\left(\mW\right) + \frac{\rho}{2}\left\|\mW + \mU^{(1)}_k - \AX\left(\mW^{(3)}_{k}\right) \right\|^2_F,
\\
\label{admmprog2op} \mW^{(2)}_{k+1} &= \argmin_{\mW}~ f_2\left(\mW\right) + \frac{\rho}{2}\left\|\mW + \mU^{(2)}_k -  \mW^{(3)}_k \right\|^2_F,
\\ 
\label{admmprog3op} \mW^{(3)}_{k+1} &= \argmin_{\mW}~ \frac{\rho}{2}\left\|\mW^{(1)}_{k+1} + \mU^{(1)}_k - \AX\left(\mW\right) \right\|^2_F + \frac{\rho}{2}\left\|\mW^{(2)}_{k+1} + \mU^{(2)}_k -  \mW \right\|^2_F,
\end{align}
and the dual updates are performed via
\begin{align*}
\mU^{(1)}_{k+1}= \mU^{(1)}_{k} + \mW^{(1)}_{k+1} - \AX\left(\mW^{(3)}_{k}\right) ,~~~ \mU^{(2)}_{k+1} = \mU^{(2)}_{k} + \mW^{(2)}_{k+1} - \mW^{(3)}_{k+1}.
\end{align*}
Based on the steps above, the following algorithm is a straightforward modification of the original Net-Trim implementation presented in operator form. 

\begin{algorithm}[H]\small 
\centerline{\caption{Implementation of the Net-Trim for Convolutional Layers}}

\begin{algorithmic}
\STATE{\textbf{input:} $\mX^{in}\in\mathbb{T}_{in}$, $\mX^{out}\in\mathbb{T}_{out}$, $\Omega$, $\mV_\Omega$, $\epsilon$, $\rho$}

\STATE{\textbf{initialize}: $\mU^{(1)}\in\mathbb{T}_{out}, \mU^{(2)}\in\mathbb{T}_{w}$ and $\mW^{(3)}\in\mathbb{T}_{w}$} \qquad   \texttt{\% all initializations can be with $\boldsymbol{0}$}\vspace{.04cm}

\WHILE{not converged}
\STATE{$\mY\leftarrow \AX\left(\mW^{(3)}\right)-\mU^{(1)}$}
 \IF{$\left \|\mY_\Omega - \mX^{out}_\Omega\right \|_F\leq \epsilon$} 
 \STATE {$\mW^{(1)}_\Omega\leftarrow \mY_\Omega$} 
 \ELSE
 \STATE{$\mW^{(1)}_\Omega\leftarrow \mX^{out}_\Omega+\epsilon\left\|\mY_\Omega - \mX^{out}_\Omega\right\|_F^{-1}\left(\mY_\Omega - \mX^{out}_\Omega\right)$} 
 \ENDIF
 \STATE{$\mW^{(1)}_{\Omega^c}\leftarrow  \mY_{\Omega^c} - (  \mY_{\Omega^c} -  \mV_{\Omega^c} )^+$}
\STATE{ $\mW^{(2)}\leftarrow S_{1/\rho}( \mW^{(3)} - \mU^{(2)} )$\quad\qquad \qquad \texttt{~~~ \% $S_{1/\rho}$  applies to each element of the matrix}}\vspace{-.0cm}
\STATE{  $\mW^{(3)}\leftarrow \argmin_{\mW}~ \frac{1}{2}\left\|  \AX\left(\mW\right) - \left(\mW^{(1)} + \mU^{(1)}\right) \right\|^2_F + \frac{1}{2}\left\|\mW- \left(\mW^{(2)} + \mU^{(2)}\right) \right\|^2_F$
}\vspace{-.0cm}
\STATE{ $\mU^{(1)} \leftarrow \mU^{(1)} + \mW^{(1)} - \AX\left(\mW^{(3)}\right)$}
\STATE{$\mU^{(2)} \leftarrow \mU^{(2)} + \mW^{(2)} - \mW^{(3)}$}
\ENDWHILE
\RETURN $\mW^{(3)}$

\end{algorithmic}
\end{algorithm}
\noindent The only undiscussed part in this algorithm is the update for $\mW^{(3)}$, which we address using an operator form of the conjugate gradient algorithm.  

\subsubsection{Least Squares Update Using an Operator Conjugate Gradient}

In this section we address the minimization 
\begin{equation}\label{eqop2}
\minimize_{\mW\in\mathbb{T}_w}~ \frac{1}{2}\left\|  \AX\left(\mW\right) - \mB \right\|^2_F + \frac{1}{2}\left\|\mW- \mC \right\|^2_F,
\end{equation}
which is central to the $\mW^{(3)}$ update in \eqref{admmprog3op}. Here  $\mB\in\mathbb{T}_{out}$ and $\mC\in\mathbb{T}_w$ are tensors, and $\AX:\mathbb{T}_w\to\mathbb{T}_{out}$ is . The minimizer to \eqref{eqop2} can be found by taking a derivative and setting it to zero, i.e.,
\begin{equation}
\AX^*\left( \AX\left(\mW\right) - \mB  \right) + \mW -\mC = \boldsymbol{0},
\end{equation}
or
\begin{equation}\label{eqop3}
\AX^*\left( \AX\left(\mW\right) \right) + \mW  =\AX^*\left(\mB\right) +\mC.
\end{equation}
Solving \eqref{eqop3} for $\mW$ is efficiently possible via the method of conjugate gradient. The following algorithm outlines the process of solving \eqref{eqop3}, which is a variant of the original CG algorithm (e.g., see \cite{CGLink}) reformulated in operator form.

\begin{algorithm}[H]\small 
\centerline{\caption{Least Squares Update in the Net-Trim Using Conjugate Gradient}}
\begin{algorithmic}

\STATE{\textbf{initialize}: $\mW_0=\boldsymbol{0}$; $~~\mR_0 = \AX^*\left(\mB\right) +\mC$; $~~\mP_0 = \mR_0$} \vspace{.04cm}

\FOR{$k = 1,\ldots,K_{\max}$} 
\STATE{$\mT_{k-1} = \AX\left(\mP_{k-1}\right)$}
\STATE{$\alpha_k = \frac{\left\| \mR_{k-1}\right\|_F^2}{\left\| \mT_{k-1} \right\|_F^2 + \left\|\mP_{k-1}  \right\|_F^2}$}\\[.1cm]
\STATE{$\mW_k = \mW_{k-1} + \alpha_k\mP_{k-1}$}
\STATE{$\mR_k = \mR_{k-1} - \alpha_k\left( \AX^*\left( \mT_{k-1}\right) + \mP_{k-1}\right)$}
\STATE{$\beta_k = \frac{\|\mR_k\|_F^2}{\|\mR_{k-1} \|_F^2}$}\\[.1cm]
\STATE{$\mP_{k} = \mR_k + \beta_k\mP_{k-1}$}
\ENDFOR
\STATE{\textbf{Output:} $\mW_{K_{\max}}$}

\end{algorithmic}
\end{algorithm}

\subsection{Experiments}\label{sec:exp}
In this section, we present more details of the experimental setup and provide additional simulations which were excluded from the original manuscript due to space limitation.

In our first set of experiments, we presented a comparison between the cascade and parallel frameworks. As stated, for this purpose we use the FC network of size $784\times 300 \times 1000 \times 100 \times 10$ (composed of four layers: $\mW_1\in\mathbb{R}^{784\times 300}, \mW_2\in\mathbb{R}^{300\times 1000}$, etc), trained to classify the MNIST dataset. Panels (a) and (b) in Figure \ref{fig2m} of the paper summarize the outcome of applying the Net-Trim parallel scheme to the trained FC model. By varying the value of $\epsilon$, one may explore different levels of layer sparsity and discrepancy. Panel (a) reports the relative value of the overall discrepancy as a function of the relative sparsity at each layer (i.e., percentage of zeros in $\hat\mW_\ell$). Each plot is obtained by varying $\epsilon$ for a range of values and retraining the FC model with 10K, 20K, 30K and the entire 55K training samples. As expected, allowing more discrepancy improves the level of sparsity. Since practically the overall discrepancy is not a good indication of the changes in the model accuracy, in panel (b) we replace it with the test accuracy of the retrained models. An interesting observation is that retraining the models with fewer samples does not significantly degrade the test accuracies and even in some cases (e.g., 30K versus 55K) it causes a slight improvement in the accuracy of the retrained models.  Panels (c) and (d) report a similar set of experiments for the cascade Net-Trim, where increasing the inflation rate away from one allows producing sparser networks.

Employing the parallel scheme (thanks to its distributable nature), and the use of a subset of the training data in the Net-Trim retraining process are both computationally attractive paths, and the experiments in Figure \ref{fig2m} indicate that at least for a reasonable sparsity range, they could be both explored without much degradation of the model accuracies. In the remainder of the experiments in this section, we will consistently use the parallel scheme for our retraining purposes, and will no more reference to the Net-Trim parallel or cascade nature.

In the next set of experiments, we investigate the additional pruning that Net-Trim brings to the architecture of neural networks beyond Dropout and $\ell_1$ regularization. For this purpose we consider the application of an $\ell_1$ regularization, Dropout and a combination of both to the training of our standard FC model. We also apply a similar set of tools to the LeNet convolutional network \cite{LecunBBH98}, which is composed of two convolutional layers (32 filters of size $5\times 5$ at the first layer, and 64 filters of similar size at the second layer, both followed by $2\times  2$ max pooling units), and two fully connected layers ($3136\times 512 \times 10$). While the linearity of the convolution operator immediately allows the application of Net-Trim, in our experiments we omit retraining the convolutional layers as the number of parameters in such layers is much less than the fully connected layers.

For both network architectures we vary $\lambda$ (the $\ell_1$ penalty), and $p$ (the Dropout probability of keeping) in a range of values that tend to produce reasonably high test accuracies. The statistics reported in Table \ref{tab1} correspond to the FC and LeNet  models, which resulted in the highest test accuracies. 
\begin{table}[htbp]
    \centering
\caption{Application of Net-Trim for different values of $\epsilon$ to the standard (FC) and convolutional (LeNet) networks trained via careful choice of the $\ell_1$ regularization and Dropout probability ($\lambda=10^{-5}$, $p=0.75$ for the FC model, and $\lambda=10^{-5}$, $p=0.5$ for the LeNet architecture); improved quantities compared to the initial models are highlighted in bold}\vspace{.2cm}\scriptsize 
\begin{tabular}{V{2.5}c|c|V{2.5}c|c|c|V{2.5}c|c|cV{2.5}}
   \clineB{3-8}{2} 
   \multicolumn{2}{cV{2.5}}{} &\multicolumn{3}{c|V{2.5}}{FC}&\multicolumn{3}{cV{2.5}}{LeNet}\\
\clineB{3-8}{1.7}
 \multicolumn{2}{cV{2.5}}{}  & Network   & Test Acc.     & Test Acc.     & Network   & Test Acc.     & Test Acc.    \\
  \multicolumn{2}{cV{2.5}}{}  & Zeros ($\%$)   & (No FT)     & With FT     & Zeros ($\%$)   & (No FT)     & With FT   \\
    \specialrule{1.2 pt}{0pt}{0pt}
    \multicolumn{2}{V{2.5}c|V{2.5}}{Initial Model}    & 43.69	& 98.65	& -- &	 33.65  &	99.57  & --  \\ \specialrule{1.2 pt}{0pt}{0pt}
    \multirow{ 7}{*}{\begin{turn}{90} Net-Trim \end{turn}} & $\epsilon = 0.01$  & \textbf{71.93}  & \textbf{98.65} &   \textbf{98.76} & \textbf{83.80} &   \textbf{99.59} &   \textbf{99.60}    
    \\ \cline{2-8}
     & $\epsilon = 0.02$   &  \textbf{76.13} & \textbf{98.65}      &     \textbf{98.72}  & \textbf{88.76}   &  \textbf{99.60} &    \textbf{99.57}      \\ \cline{2-8}
     & $\epsilon = 0.04$   &  \textbf{80.02} & 98.56        &     \textbf{98.66}  & \textbf{92.75}   &  99.54 &    99.53      \\ \cline{2-8}
     & $\epsilon = 0.06$   &  \textbf{81.98} & 98.54    &      98.59 &  \textbf{94.46}   &     99.49     &    99.47
    \\ \cline{2-8}
     & $\epsilon = 0.08$   &  \textbf{83.34} & 98.36    &       98.48 & \textbf{95.40}    &    99.35   &      99.44     \\ \cline{2-8}
     & $\epsilon = 0.1$   &   \textbf{84.30} & 98.08    &       98.38 & \textbf{96.01}    &    98.26   &      99.35     \\
     \cline{2-8}
     & $\epsilon = 0.2$   &  \textbf{86.99} & 96.76    &       97.88 & \textbf{97.37}    &    98.83   &      99.22     \\
     \cline{2-8}
     & $\epsilon = 0.3$   &  \textbf{88.61} & 94.69    &       97.31 & \textbf{97.89}    &    98.61   &      99.07     \\
      \specialrule{1.2 pt}{0pt}{0pt}
     \end{tabular}%
  \label{tab1}%
\end{table}
\normalsize 
For both architectures the best results happened when the Dropout and $\ell_1$ regularization were applied simultaneously.  The first row reports the initial model statistics and the subsequent rows correspond to the application of the Net-Trim using different values of $\epsilon$. In this experiment the third column of each architecture section corresponds to an additional fine-tuning step after Net-Trim prunes the network. This (optional) step uses the Net-Trim solution as an initialization for a secondary training, which only applies to the non-zero weights identified by the Net-Trim. Such fine-tuning often results in an improvement in the generalization error without changing the sparsity of the network. 

A quick assessment of Table \ref{tab1} reveals that applying Net-Trim can significantly improve the sparsity (and even at the same time the accuracy) of the models. For instance, in the FC model we can improve the test accuracy to 98.76\%, and at the same time increase the percentage of network zeros from 43.69\% to 71.93\%. A similar trend holds for the LeNet model. If we allow some degradation in the test accuracy, the percentage of zeros can be significantly increased to 88.61\% in the FC model, and 97.89\% in the LeNet architecture.

\begin{figure}[htb!]\vspace{.5cm}\hspace{-.55cm}
\begin{tabular}{c}
\begin{overpic}[ height=.314\textwidth,tics=10]{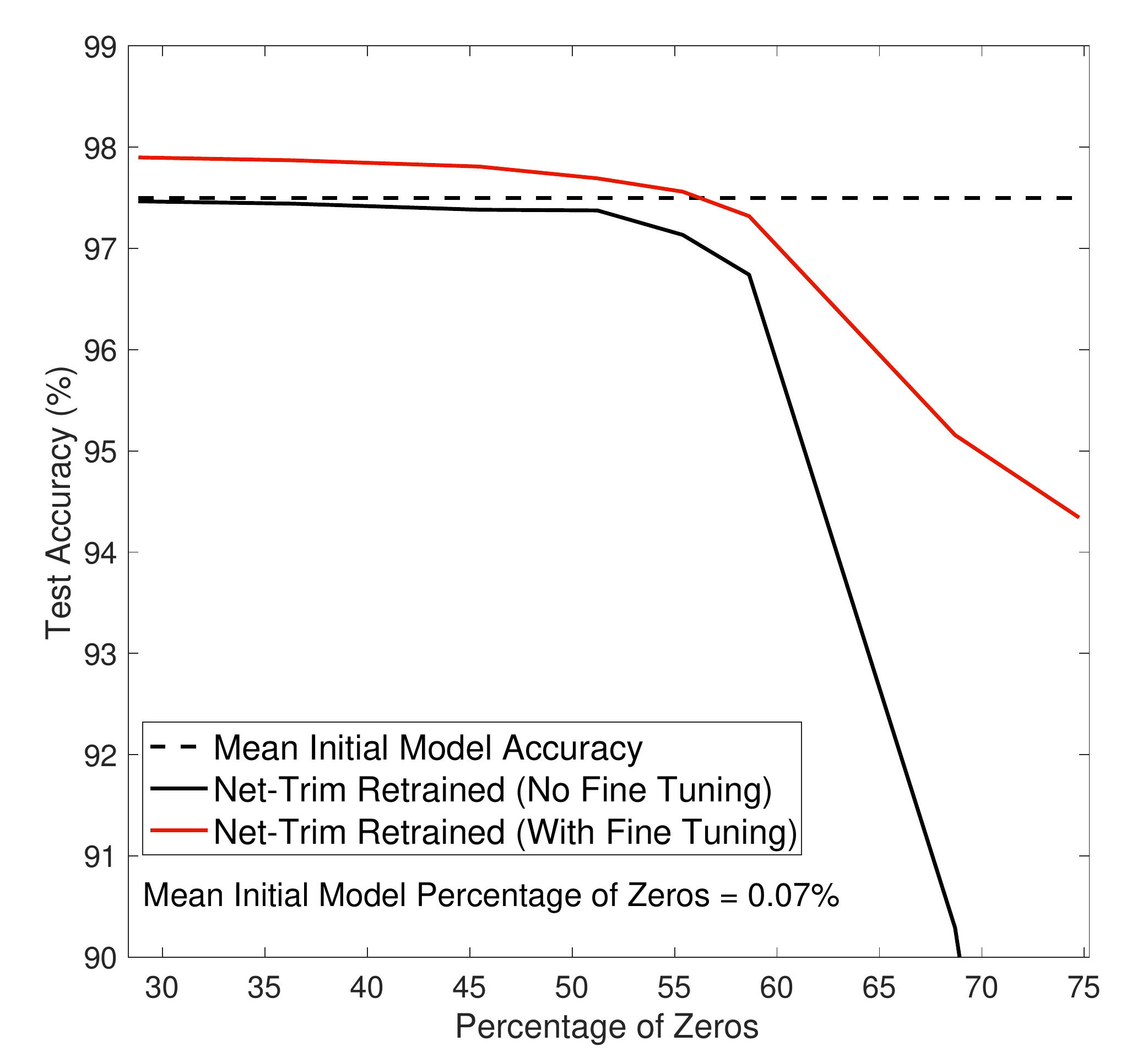}
\put (49,-5.5) {\scalebox{.85}{(a)}}
\end{overpic}\hspace{-.35cm}
\begin{overpic}[ height=.314\textwidth,tics=10]{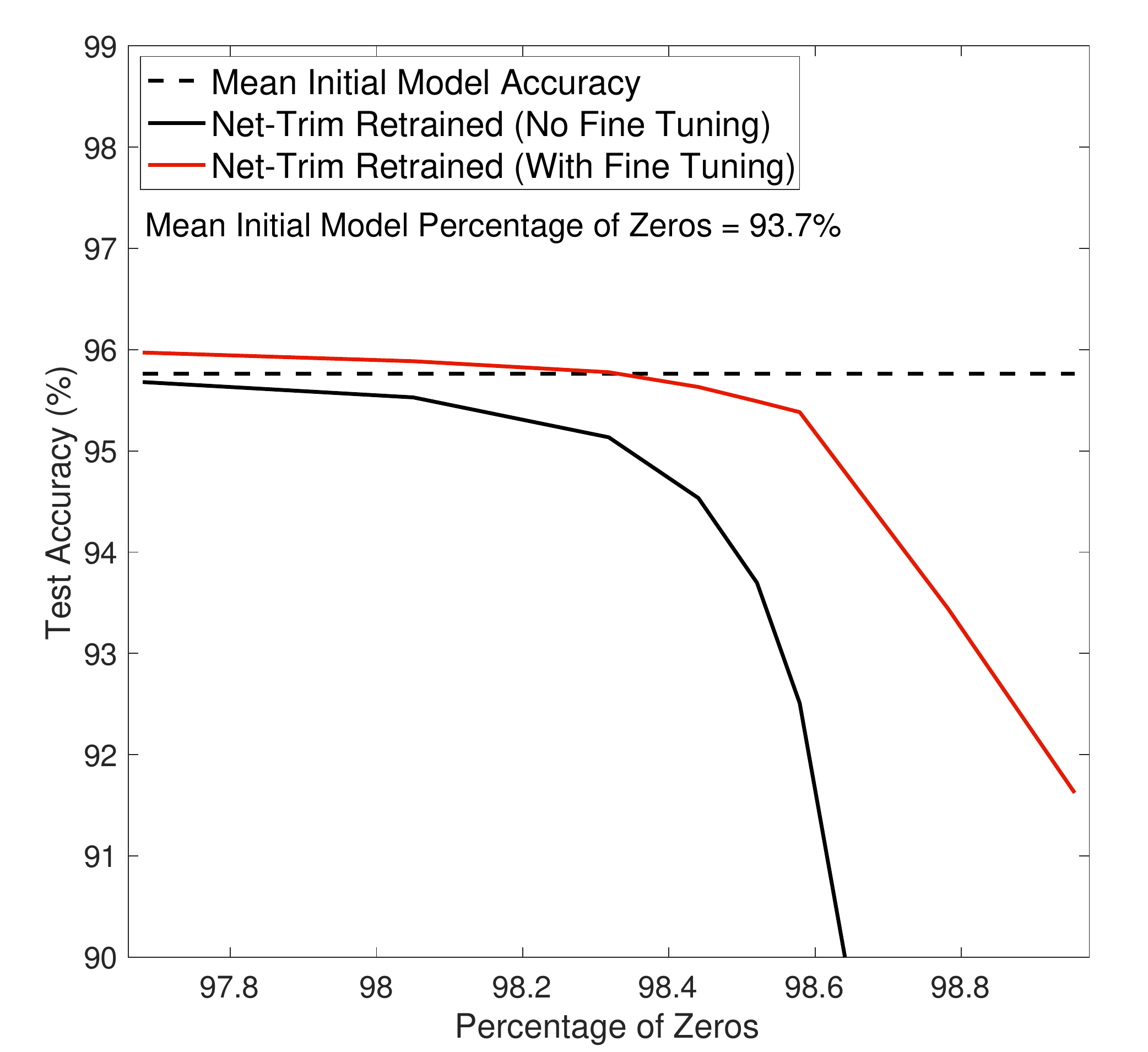}
\put (49,-5.5) {\scalebox{.85}{(b)}}
\end{overpic}\hspace{-.35cm}
\begin{overpic}[ height=.314\textwidth,tics=10]{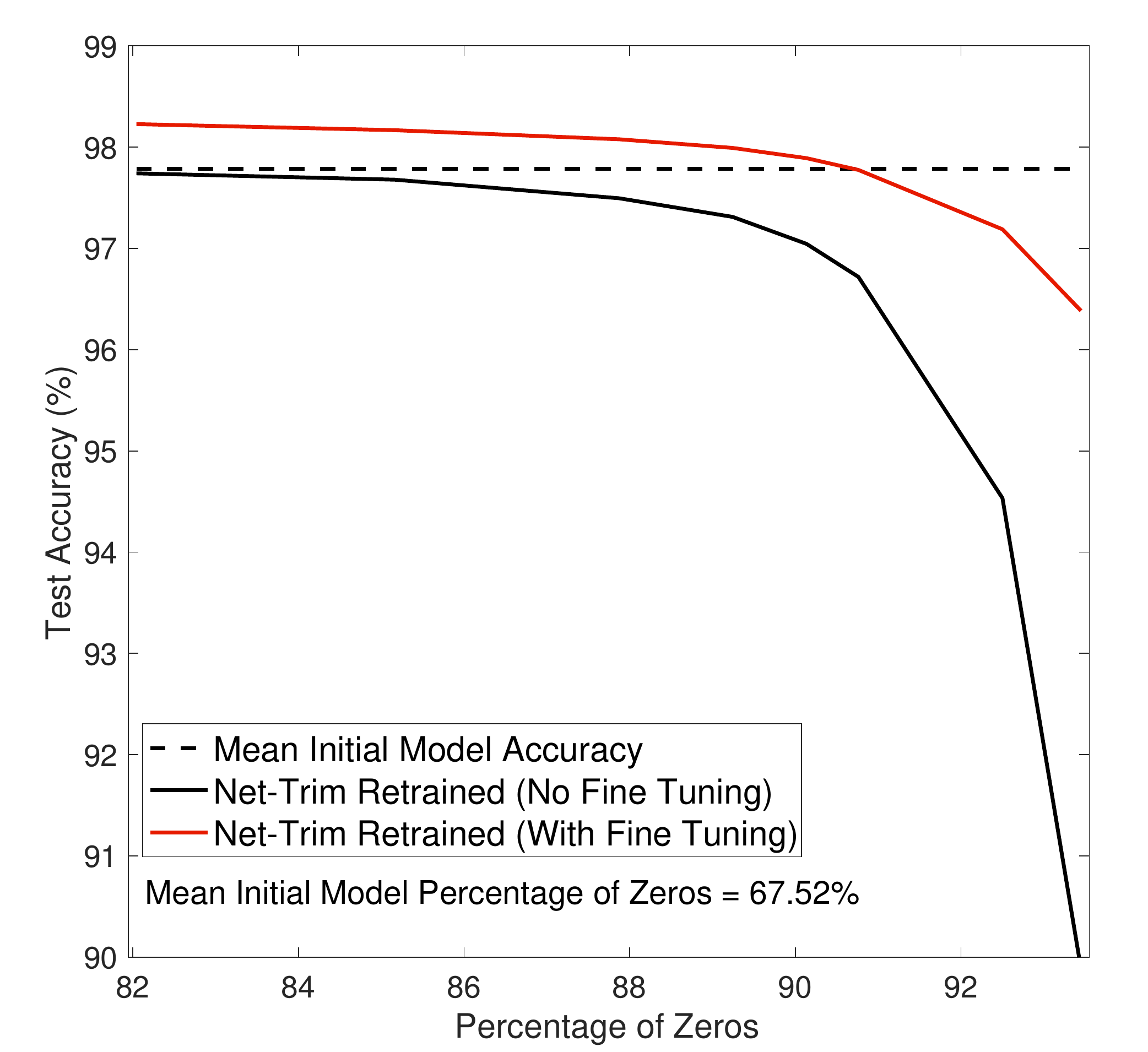}
\put (49,-5.5) {\scalebox{.85}{(c)}}
\end{overpic}
\end{tabular}
 \caption{Mean test accuracy vs mean model sparsity after the application of Net-Trim for FC network initially regularized via $\ell_1$ penalty, Dropout, or both (the regularization parameter and Dropout probability are picked from a range of values and mean quantities are reported); (a) a model trained with Dropout only: $0.3\leq p \leq 0.8$; (b) a model trained with $\ell_1$ penalty only: $10^{-5}\leq \lambda \leq 5\times 10^{-3}$; (c) a model trained with Dropout and $\ell_1$: $10^{-5}\leq \lambda \leq 2\times 10^{-4}$, $0.5\leq p \leq 0.75$;}\label{fig3}
\end{figure}

Table \ref{tab1} only reports the Net-Trim performance on the most accurate models. In Figure \ref{fig3m} of the paper we presented a more comprehensive set of experiments on the LeNet Network. Figure \ref{fig3} shows a similar set of experiments on the FC network.  
In these experiments the mean test accuracy and initial model sparsity are reported for the cases of Dropout, $\ell_1$ regularization, and a combination of both. For each setup the tuning parameters ($\lambda$, $p$, or both) are varied in a range of values and unlike Table \ref{tab1}, the mean quantities are reported. For instance, panel (a) indicates that  applying the Dropout to the FC model with $0.3\leq p\leq 0.8$ yields an average network zero percentage of $0.07\%$, and approximately 97.5\% test accuracy. However, applying Net-Trim along with the FT step, can elevate the average accuracy to around 98\%, and at the same time increase the network percentage of zeros to almost 45\%. The plot also reveals that with no loss in the model accuracies, we can improve the sparsity of the models to up to 56\% (corresponding to the point where the red and the dashed lines intersect).

\begin{figure}[htb!]\vspace{.5cm}\hspace{-.55cm}
\centering\begin{tabular}{c}
\begin{overpic}[ height=.3\textwidth,tics=10]{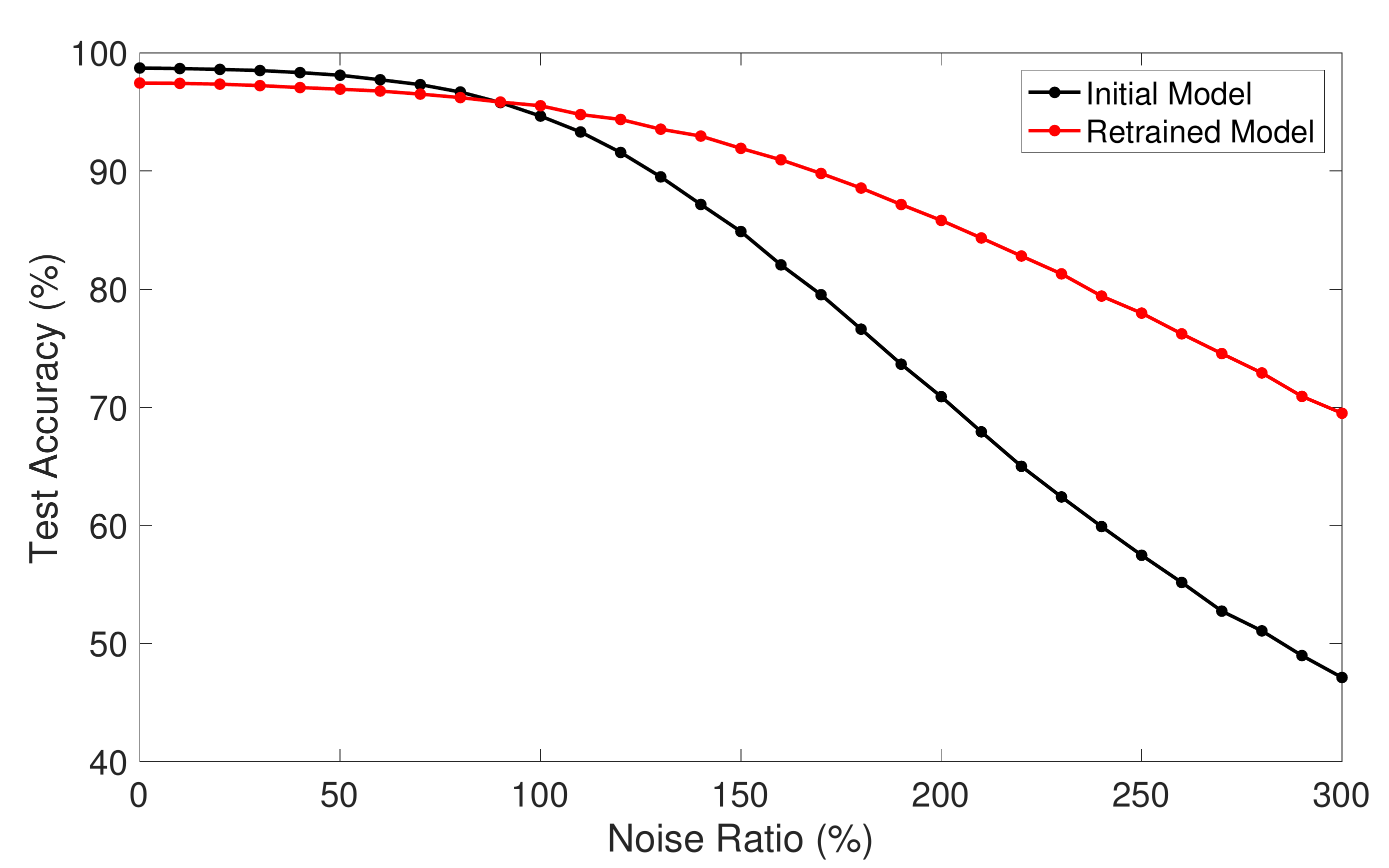}
\put (49,-3.5) {\scalebox{.85}{(a)}}
\end{overpic}\hspace{-.1cm}
\begin{overpic}[ height=.3\textwidth,tics=10]{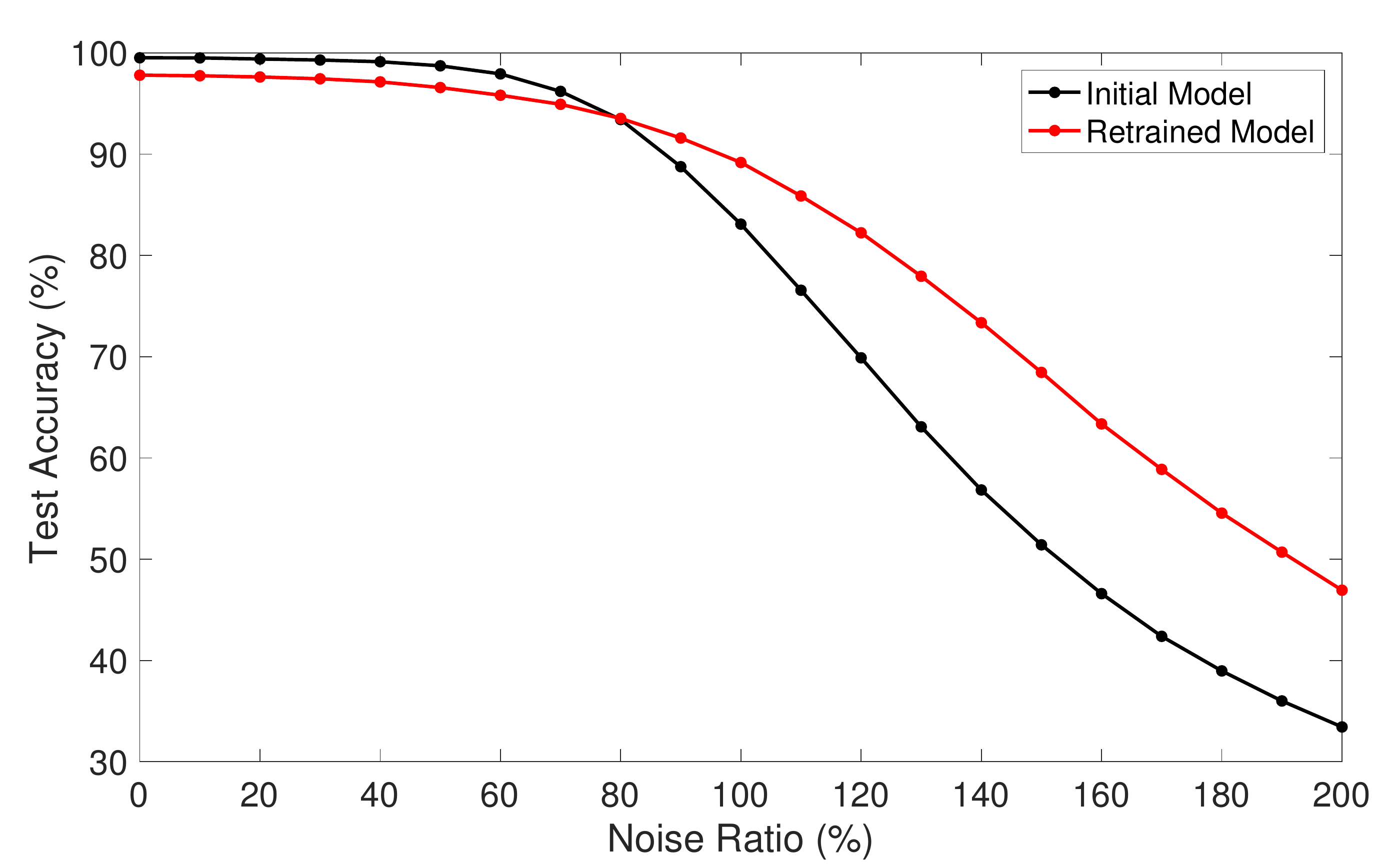}
\put (49,-3.5) {\scalebox{.85}{(b)}}
\end{overpic}\hspace{-.35cm}
\end{tabular}
 \caption{Noise robustness of initial and retrained networks; (a) FC; (b) LeNet}\label{fig3.5}
\end{figure}

An assessment of all panels (specifically the crossing of the red curves and the dashed lines) reveals that in all three scenarios (Dropout, $\ell_1$ regularization and a combination of both), and for both architectures (FC and LeNet), an additional application of Net-Trim can improve the models both in terms of accuracy and the number of underlying parameters. Even in cases that the accuracy is degraded to some extent, but the model is significantly pruned, the pruned network may be considered a more reliable model. In Figure \eqref{fig3.5} we have demonstrated the FC and LeNet models initially trained with Dropout and retrained using Net-Trim. Despite an accuracy loss of 1.3\% for the FC model, and 1.7\% for the LeNet model, the percentage of zeros have been increased to 63.32\% and 96.8\%, respectively. As a result of this reduction, when the models are tested with different noisy versions of the original test set, the reduced models exhibit a lower accuracy degradation (i.e., more robustness) to the noise increase. 

Thanks to the simple implementation of Net-Trim, in the aforementioned experiments, the retraining of the layer matrices was only in order of few minutes on a standard desktop computer, while in the majority of the cases, the initial training of the networks took much longer. We would like to note that we did not make any efforts to optimize the Net-Trim code and fully exploit the parallel features (e.g., matrix products, processing of layers in parallel, etc). The distributable nature of our implementation supports yet much faster software than the one currently present.

In the paper we also compared Net-Trim with the HPTD \cite{han2015learning}.  The HPTD algorithm does not come with any performance guarantees, however, the basic implementation idea has made it a widespread tool in the network compression community. Using tools such as quantization and Huffman coding, more advanced frameworks such as the Deep Compression \cite{han2015deep_compression} have been developed later. However, their focus is mainly compressing the network parameters on the memory, and HPTD pruning scheme is yet the most relevant single-module framework that could be compared with Net-Trim.

\begin{figure}[htb!]\vspace{.5cm}\hspace{-.13cm}
\centering \begin{tabular}{c}
\begin{overpic}[ trim={1.5cm .8cm 0 0},clip, height=.302\textwidth,tics=10]{All_FC.pdf}
\put (-2,35){\rotatebox{90}{\scalebox{.5}{Test Accuracy (\%)}}}
\put (35,-3){\rotatebox{0}{\scalebox{.5}{Percentage of Zeros}}}
\end{overpic}\hspace{-.08cm}
\begin{overpic}[ trim={.85cm 1cm 0 0},clip,height=.30\textwidth,tics=10]{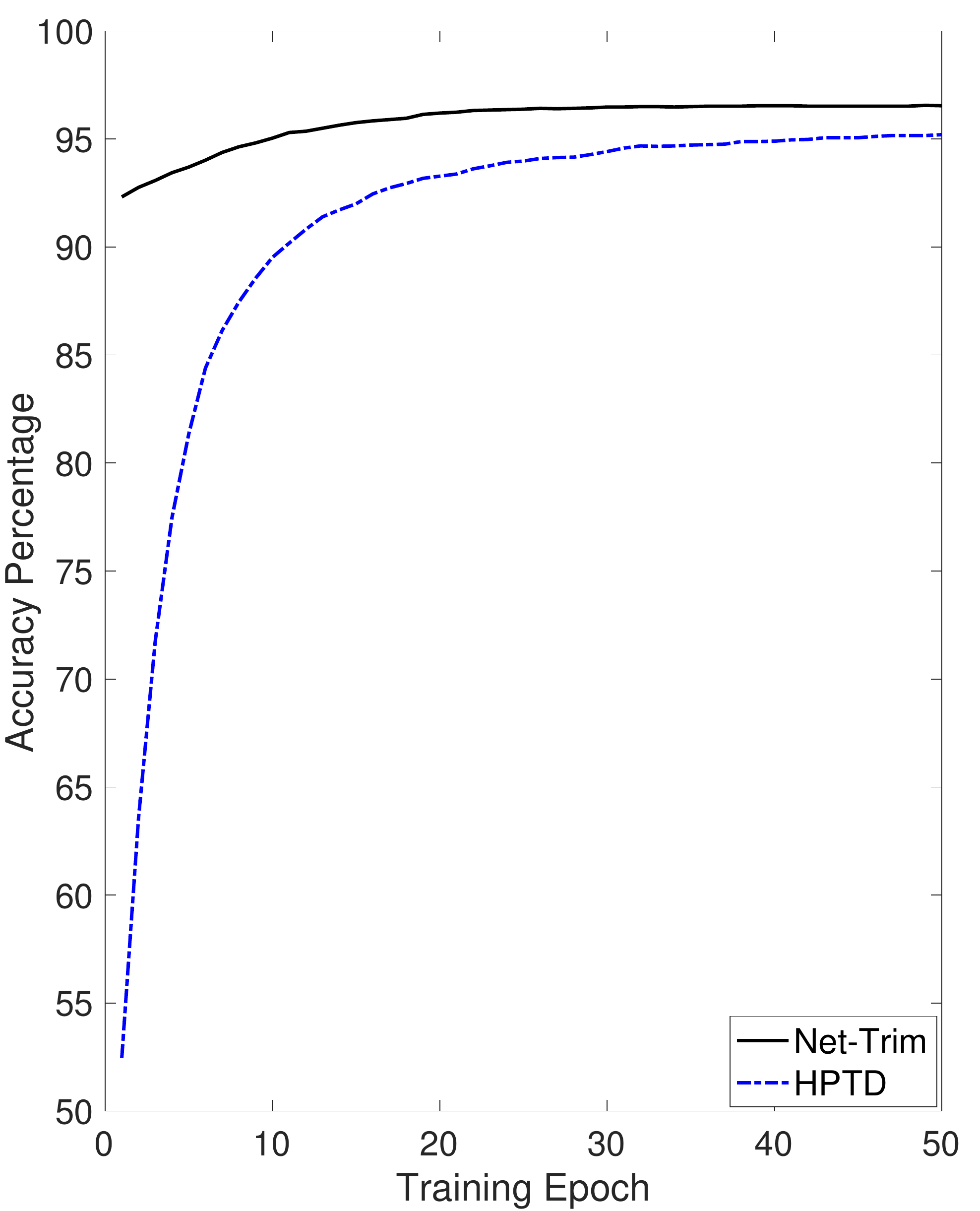}
\put (40,-9) {\scalebox{.85}{(a)}}
\put (-3,35){\rotatebox{90}{\scalebox{.5}{Test Accuracy (\%)}}}
\put (30,-3){\rotatebox{0}{\scalebox{.5}{Fine-Tuning Epoch}}}
\end{overpic}\hspace{.12cm}
\begin{overpic}[ trim={.85cm .8cm 0 0},clip,height=.302\textwidth,tics=10]{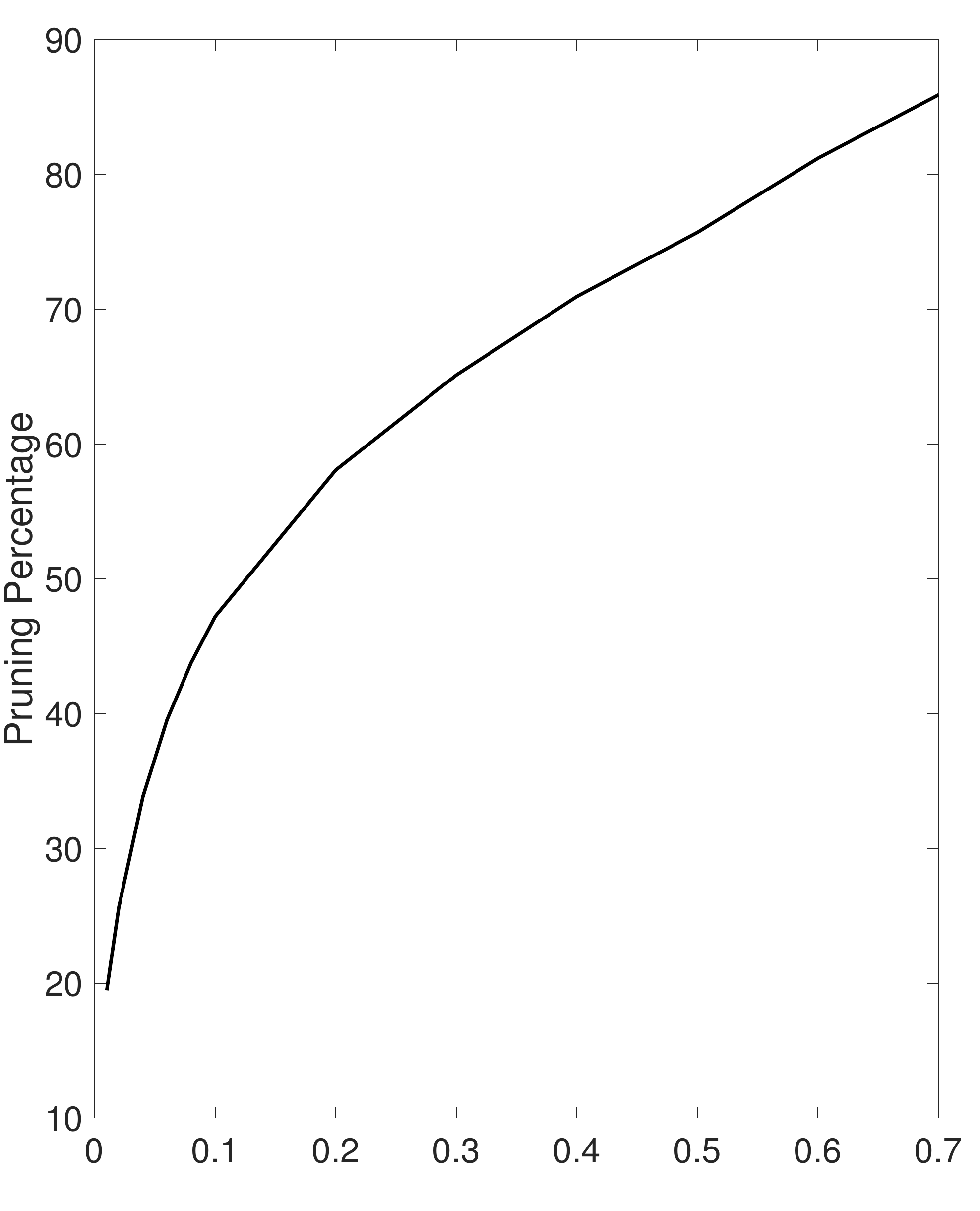}
\put (40,-3.5) {\scalebox{.85}{$\epsilon$}}
\put (-3,35){\rotatebox{90}{\scalebox{.5}{Percentage of Zeros}}}
\end{overpic}\hspace{.03cm}
\\[.4cm]
\begin{overpic}[trim={1.4cm .8cm 0 0},clip,height=.303\textwidth,tics=10]{All_Lenet.pdf}
\put (-2,35){\rotatebox{90}{\scalebox{.5}{Test Accuracy (\%)}}}
\put (35,-3){\rotatebox{0}{\scalebox{.5}{Percentage of Zeros}}}
\end{overpic}\hspace{-.08cm}
\begin{overpic}[ trim={.85cm 1cm 0 0},clip, height=.30\textwidth,tics=10]{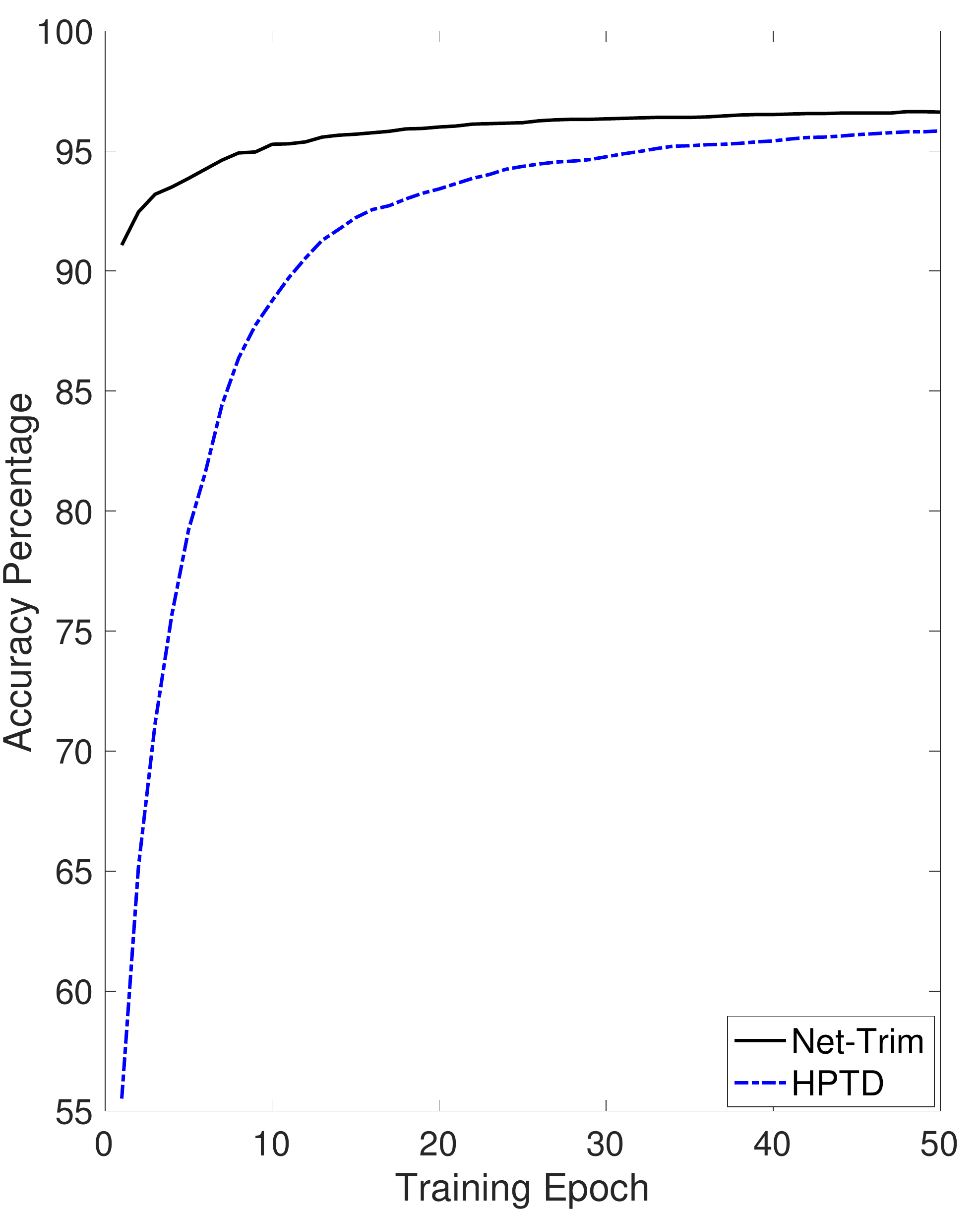}
\put (40,-9) {\scalebox{.85}{(b)}}
\put (-3,35){\rotatebox{90}{\scalebox{.5}{Test Accuracy (\%)}}}
\put (30,-3){\rotatebox{0}{\scalebox{.5}{Fine-Tuning Epoch}}}
\end{overpic}\hspace{.12cm}
\begin{overpic}[ trim={.85cm .8cm 0 0},clip,height=.302\textwidth,tics=10]{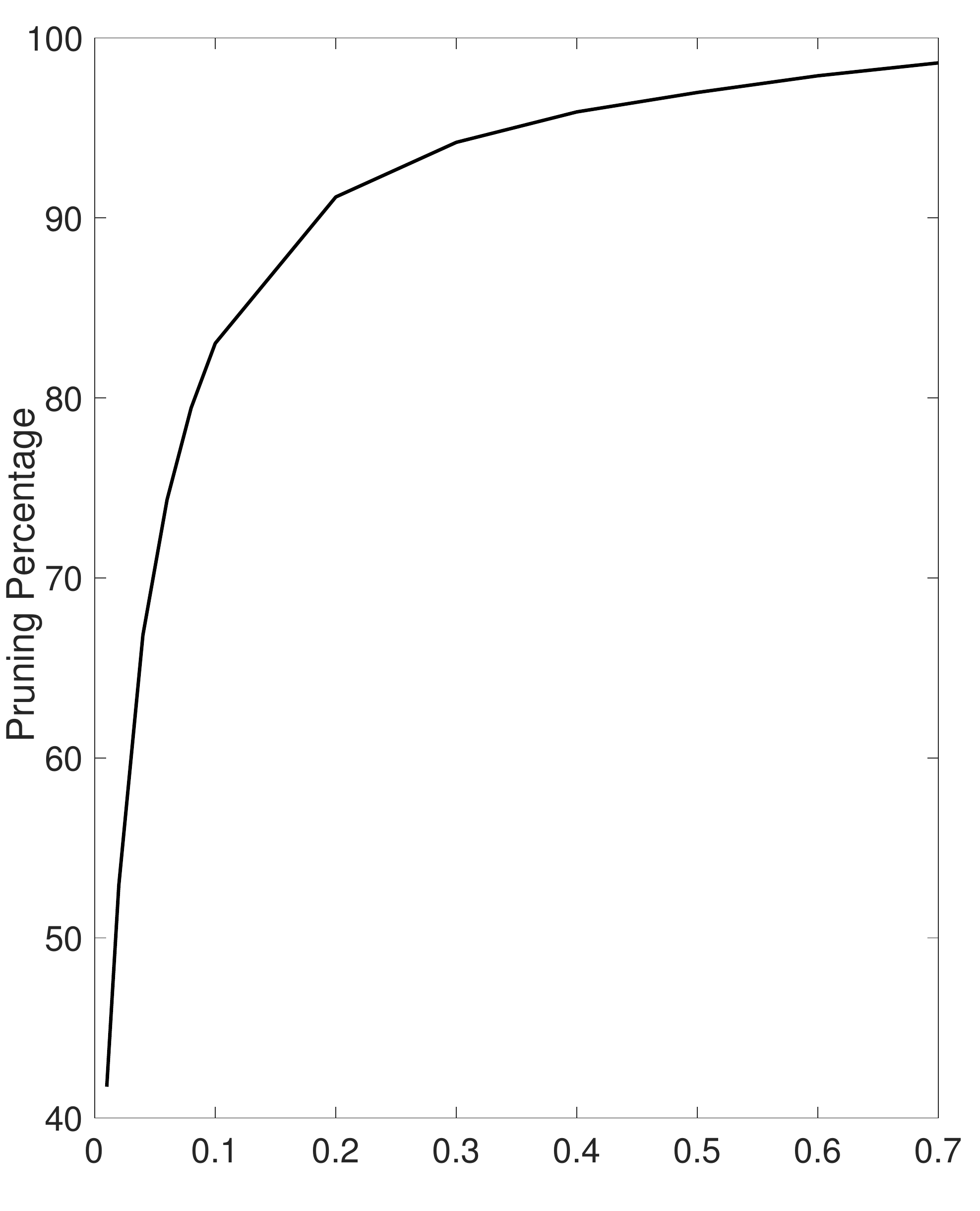}
\put (40,-3.5) {\scalebox{.85}{$\epsilon$}}
\put (-1.57,96) {\scalebox{.43}{1}}
\put (-3,35){\rotatebox{90}{\scalebox{.5}{Percentage of Zeros}}}
\end{overpic}\vspace{.2cm}
\end{tabular}
 \caption{Comparison of Net-Trim and HPTD in different settings for (a) FC model, (b) LeNet model; the left panels compare Net-Trim and HPTD test accuracy vs percentage of zeros, without fine-tuning, and with fine-tuning using 10 and 30 epochs; middle panels show the number of fine-tuning epochs and the acquired accuracy using Net-Trim and HPTD; the right panels indicate the percentage of zeros as a function $\epsilon$ for Net-Trim}\label{fig4}
\end{figure}
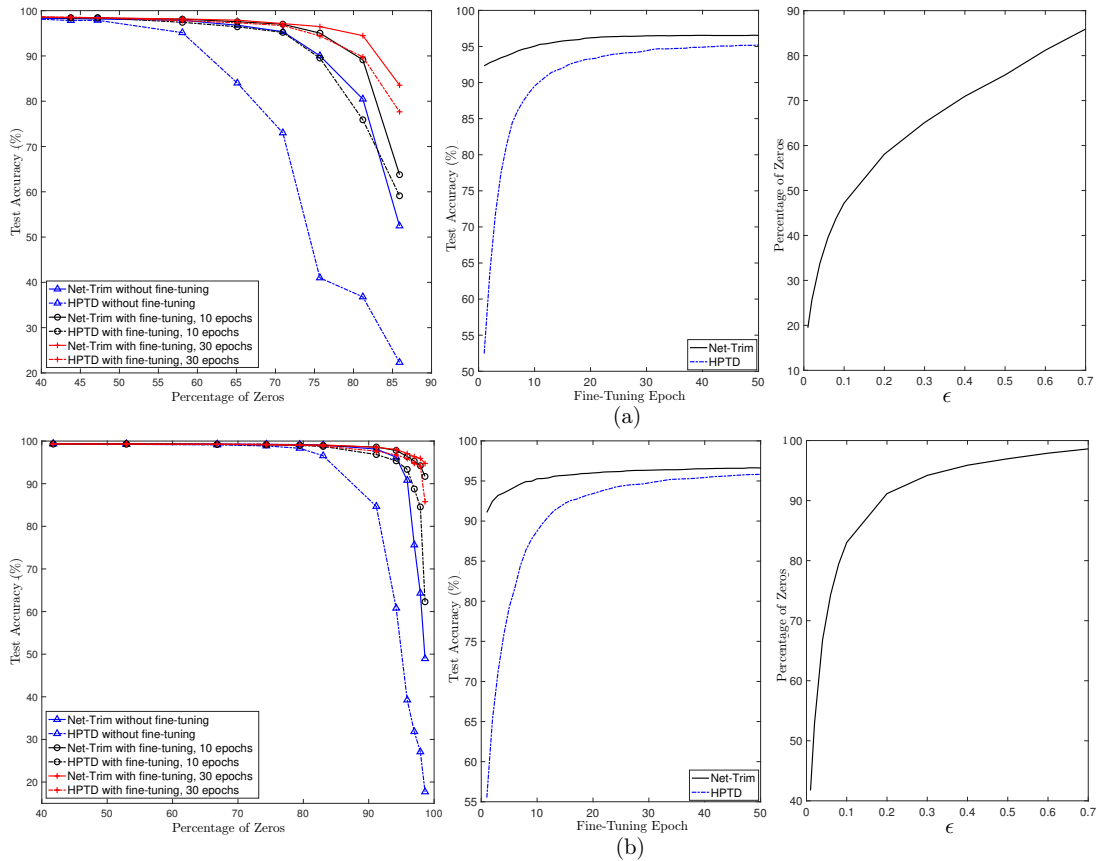

With reference to Figure \ref{fig4m} of the paper, Figure \ref{fig4} presents more details of the comparison between the Net-Trim and HPTD on the FC and LeNet models. 
For the Net-Trim we use different values of $\epsilon$ to prune the trained networks. To compare the method with the HPTD, after each application of the Net-Trim and counting the number of zeros, the same number of elements are truncated from the initial network to be used for the HPTD implementation. HPTD is followed by a fine-tuning step after the truncation, which is also an optional task for Net-Trim. Nevertheless, both algorithms are compared without fine-tuning, or with fine-tuning using 10 or 30 epochs. 
The left plots in panels (a) and (b) show that in all scenarios Net-Trim outperforms HPTD in generating more accurate models when the levels of sparsity are matched. The middle plots also show the improvements in the accuracy as a function of the number of epochs required in the fine-tuning process for the two schemes. In both scenarios, Net-Trim requires only few epochs to achieve the top accuracy, while achieving such level of accuracy for the HPTD is either not feasible or takes many fine-tuning epochs.

\begin{figure}[htb!]\vspace{.5cm}\hspace{-.13cm}
\centering \begin{tabular}{c}
\begin{overpic}[ trim={1.8cm .8cm 0 0},clip,height=.302\textwidth,tics=10]{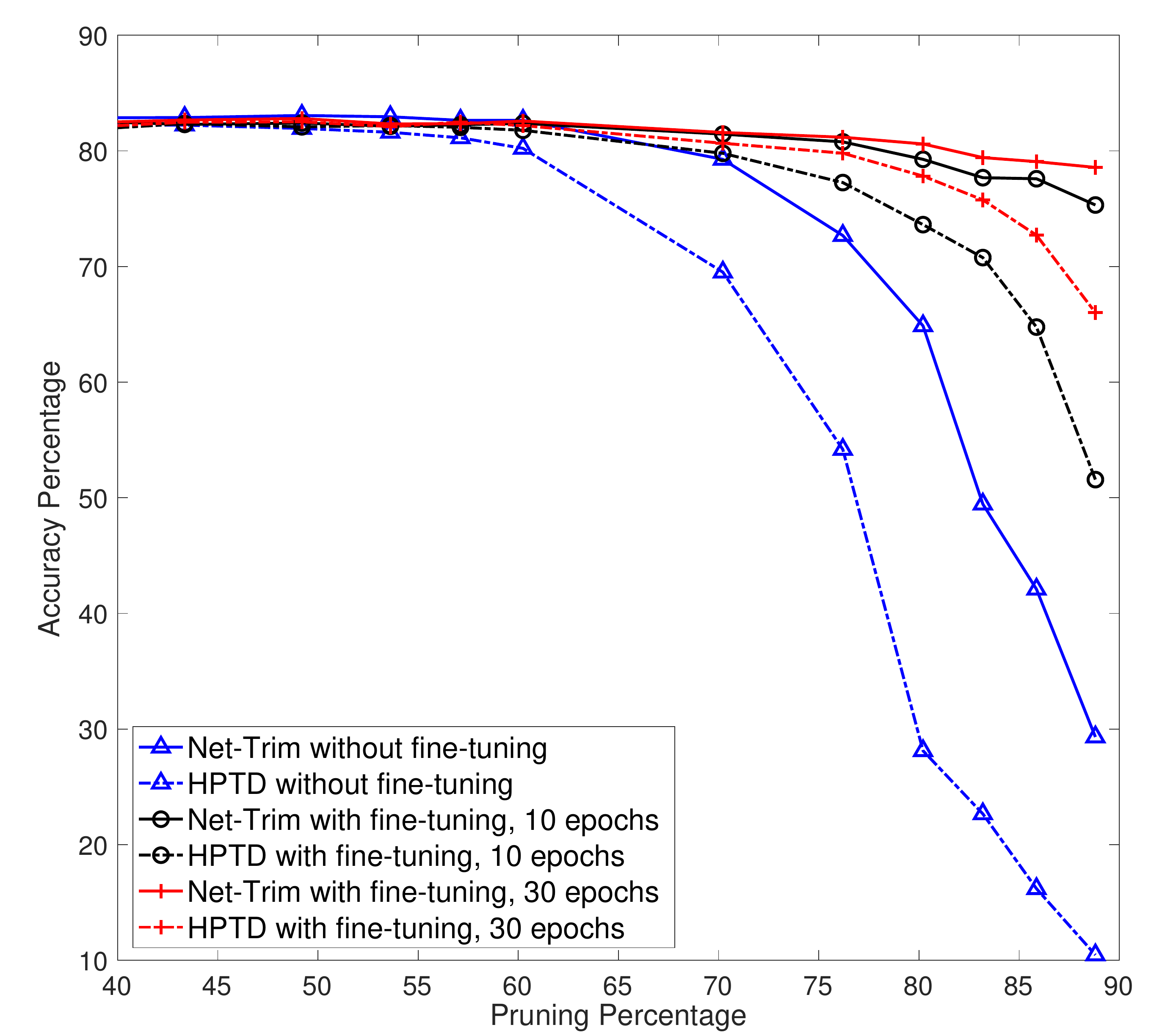}
\put (-4,35){\rotatebox{90}{\scalebox{.5}{Test Accuracy (\%)}}}
\put (35,-3){\rotatebox{0}{\scalebox{.5}{Percentage of Zeros}}}
\end{overpic}\hspace{-.06cm}
\begin{overpic}[ trim={.95cm 1cm 0 0},clip,height=.30\textwidth,tics=10]{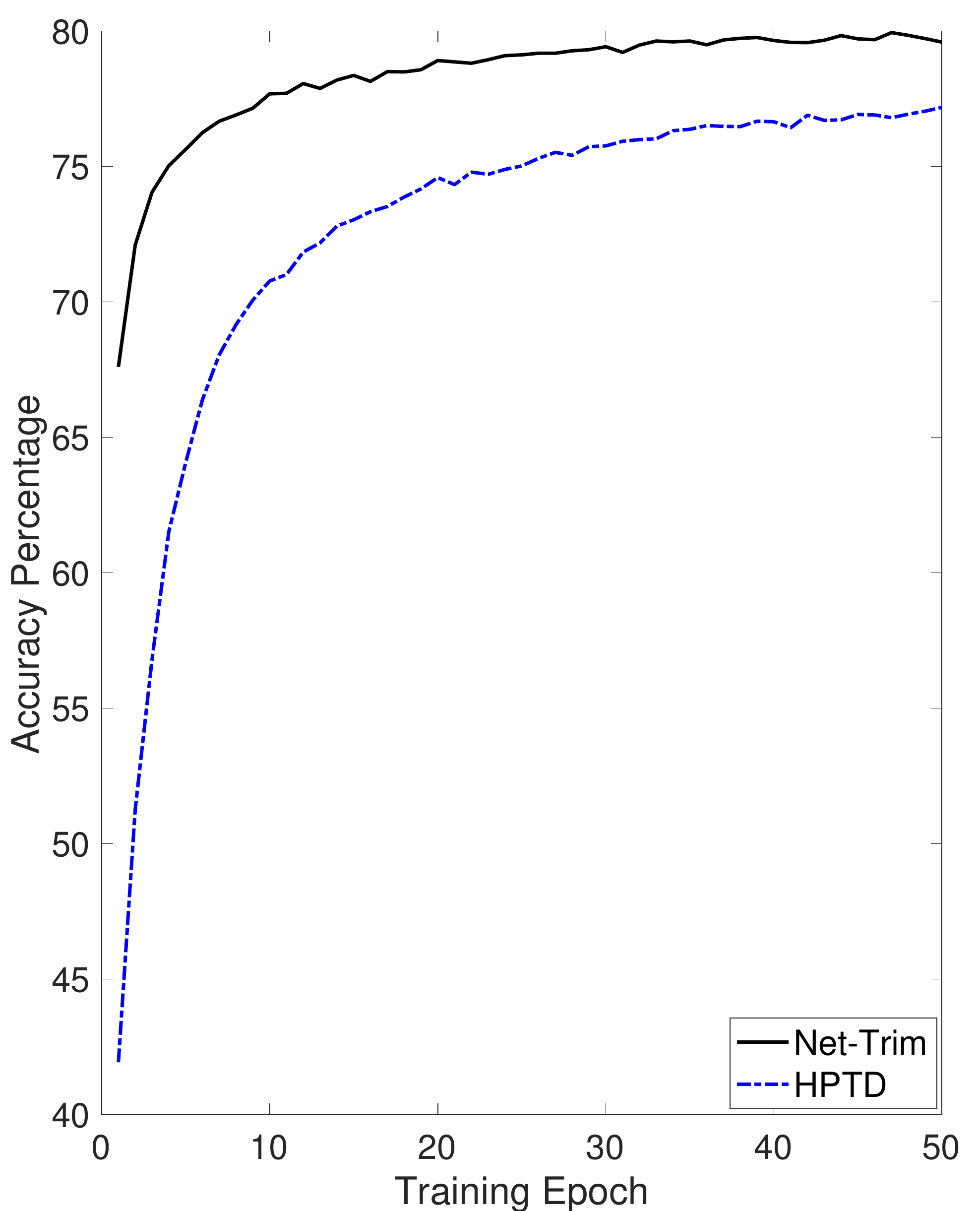}
\put (40,-9.5) {\scalebox{.85}{(a)}}
\put (-4,35){\rotatebox{90}{\scalebox{.5}{Test Accuracy (\%)}}}
\put (30,-3){\rotatebox{0}{\scalebox{.5}{Fine-Tuning Epoch}}}
\end{overpic}\hspace{.12cm}
\begin{overpic}[ trim={.85cm .8cm 0 0},clip,height=.302\textwidth,tics=10]{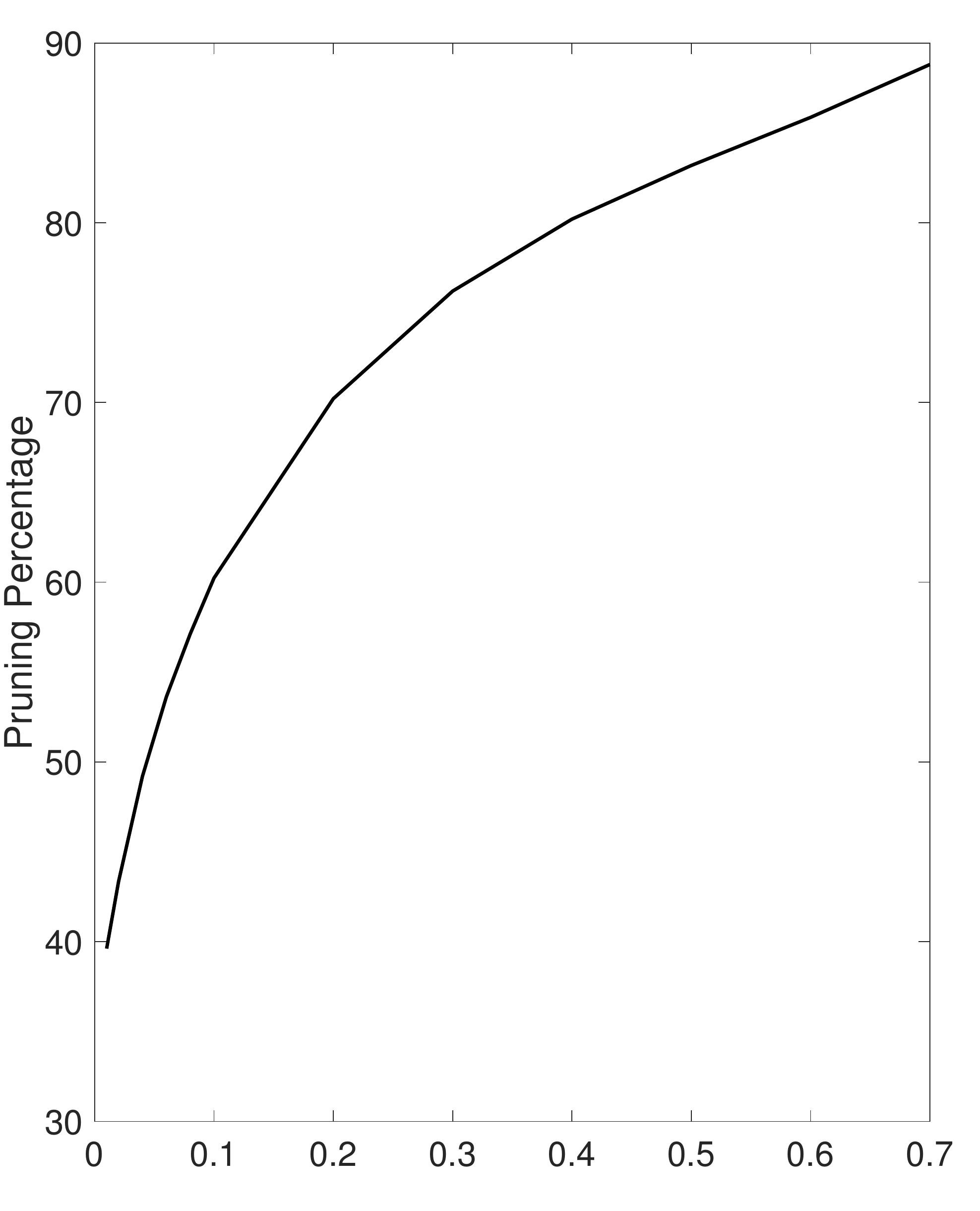}
\put (40,-3.5) {\scalebox{.85}{$\epsilon$}}
\put (-4,35){\rotatebox{90}{\scalebox{.5}{Percentage of Zeros}}}
\end{overpic}\hspace{-.12cm}
\\[.4cm]
\hspace{.1cm}\begin{overpic}[ trim={1.8cm .8cm 0 0},clip,height=.302\textwidth,tics=10]{All_Cifar50.pdf}
\put (-4,35){\rotatebox{90}{\scalebox{.5}{Test Accuracy (\%)}}}
\put (35,-3){\rotatebox{0}{\scalebox{.5}{Percentage of Zeros}}}
\end{overpic}\hspace{-.06cm}
\begin{overpic}[ trim={.9cm 1cm 0 0},clip,height=.30\textwidth,tics=10]{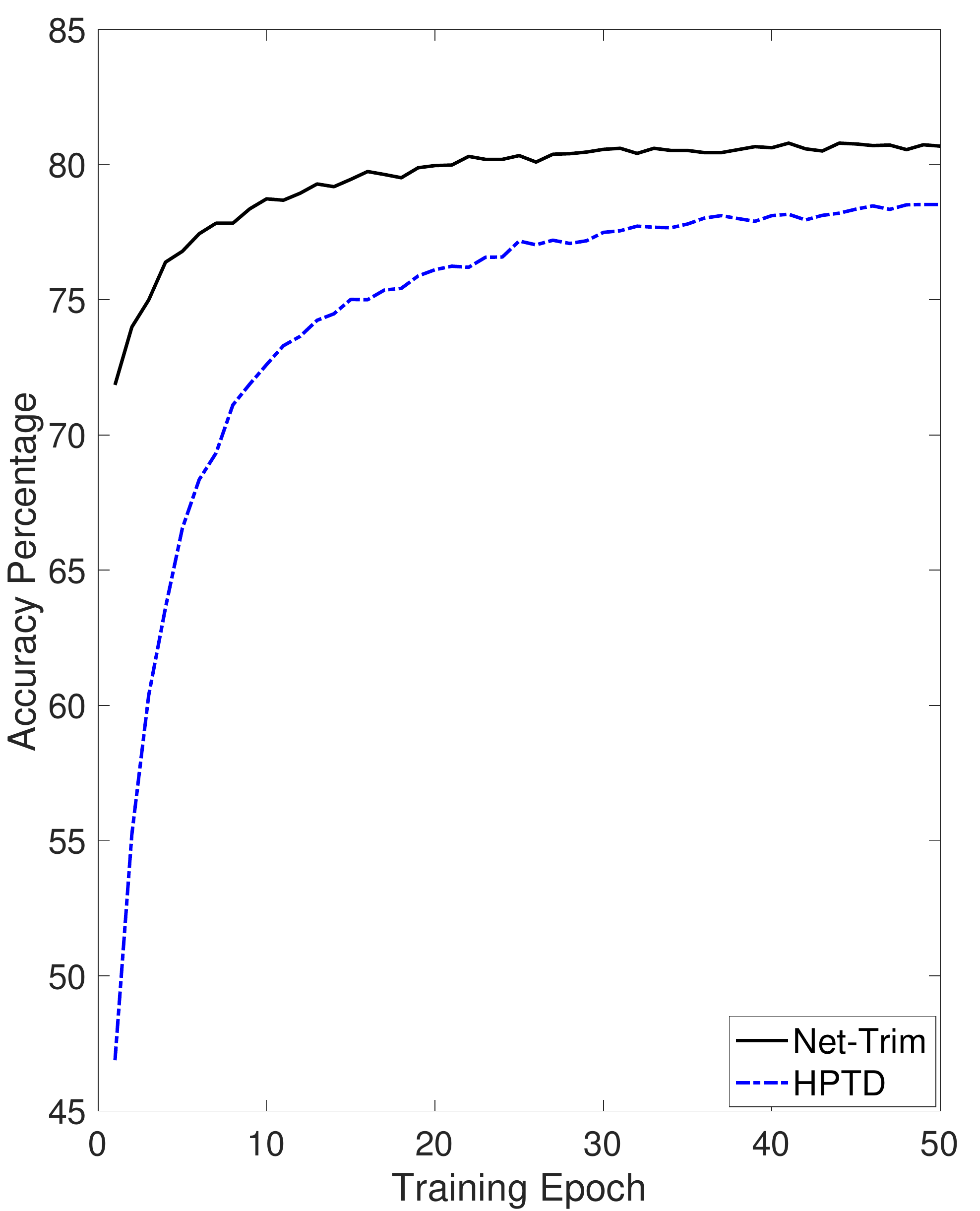}
\put (40,-9.5) {\scalebox{.85}{(b)}}
\put (-4,35){\rotatebox{90}{\scalebox{.5}{Test Accuracy (\%)}}}
\put (30,-3){\rotatebox{0}{\scalebox{.5}{Fine-Tuning Epoch}}}
\end{overpic}\hspace{.12cm}
\begin{overpic}[ trim={.85cm .8cm 0 0},clip,height=.302\textwidth,tics=10]{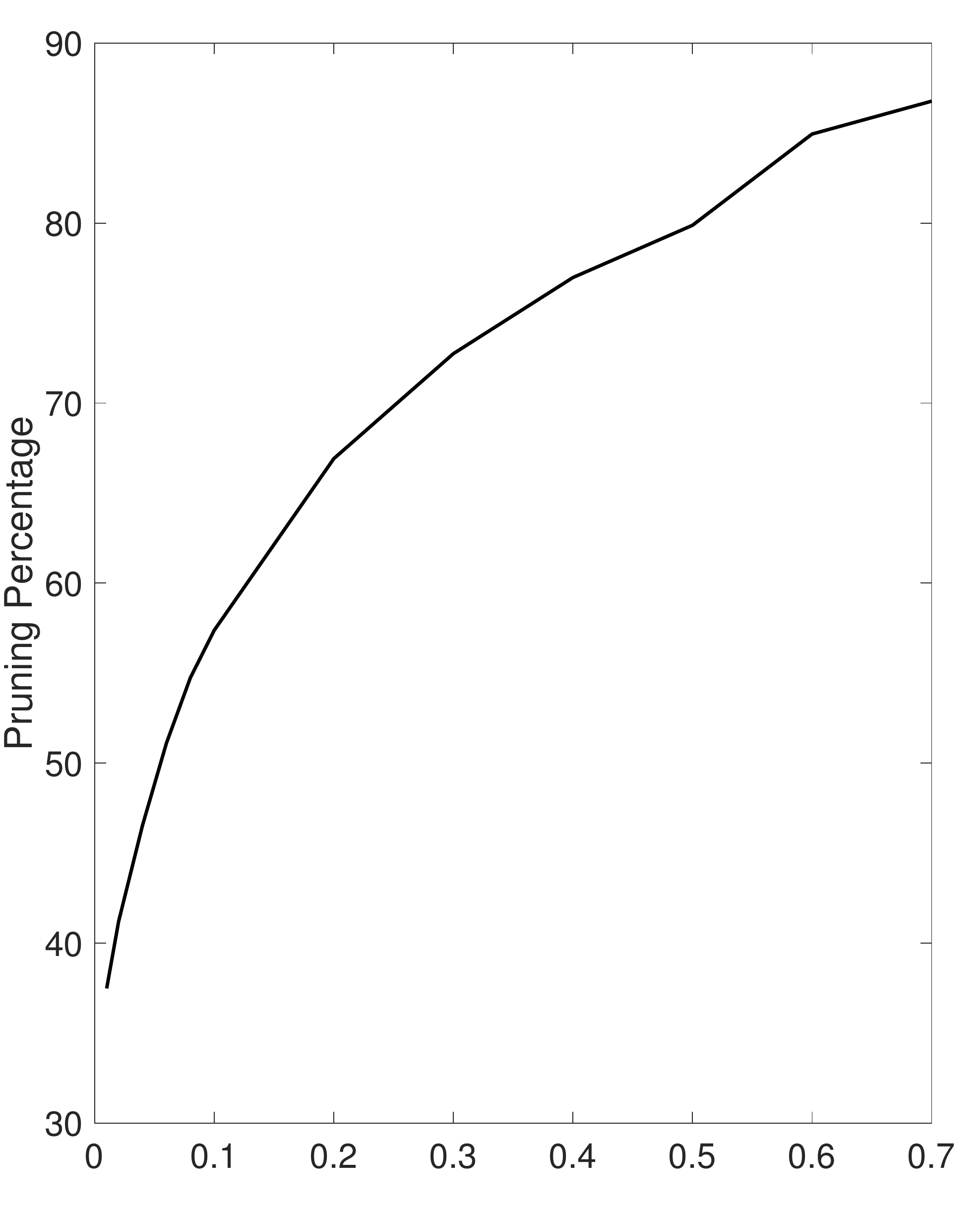}
\put (40,-3.5) {\scalebox{.85}{$\epsilon$}}
\put (-4,35){\rotatebox{90}{\scalebox{.5}{Percentage of Zeros}}}
\end{overpic}\hspace{-.03cm}
\\[.4cm]
\hspace{.1cm}\begin{overpic}[ trim={1.8cm .8cm 0 0},clip,height=.302\textwidth,tics=10]{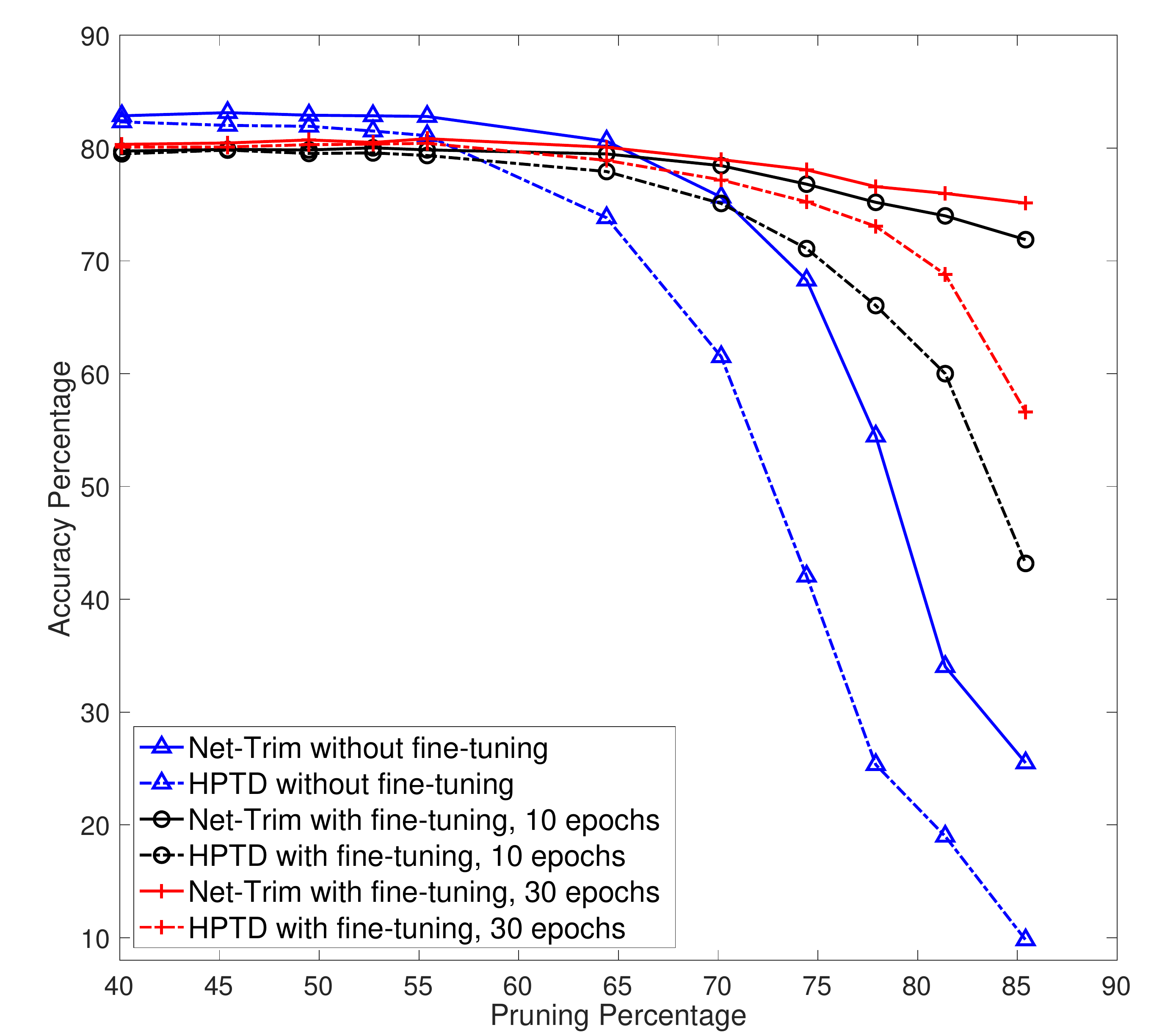}
\put (-4,35){\rotatebox{90}{\scalebox{.5}{Test Accuracy (\%)}}}
\put (35,-3){\rotatebox{0}{\scalebox{.5}{Percentage of Zeros}}}
\end{overpic}\hspace{-.06cm}
\begin{overpic}[ trim={.85cm 1cm 0 0},clip, height=.30\textwidth,tics=10]{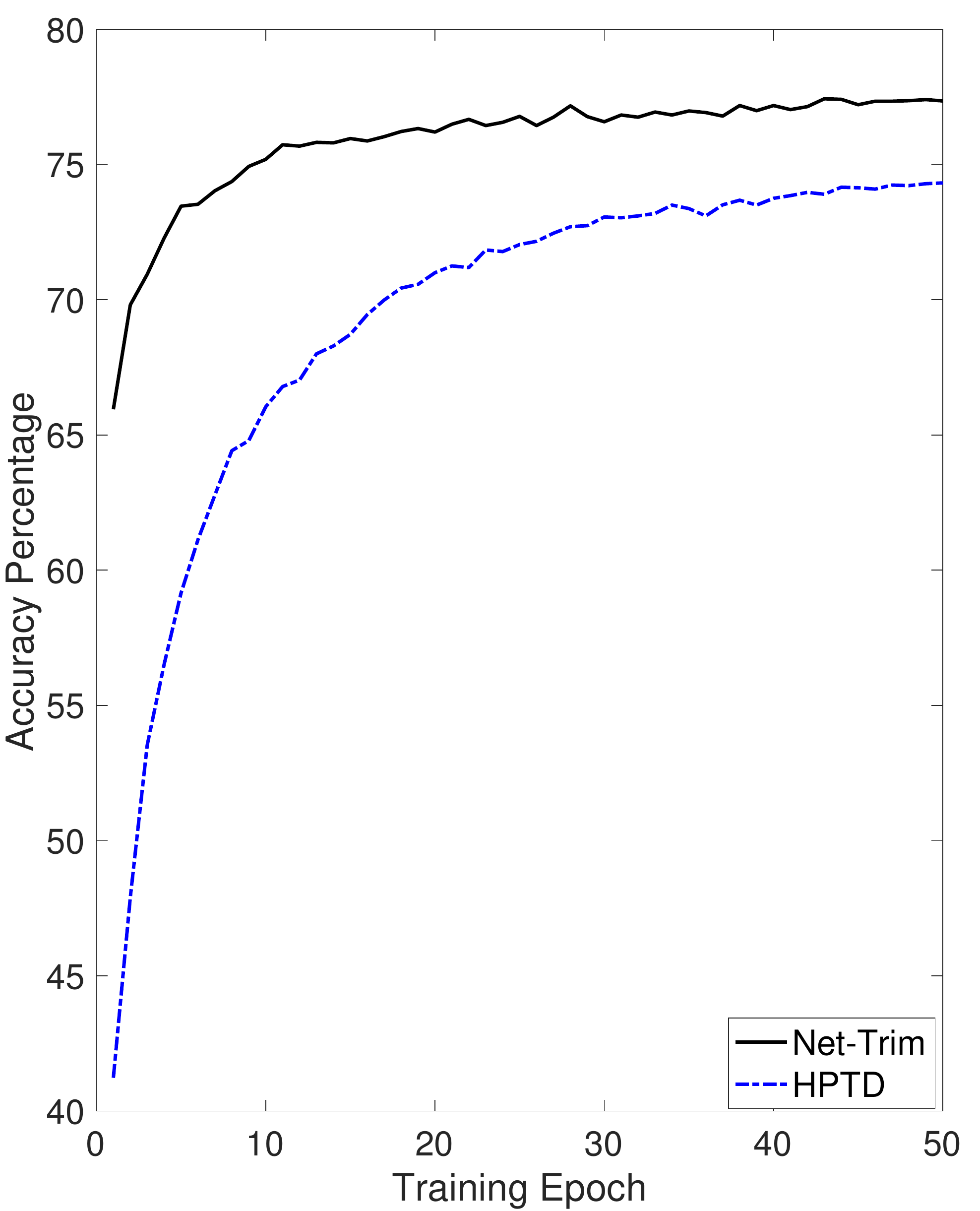}
\put (40,-9.5) {\scalebox{.85}{(c)}}
\put (-4,35){\rotatebox{90}{\scalebox{.5}{Test Accuracy (\%)}}}
\put (30,-3){\rotatebox{0}{\scalebox{.5}{Fine-Tuning Epoch}}}
\end{overpic}\hspace{.12cm}
\begin{overpic}[ trim={.85cm .8cm 0 0},clip,height=.302\textwidth,tics=10]{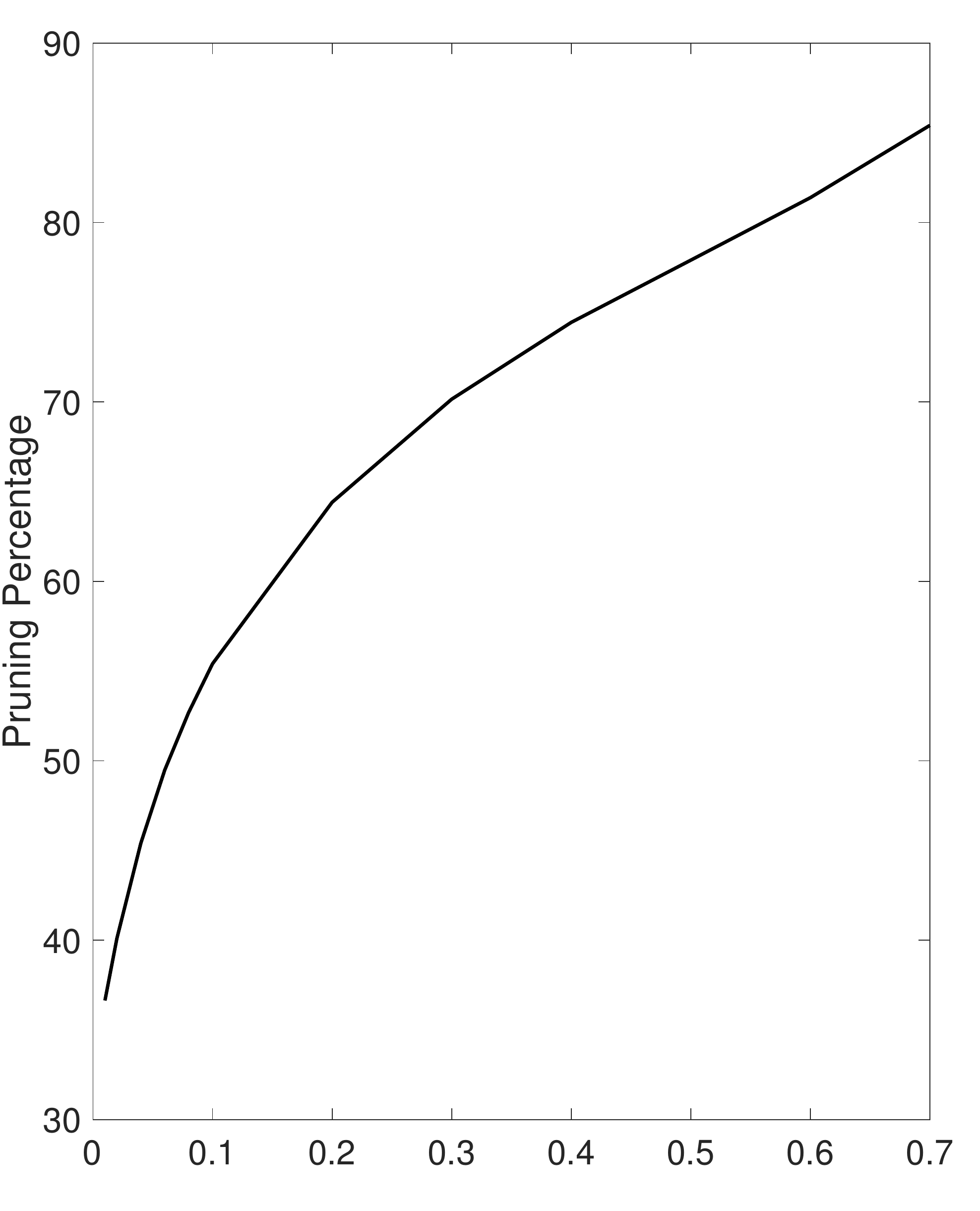}
\put (40,-3.5) {\scalebox{.85}{$\epsilon$}}
\put (-4,35){\rotatebox{90}{\scalebox{.5}{Percentage of Zeros}}}
\end{overpic}\vspace{.13cm}
\end{tabular}
 \caption{Similar comparison plots as in Figure \ref{fig4} for CIFAR-10: (a) retraining of Net-Trim and HPTD performed using 25K samples; (b) retraining performed using 50K samples; (c) retraining performed using 75K samples}\label{fig6}
\end{figure}
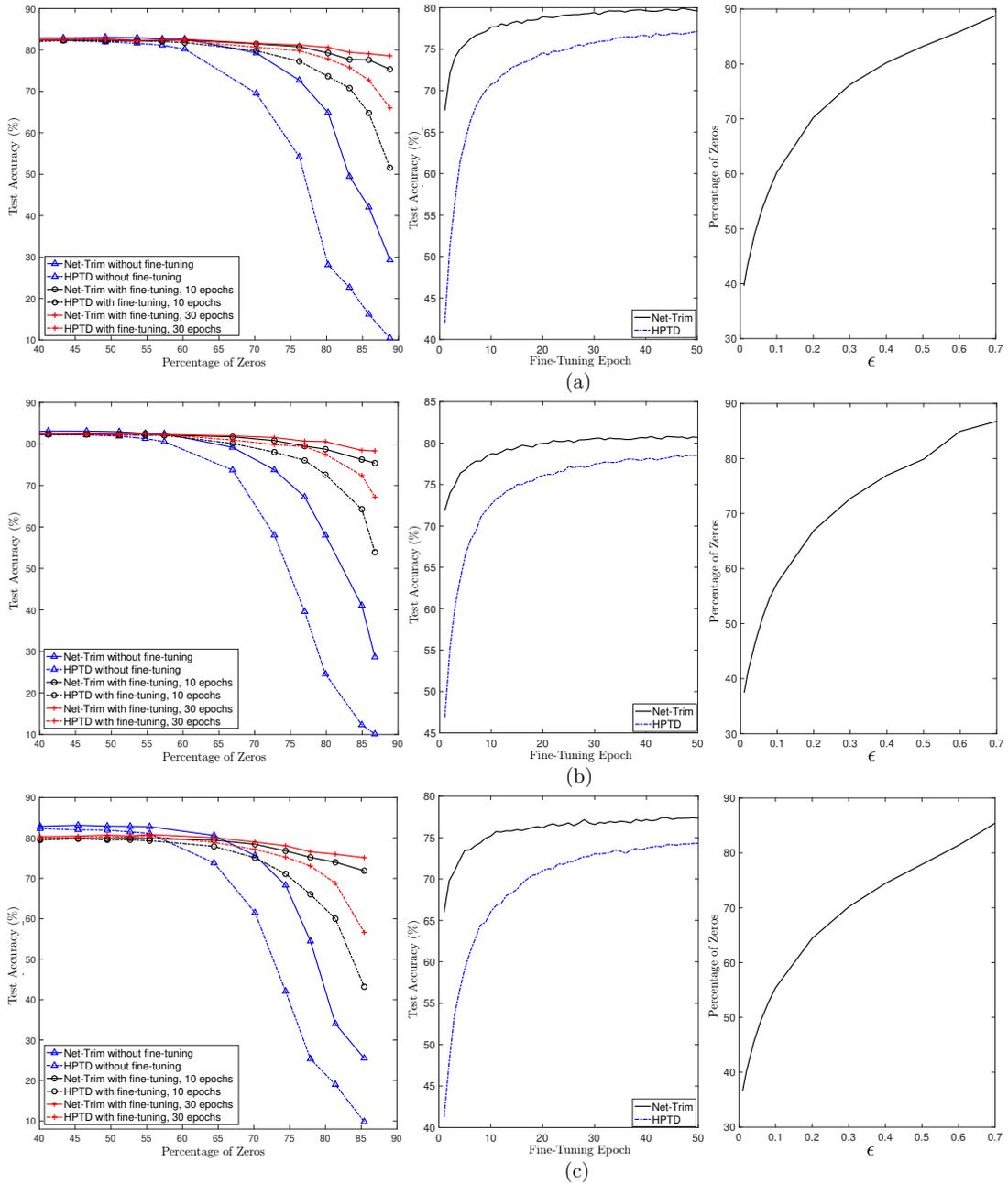

Figure \ref{fig6} demonstrates another set of comparative experiments between Net-Trim and HPTD, performed on a much larger augmented dataset. The reference training set is the CIFAR10 color-image database, which contains 50K samples of size $32\times 32$ from ten classes \cite{krizhevskyconvolutional, KrizhevskySH12}. 

%
In order to obtain higher test accuracies, the training images are multiplicated by 
taking $24\times 24$ windows to randomly crop them, and each cropped image is horizontally flipped with probability $0.5$. This process augments the training set to 6,400,000 samples. The neural network employed to address the initial classification problem is convolutional, where the first layer of the trained network uses 64 filters of size $5\times 5\times 3$, followed by a max pooling unit (size: $3\times 3$, stride: $2\times 2$). The second layer is also convolutional with 64 filters of size $5\times 5\times 64$ and a similar max pooling unit. The remainder of the network contains three fully connected layers ($3136\times 384\times 192\times 10$).

For this relatively large dataset we also go through the exercise of retraining the Net-Trim with only part of the training samples, specifically 25K, 50K and 75K samples of the entire 6.4M training set. A similar set of comparisons between the Net-Trim and the HPTD as in Figure \ref{fig4} is performed, noting that the fine-tuning step for both schemes is carried out using all the training samples. Similar to the previous experiment, Net-Trim consistently outperforms HPTD in all similar setups. Aside from such superiority, we highlight the possibility of retraining Net-Trim using only part of the training samples. For instance, a comparison of panels (a) and (b) shows that almost identical results can be achieved in terms of accuracy versus sparsity, when Net-Trim is solved with 25K samples instead of 50K samples. For instance, for both panels, a Net-Trim application followed by a single fine-tuning step can increase the percentage of the zeros in the network to more than 80\%, with almost no loss in the model accuracy. Basically, as also discussed previously with reference to Figure \ref{fig2m}, for large data sets formulating the Net-Trim with only a portion of the data can be considered as a general computation shortcut. 

Another interesting observation, which is more apparent on the left plot of panel (c), is that fine-tuning does not always improve the accuracy of the models after the application of Net-Trim, and especially in low pruning regimes may cause degrading the accuracy due to phenomena such as overfitting.  For example, in panel (c), up to a pruning percentage of almost 65\%, a fine-tuning step after the Net-Trim slightly degrades the accuracy. While a fine-tuning step is likely to help in the majority of cases, our access to both Net-Trim's plain outcome, and the fine-tuned version provides the flexibility of picking the most compressed and accurate model among the two.

\end{document}